\documentclass[sigconf]{acmart}
\AtBeginDocument{%
  }

\copyrightyear{2024}
\acmYear{2024}
\setcopyright{rightsretained}
\acmConference[FAccT '24]{The 2024 ACM Conference on Fairness, Accountability, and Transparency}{June
3--6, 2024}{Rio de Janeiro, Brazil}
\acmBooktitle{The 2024 ACM Conference on Fairness, Accountability, and Transparency (FAccT '24), June 3--6,
2024, Rio de Janeiro, Brazil}\acmDOI{10.1145/3630106.3658921}
\acmISBN{979-8-4007-0450-5/24/06}

\usepackage{mathtools}

\usepackage{url,stfloats}
\usepackage{subfigure}

\usepackage{amsfonts,bm}

\DeclareMathOperator*{\argmin}{arg\,min}

\usepackage{pifont}
\newcommand{\xmark}{\ding{55}}%
\theoremstyle{plain}
\newtheorem{theorem}{Theorem}[section]

\newtheorem{corollary}[theorem]{Corollary}
\theoremstyle{definition}

\newtheorem{example}[theorem]{Example}
\theoremstyle{remark}

\usepackage[capitalize,noabbrev]{cleveref}
\begin{document}

\def\eps{{\epsilon}}

\newcommand{\hr}{x_{\operatorname{HR}}}
\newcommand{\lhr}{l_{\operatorname{HR}}}
\newcommand{\hathr}{\hat{x}_{\operatorname{HR}}}
\newcommand{\hatlhr}{\hat{l}_{\operatorname{HR}}}
\newcommand{\lr}{x_{\operatorname{LR}}}
\newcommand{\dataset}{UnfairFace}
\newcommand{\myP}{\mathbb{P}}
\newcommand{\Phr}{\mathbb{P}_{\operatorname{HR}}}
\newcommand{\Px}{\mathbb{P}_{x}}
\newcommand{\PAy}{\mathbb{P}_{\mathcal{A}\vert y}}
\newcommand{\Pxy}{\mathbb{P}_{x, y}}
\newcommand{\Phatxy}{\mathbb{P}_{\hat{x}\vert y}}
\newcommand{\Phatx}{\mathbb{P}_{\hat{x}}}
\newcommand{\Phatxx}{\mathbb{P}_{\hat{x} \vert x}}
\newcommand{\Phatxyi}{\mathbb{P}_{\hat{x}\vert y^{(i)}}}
\newcommand{\Py}{\mathbb{P}_{y}}
\newcommand{\Pycondx}{\mathbb{P}_{y \vert x}}
\newcommand{\Pxcondy}{\mathbb{P}_{x \vert y}}
\newcommand{\PA}{\mathbb{P}_{\mathcal{A}}}
\newcommand{\PAx}{\mathbb{P}_{\mathcal{A} \vert x}}
\newcommand{\Phathrlr}{\mathbb{P}_{\hat{\operatorname{HR}} \vert \operatorname{LR}}}
\newcommand{\Phathrhr}{\mathbb{P}_{\hat{\operatorname{HR}} \vert \operatorname{HR}}}
\newcommand{\Phathr}{\mathbb{P}_{\hat{\operatorname{HR}}}}
\newcommand{\Ehr}{\mathbb{E}_{\hr \sim \Phr}}
\newcommand{\Prdp}{\mathbb{P}_{\operatorname{RDP}}}
\newcommand{\Ppr}{\mathbb{P}_{\operatorname{PR}}}
\newcommand{\Pcpr}{\mathbb{P}_{\operatorname{CPR}}}
\newcommand{\Pcprcond}{\mathbb{P}_{\operatorname{CPR}\vert \lr}}
\newcommand{\Pcprcondy}{\mathbb{P}_{\operatorname{CPR}\vert y}}
\newcommand{\Pucpr}{\mathbb{P}_{\operatorname{UCPR}}}

\def\reta{{\textnormal{$\eta$}}}
\def\ra{{\textnormal{a}}}
\def\rb{{\textnormal{b}}}
\def\rc{{\textnormal{c}}}
\def\rd{{\textnormal{d}}}
\def\re{{\textnormal{e}}}
\def\rf{{\textnormal{f}}}
\def\rg{{\textnormal{g}}}
\def\rh{{\textnormal{h}}}
\def\ri{{\textnormal{i}}}
\def\rj{{\textnormal{j}}}
\def\rk{{\textnormal{k}}}
\def\rl{{\textnormal{l}}}
\def\rn{{\textnormal{n}}}
\def\ro{{\textnormal{o}}}
\def\rp{{\textnormal{p}}}
\def\rq{{\textnormal{q}}}
\def\rr{{\textnormal{r}}}
\def\rs{{\textnormal{s}}}
\def\rt{{\textnormal{t}}}
\def\ru{{\textnormal{u}}}
\def\rv{{\textnormal{v}}}
\def\rw{{\textnormal{w}}}
\def\rx{{\textnormal{x}}}
\def\ry{{\textnormal{y}}}
\def\rz{{\textnormal{z}}}

\def\rvepsilon{{\mathbf{\epsilon}}}
\def\rvtheta{{\mathbf{\theta}}}
\def\rva{{\mathbf{a}}}
\def\rvb{{\mathbf{b}}}
\def\rvc{{\mathbf{c}}}
\def\rvd{{\mathbf{d}}}
\def\rve{{\mathbf{e}}}
\def\rvf{{\mathbf{f}}}
\def\rvg{{\mathbf{g}}}
\def\rvh{{\mathbf{h}}}
\def\rvu{{\mathbf{i}}}
\def\rvj{{\mathbf{j}}}
\def\rvk{{\mathbf{k}}}
\def\rvl{{\mathbf{l}}}
\def\rvm{{\mathbf{m}}}
\def\rvn{{\mathbf{n}}}
\def\rvo{{\mathbf{o}}}
\def\rvp{{\mathbf{p}}}
\def\rvq{{\mathbf{q}}}
\def\rvr{{\mathbf{r}}}
\def\rvs{{\mathbf{s}}}
\def\rvt{{\mathbf{t}}}
\def\rvu{{\mathbf{u}}}
\def\rvv{{\mathbf{v}}}
\def\rvw{{\mathbf{w}}}
\def\rvx{{\mathbf{x}}}
\def\rvy{{\mathbf{y}}}
\def\rvz{{\mathbf{z}}}

\def\erva{{\textnormal{a}}}
\def\ervb{{\textnormal{b}}}
\def\ervc{{\textnormal{c}}}
\def\ervd{{\textnormal{d}}}
\def\erve{{\textnormal{e}}}
\def\ervf{{\textnormal{f}}}
\def\ervg{{\textnormal{g}}}
\def\ervh{{\textnormal{h}}}
\def\ervi{{\textnormal{i}}}
\def\ervj{{\textnormal{j}}}
\def\ervk{{\textnormal{k}}}
\def\ervl{{\textnormal{l}}}
\def\ervm{{\textnormal{m}}}
\def\ervn{{\textnormal{n}}}
\def\ervo{{\textnormal{o}}}
\def\ervp{{\textnormal{p}}}
\def\ervq{{\textnormal{q}}}
\def\ervr{{\textnormal{r}}}
\def\ervs{{\textnormal{s}}}
\def\ervt{{\textnormal{t}}}
\def\ervu{{\textnormal{u}}}
\def\ervv{{\textnormal{v}}}
\def\ervw{{\textnormal{w}}}
\def\ervx{{\textnormal{x}}}
\def\ervy{{\textnormal{y}}}
\def\ervz{{\textnormal{z}}}

\def\rmA{{\mathbf{A}}}
\def\rmB{{\mathbf{B}}}
\def\rmC{{\mathbf{C}}}
\def\rmD{{\mathbf{D}}}
\def\rmE{{\mathbf{E}}}
\def\rmF{{\mathbf{F}}}
\def\rmG{{\mathbf{G}}}
\def\rmH{{\mathbf{H}}}
\def\rmI{{\mathbf{I}}}
\def\rmJ{{\mathbf{J}}}
\def\rmK{{\mathbf{K}}}
\def\rmL{{\mathbf{L}}}
\def\rmM{{\mathbf{M}}}
\def\rmN{{\mathbf{N}}}
\def\rmO{{\mathbf{O}}}
\def\rmP{{\mathbf{P}}}
\def\rmQ{{\mathbf{Q}}}
\def\rmR{{\mathbf{R}}}
\def\rmS{{\mathbf{S}}}
\def\rmT{{\mathbf{T}}}
\def\rmU{{\mathbf{U}}}
\def\rmV{{\mathbf{V}}}
\def\rmW{{\mathbf{W}}}
\def\rmX{{\mathbf{X}}}
\def\rmY{{\mathbf{Y}}}
\def\rmZ{{\mathbf{Z}}}

\def\ermA{{\textnormal{A}}}
\def\ermB{{\textnormal{B}}}
\def\ermC{{\textnormal{C}}}
\def\ermD{{\textnormal{D}}}
\def\ermE{{\textnormal{E}}}
\def\ermF{{\textnormal{F}}}
\def\ermG{{\textnormal{G}}}
\def\ermH{{\textnormal{H}}}
\def\ermI{{\textnormal{I}}}
\def\ermJ{{\textnormal{J}}}
\def\ermK{{\textnormal{K}}}
\def\ermL{{\textnormal{L}}}
\def\ermM{{\textnormal{M}}}
\def\ermN{{\textnormal{N}}}
\def\ermO{{\textnormal{O}}}
\def\ermP{{\textnormal{P}}}
\def\ermQ{{\textnormal{Q}}}
\def\ermR{{\textnormal{R}}}
\def\ermS{{\textnormal{S}}}
\def\ermT{{\textnormal{T}}}
\def\ermU{{\textnormal{U}}}
\def\ermV{{\textnormal{V}}}
\def\ermW{{\textnormal{W}}}
\def\ermX{{\textnormal{X}}}
\def\ermY{{\textnormal{Y}}}
\def\ermZ{{\textnormal{Z}}}

\def\vzero{{\mathbf{0}}}
\def\vone{{\mathbf{1}}}
\def\vmu{{\mathbf{\mu}}}
\def\vtheta{{\boldsymbol{\theta}}}
\def\va{{\mathbf{a}}}
\def\vb{{\mathbf{b}}}
\def\vc{{\mathbf{c}}}
\def\vd{{\mathbf{d}}}
\def\ve{{\mathbf{e}}}
\def\vf{{\mathbf{f}}}
\def\vg{{\mathbf{g}}}
\def\vh{{\mathbf{h}}}
\def\vi{{\mathbf{i}}}
\def\vj{{\mathbf{j}}}
\def\vk{{\mathbf{k}}}
\def\vl{{\mathbf{l}}}
\def\vm{{\mathbf{m}}}
\def\vn{{\mathbf{n}}}
\def\vo{{\mathbf{o}}}
\def\vp{{\mathbf{p}}}
\def\vq{{\mathbf{q}}}
\def\vr{{\mathbf{r}}}
\def\vs{{\mathbf{s}}}
\def\vt{{\mathbf{t}}}
\def\vu{{\mathbf{u}}}
\def\vv{{\mathbf{v}}}
\def\vw{{\mathbf{w}}}
\def\vx{{\mathbf{x}}}
\def\vy{{\mathbf{y}}}
\def\vz{{\mathbf{z}}}

\def\evalpha{{\alpha}}
\def\evbeta{{\beta}}
\def\evepsilon{{\epsilon}}
\def\evlambda{{\lambda}}
\def\evomega{{\omega}}
\def\evmu{{\mu}}
\def\evpsi{{\psi}}
\def\evsigma{{\sigma}}
\def\evtheta{{\theta}}
\def\eva{{a}}
\def\evb{{b}}
\def\evc{{c}}
\def\evd{{d}}
\def\eve{{e}}
\def\evf{{f}}
\def\evg{{g}}
\def\evh{{h}}
\def\evi{{i}}
\def\evj{{j}}
\def\evk{{k}}
\def\evl{{l}}
\def\evm{{m}}
\def\evn{{n}}
\def\evo{{o}}
\def\evp{{p}}
\def\evq{{q}}
\def\evr{{r}}
\def\evs{{s}}
\def\evt{{t}}
\def\evu{{u}}
\def\evv{{v}}
\def\evw{{w}}
\def\evx{{x}}
\def\evy{{y}}
\def\evz{{z}}

\def\mA{{\mathbf{A}}}
\def\mB{{\mathbf{B}}}
\def\mC{{\mathbf{C}}}
\def\mD{{\mathbf{D}}}
\def\mE{{\mathbf{E}}}
\def\mF{{\mathbf{F}}}
\def\mG{{\mathbf{G}}}
\def\mH{{\mathbf{H}}}
\def\mI{{\mathbf{I}}}
\def\mJ{{\mathbf{J}}}
\def\mK{{\mathbf{K}}}
\def\mL{{\mathbf{L}}}
\def\mM{{\mathbf{M}}}
\def\mN{{\mathbf{N}}}
\def\mO{{\mathbf{O}}}
\def\mP{{\mathbf{P}}}
\def\mQ{{\mathbf{Q}}}
\def\mR{{\mathbf{R}}}
\def\mS{{\mathbf{S}}}
\def\mT{{\mathbf{T}}}
\def\mU{{\mathbf{U}}}
\def\mV{{\mathbf{V}}}
\def\mW{{\mathbf{W}}}
\def\mX{{\mathbf{X}}}
\def\mY{{\mathbf{Y}}}
\def\mZ{{\mathbf{Z}}}
\def\mBeta{{\mathbf{\beta}}}
\def\mPhi{{\mathbf{\Phi}}}
\def\mLambda{{\mathbf{\Lambda}}}
\def\mSigma{{\mathbf{\Sigma}}}

\newcommand{\tens}[1]{\bm{\mathsfit{#1}}}
\def\tA{{\tens{A}}}
\def\tB{{\tens{B}}}
\def\tC{{\tens{C}}}
\def\tD{{\tens{D}}}
\def\tE{{\tens{E}}}
\def\tF{{\tens{F}}}
\def\tG{{\tens{G}}}
\def\tH{{\tens{H}}}
\def\tI{{\tens{I}}}
\def\tJ{{\tens{J}}}
\def\tK{{\tens{K}}}
\def\tL{{\tens{L}}}
\def\tM{{\tens{M}}}
\def\tN{{\tens{N}}}
\def\tO{{\tens{O}}}
\def\tP{{\tens{P}}}
\def\tQ{{\tens{Q}}}
\def\tR{{\tens{R}}}
\def\tS{{\tens{S}}}
\def\tT{{\tens{T}}}
\def\tU{{\tens{U}}}
\def\tV{{\tens{V}}}
\def\tW{{\tens{W}}}
\def\tX{{\tens{X}}}
\def\tY{{\tens{Y}}}
\def\tZ{{\tens{Z}}}

\def\gA{{\mathcal{A}}}
\def\gB{{\mathcal{B}}}
\def\gC{{\mathcal{C}}}
\def\gD{{\mathcal{D}}}
\def\gE{{\mathcal{E}}}
\def\gF{{\mathcal{F}}}
\def\gG{{\mathcal{G}}}
\def\gH{{\mathcal{H}}}
\def\gI{{\mathcal{I}}}
\def\gJ{{\mathcal{J}}}
\def\gK{{\mathcal{K}}}
\def\gL{{\mathcal{L}}}
\def\gM{{\mathcal{M}}}
\def\gN{{\mathcal{N}}}
\def\gO{{\mathcal{O}}}
\def\gP{{\mathcal{P}}}
\def\gQ{{\mathcal{Q}}}
\def\gR{{\mathcal{R}}}
\def\gS{{\mathcal{S}}}
\def\gT{{\mathcal{T}}}
\def\gU{{\mathcal{U}}}
\def\gV{{\mathcal{V}}}
\def\gW{{\mathcal{W}}}
\def\gX{{\mathcal{X}}}
\def\gY{{\mathcal{Y}}}
\def\gZ{{\mathcal{Z}}}

\def\sA{{\mathbb{A}}}
\def\sB{{\mathbb{B}}}
\def\sC{{\mathbb{C}}}
\def\sD{{\mathbb{D}}}
\def\sF{{\mathbb{F}}}
\def\sG{{\mathbb{G}}}
\def\sH{{\mathbb{H}}}
\def\sI{{\mathbb{I}}}
\def\sJ{{\mathbb{J}}}
\def\sK{{\mathbb{K}}}
\def\sL{{\mathbb{L}}}
\def\sM{{\mathbb{M}}}
\def\sN{{\mathbb{N}}}
\def\sO{{\mathbb{O}}}
\def\sP{{\mathbb{P}}}
\def\sQ{{\mathbb{Q}}}
\def\sR{{\mathbb{R}}}
\def\sS{{\mathbb{S}}}
\def\sT{{\mathbb{T}}}
\def\sU{{\mathbb{U}}}
\def\sV{{\mathbb{V}}}
\def\sW{{\mathbb{W}}}
\def\sX{{\mathbb{X}}}
\def\sY{{\mathbb{Y}}}
\def\sZ{{\mathbb{Z}}}

\def\emLambda{{\Lambda}}
\def\emA{{A}}
\def\emB{{B}}
\def\emC{{C}}
\def\emD{{D}}
\def\emE{{E}}
\def\emF{{F}}
\def\emG{{G}}
\def\emH{{H}}
\def\emI{{I}}
\def\emJ{{J}}
\def\emK{{K}}
\def\emL{{L}}
\def\emM{{M}}
\def\emN{{N}}
\def\emO{{O}}
\def\emP{{P}}
\def\emQ{{Q}}
\def\emR{{R}}
\def\emS{{S}}
\def\emT{{T}}
\def\emU{{U}}
\def\emV{{V}}
\def\emW{{W}}
\def\emX{{X}}
\def\emY{{Y}}
\def\emZ{{Z}}
\def\emSigma{{\Sigma}}

\newcommand{\etens}[1]{\mathsfit{#1}}
\def\etLambda{{\etens{\Lambda}}}
\def\etA{{\etens{A}}}
\def\etB{{\etens{B}}}
\def\etC{{\etens{C}}}
\def\etD{{\etens{D}}}
\def\etE{{\etens{E}}}
\def\etF{{\etens{F}}}
\def\etG{{\etens{G}}}
\def\etH{{\etens{H}}}
\def\etI{{\etens{I}}}
\def\etJ{{\etens{J}}}
\def\etK{{\etens{K}}}
\def\etL{{\etens{L}}}
\def\etM{{\etens{M}}}
\def\etN{{\etens{N}}}
\def\etO{{\etens{O}}}
\def\etP{{\etens{P}}}
\def\etQ{{\etens{Q}}}
\def\etR{{\etens{R}}}
\def\etS{{\etens{S}}}
\def\etT{{\etens{T}}}
\def\etU{{\etens{U}}}
\def\etV{{\etens{V}}}
\def\etW{{\etens{W}}}
\def\etX{{\etens{X}}}
\def\etY{{\etens{Y}}}
\def\etZ{{\etens{Z}}}

\newcommand{\pdata}{p_{\rm{data}}}
\newcommand{\ptrain}{\hat{p}_{\rm{data}}}
\newcommand{\Ptrain}{\hat{P}_{\rm{data}}}
\newcommand{\pmodel}{p_{\rm{model}}}
\newcommand{\Pmodel}{P_{\rm{model}}}
\newcommand{\ptildemodel}{\tilde{p}_{\rm{model}}}
\newcommand{\pencode}{p_{\rm{encoder}}}
\newcommand{\pdecode}{p_{\rm{decoder}}}
\newcommand{\precons}{p_{\rm{reconstruct}}}

\newcommand{\laplace}{\mathrm{Laplace}} %

\newcommand{\E}{\mathbb{E}}
\newcommand{\Ls}{\mathcal{L}}
\newcommand{\R}{\mathbb{R}}
\newcommand{\emp}{\tilde{p}}
\newcommand{\reg}{\lambda}
\newcommand{\rect}{\mathrm{rectifier}}
\newcommand{\softmax}{\mathrm{softmax}}
\newcommand{\sigmoid}{\sigma}
\newcommand{\softplus}{\zeta}
\newcommand{\KL}{D_{\mathrm{KL}}}
\newcommand{\Var}{\mathrm{Var}}
\newcommand{\standarderror}{\mathrm{SE}}
\newcommand{\Cov}{\mathrm{Cov}}
\newcommand{\normlzero}{L^0}
\newcommand{\normlone}{L^1}
\newcommand{\normltwo}{L^2}
\newcommand{\normlp}{L^p}
\newcommand{\normmax}{L^\infty}

\newcommand{\parents}{Pa} %

\newcommand{\dint}{\text{d}}
\newcommand{\vphi}{\boldsymbol{\phi}}
\newcommand{\vpi}{\boldsymbol{\pi}}
\newcommand{\vpsi}{\boldsymbol{\psi}}
\newcommand{\vomg}{\boldsymbol{\omega}}
\newcommand{\vsigma}{\boldsymbol{\sigma}}
\newcommand{\vzeta}{\boldsymbol{\zeta}}
\renewcommand{\vx}{\mathbf{x}}
\renewcommand{\vy}{\mathbf{y}}
\renewcommand{\vz}{\mathbf{z}}
\renewcommand{\vh}{\mathbf{h}}
\renewcommand{\b}{\mathbf}
\renewcommand{\vec}{\text{vec}}
\newcommand{\vecemph}{\text{\emph{vec}}}
\newcommand{\mvn}{\mathcal{MN}}
\newcommand{\N}{\mathcal{N}}
\newcommand{\diag}[1]{\text{diag}(#1)}
\newcommand{\diagemph}[1]{\text{\emph{diag}}(#1)}
\newcommand{\tr}[1]{\text{tr}(#1)}
\renewcommand{\R}{\mathbb{R}}
\renewcommand{\E}{\mathbb{E}}
\newcommand{\D}{\mathcal{D}}

\title{Benchmarking the Fairness of Image Upsampling Methods}

\author{Mike Laszkiewicz}
\email{Mike.Laszkiewicz@rub.de}
\affiliation{%
  \institution{Faculty of Computer Science, Ruhr University}
\streetaddress{Universitaetsstrasse 140}
  \city{Bochum}
  \state{North Rhine-Westphalia}
  \country{Germany}
  \postcode{44801}
}

\author{Imant Daunhawer}
\affiliation{%
  \institution{Department of Computer Science, ETH Zurich}
  \streetaddress{Universitaetsstrasse 6}
  \city{Zurich}
  \state{Zurich}
  \country{Switzerland}
  \postcode{8092}
}

\author{Julia E. Vogt}
\authornote{Joint supervision.}
\affiliation{%
  \institution{Department of Computer Science, ETH Zurich}
\streetaddress{Universitaetsstrasse 6}
  \city{Zurich}
  \state{Zurich}
  \country{Switzerland}
  \postcode{8092}
}

\author{Asja Fischer}
\authornotemark[1]
\affiliation{%
  \institution{Faculty of Computer Science, Ruhr University}
\streetaddress{Universitaetsstrasse 140}
  \city{Bochum}
  \state{North Rhine-Westphalia}
  \country{Germany}
  \postcode{44801}
}

\author{Johannes Lederer}
\authornotemark[1]
\affiliation{%
  \institution{Department of Mathematics, Computer Science, and Natural Sciences, University of Hamburg}
\streetaddress{Bundesstrasse 55}
  \city{Hamburg}
  \state{Hamburg}
  \country{Germany}
  \postcode{20146}
}

\renewcommand{\shortauthors}{Laszkiewicz et al.}

\begin{abstract}
    Recent years have witnessed a rapid development of deep generative models for creating synthetic media, such as images and videos.  
    While the practical applications of these models in everyday tasks are enticing, it is crucial to assess the inherent risks regarding their fairness. 
    In this work, we introduce a %
    comprehensive
    framework for benchmarking the performance and fairness of conditional generative models.
    We develop a set of metrics---inspired by their supervised fairness counterparts---to evaluate the models on their fairness and diversity.  
    Focusing on the specific application of image upsampling, we create a benchmark covering a wide variety of modern upsampling methods. %
    As part of the benchmark, we introduce \dataset{}, 
    a subset of FairFace that replicates the racial distribution of common large-scale face datasets.
    Our empirical study highlights the importance of using an unbiased training set and reveals variations in how the algorithms respond to dataset imbalances. 
    Alarmingly, we find that none of the considered methods produces statistically fair and diverse results. All experiments can be reproduced using our provided repository.\footnote{\url{https://github.com/MikeLasz/Benchmarking-Fairness-ImageUpsampling}}
\end{abstract}



\begin{teaserfigure}
  \centering
  \includegraphics[width=\textwidth]{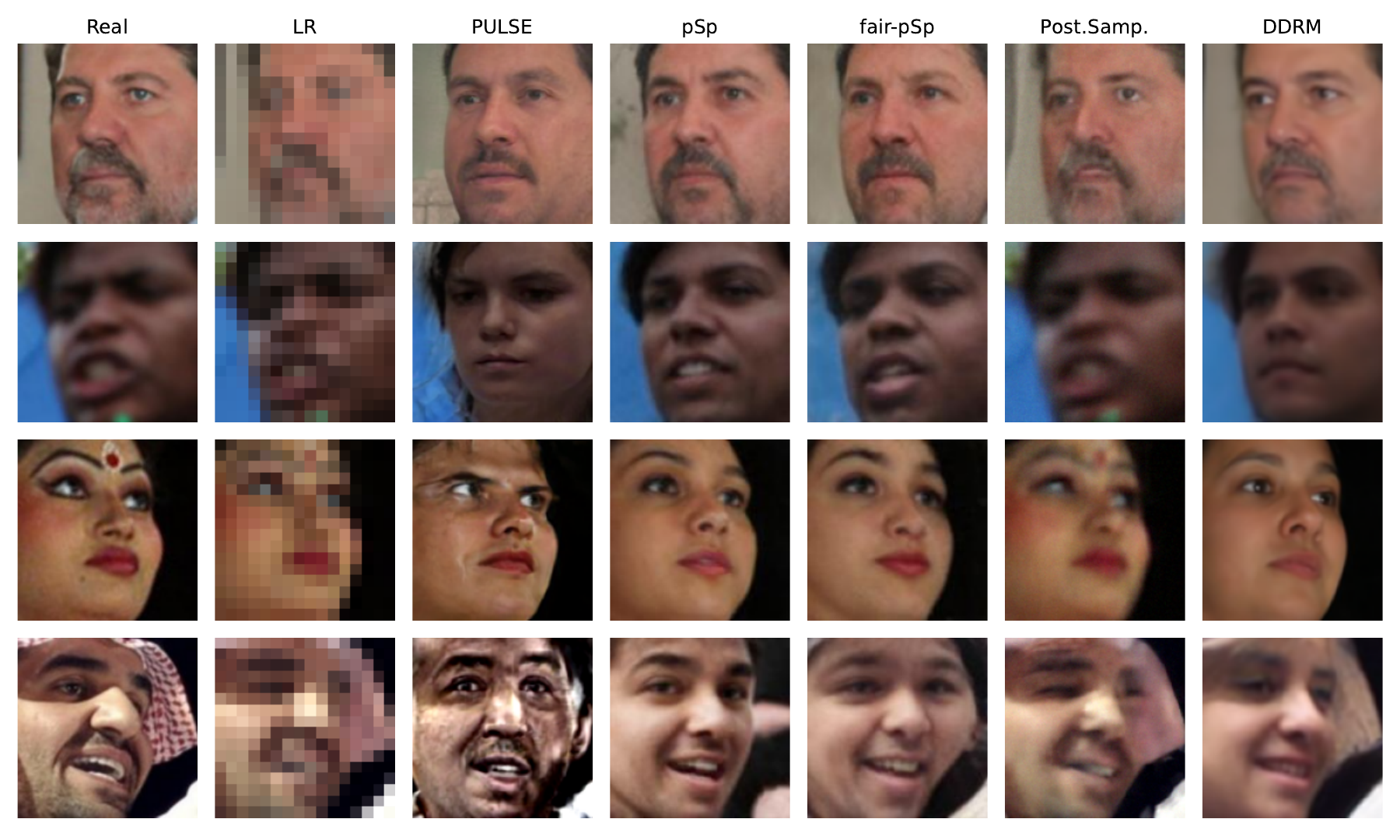}
  \caption[]{Upsampling algorithms fail to reconstruct samples from minority groups: Reconstructing low-resolution images using models trained on \dataset{}, a subset of the FairFace dataset replicating the racial distribution of common large-scale datasets.}
  \label{fig:teaser}
  \Description{Comparing the reconstructions based on several image upsampling algorithms. The models were trained on UnfairFace.}
\end{teaserfigure}

\settopmatter{printfolios=true}
\maketitle

\section{Introduction}
Remarkable advancements in deep generative models 
(e.g. \citep{karras2021alias, Sauer2021, rombach2022high, hatamizadeh2023diffit}) give rise to striking applications, including image inpainting (e.g. \citep{richardson2021encoding}), style transfer (e.g. \citep{CycleGAN2017}), image superresolution (e.g. \citep{chen2023activating}), and text-to-image translation (e.g. \citep{ramesh2022hierarchical}).
Despite their apparent success, our understanding of these models and the biases they are subject to lags behind, often revealing unexpected fairness violations. 
Earlier research on fairness in deep learning discovered biases in supervised prediction models, leading to racist \citep{police_racist, zong2022medfair}, sexist \citep{sexist_ml, google_sexist}, and homophobic \citep{google_homophobic} decisions, which can be critical in many applications, such as criminal justice, medicine, hiring, or personalized advertising.
These observations sparked a general interest in a research field that concentrates on developing and quantifying fair machine learning models \citep{review_fairness, du2020fairness, barocas-hardt-narayanan, zong2022medfair}. 
However, compared to the supervised setting, there is a noticeable gap in addressing fairness in generative models~\citep{choi2020fair, humayun2022magnet} and conditional generative models~\citep{jalal2021fairness, tanjim2022debiasing}.

In this work, we focus on the latter and investigate conditional generative models regarding their fairness and diversity. %
We introduce a novel set of metrics that serves as a practical tool for assessing fairness violations of conditional generative models. These metrics are associated with the fairness definitions introduced by~\citet{jalal2021fairness} in the sense that fairness is achieved if and only if our fairness violation metrics equal zero. 
These metrics form the foundation of a comprehensive framework that supplements standard performance criteria with fairness and diversity metrics. Based on these metrics, we design a versatile benchmark, facilitating a unified evaluation of the performance, fairness, and diversity of a wide variety of conditional generative models.
We showcase the capacities of the proposed framework by evaluating one particular type of conditional generative modeling, namely \emph{image upsampling}, which has triggered a compelling discussion on racial biases in deep generative models~\citep{obama_vice}.
We leverage the FairFace dataset~\citep{karkkainen2021fairface} to introduce \emph{\dataset{}}, a subset of FairFace mimicking the distribution of common large-scale face datasets. Because FairFace provides race\footnote{We specify race in terms of phenotype and appearance, and ethnicity as a shared culture~\citep{balestra2018diversity, canadian2022guidance}. We are aware that race and ethnicity are social constructs and that our definitions do not reflect the underlying complexities apparent in society. An elaborate ethics statement is provided in Section~\ref{sec:ethical}.} labels, it allows us to conduct an in-depth empirical evaluation of a range of upsampling methods and their influence on dataset bias. 
Our evaluation demonstrates that, while none of the methods are significantly fair, the degree of fairness varies significantly across methods. Unsurprisingly, we find that a racial bias in the training data reduces the fairness of most models. Notably, the effect is most apparent for Denoising Diffusion Restoration Models (DDRM~\citep{kawar2022denoising}), which
exhibit the highest discrepancy to fairness among all investigated models when trained on \dataset{}, but the least discrepancy to fairness when trained on FairFace. 
In contrast, the fairness discrepancy of Posterior Sampling~\citep{jalal2021fairness} remains robust despite biased training data. 
These findings highlight the importance of careful data acquisition and emphasize how biases present in the dataset can impact model selection.

\paragraph{Notation}
In this work, we define performance and fairness using probabilistic terminology. Specifically, we denote target samples by $x\in \mathcal{X}$ and conditional arguments by $y\in \mathcal{Y}$. Target samples and conditions follow the distribution $\Pxy$ with marginals $\Px$ and $\Py$, respectively. 
$\Pycondx$  denotes the conditional distribution of $y$  given a target $x$ and  $\Pxcondy$ the conditional distribution of $x$ given a condition $y$. The conditional generative model generates samples $\hat{x}$ conditioned on $y$, which follows the distribution $\Phatxy$. 
We denote the distribution of $\Phatxy$ marginalized over all $y$ by $\Phatx$.
Moreover, we denote by $\Phatxx$ the conditional distribution 
that one obtains by marginalizing the product $\Phatxy \cdot \Pycondx$ over all $y$.  
For the sake of clarity, we use the notation $\myP{}(\hat{x} \in A \vert\,  x\in B) := \Phatxx(A\vert\, B)$ for measurable sets $A, \, B$ in the domain of $\hat{x}$ and $x$, respectively. We use a similar notation for all other distributions introduced in this paragraph. 

\section{Related Work}

\subsection{Fairness for Supervised Models and Unconditional Generative Models}
The first investigations on fairness of machine learning models were conducted for supervised models\citep{mehrabi2021survey}. Research in that field has led to several approaches, such as fair representation learning~\citep{zemel2013learning, xu2018fairgan, sattigeri2018fairness, balunovic2021fair},  constrained optimization~\citep{agarwal2018reductions, oneto2020general}, adversarial training~\citep{edwards2015censoring}, and post-processing methods~\citep{hardt2016equality, chzhen2020fair}. 
These works introduced numerous different definitions of fairness---for instance, demographic parity~\citep{dwork2012fairness} and equalized odds~\citep{hardt2016equality}---that guarantee that the model predictions are not driven by sensitive features. These definitions do not apply to unconditional generative models, whose aim is to generate synthetic data. Instead, fairness for unconditional generative models is typically phrased in terms of feature diversity, which is often violated in practice. 
For instance, deep generative models have been observed to generate less diverse samples \citep{jain2020imperfect}, which is amplified by common practices such as the truncation trick \citep{maluleke2022studying}.
Moreover, empirical results demonstrate that 
an increase in model performance correlates with a decrease in sample diversity \citep{salminen, zameshina2022fairness}.
To resolve the issue, given a fair reference dataset, \citet{choi2020fair} and  \citet{zameshina2022fairness} enhance the diversity by training a model with importance weighting and stratification, \citet{yu2020inclusive} introduce another training loss based on implicit maximum likelihood, \citet{teo2022fair} apply modern transfer learning techniques to the reference dataset, whereas other works refine the diversity of the generative models by latent code manipulation \citep{tan2020improving, humayun2022magnet}. 

\subsection{Fairness for Conditional Generative Models}\label{subsec:prior_fairness_cond}
Conditional generative models create a new sample $\hat{x}$ given a condition $y$, which could, for instance, be a low-resolution image in image upsampling (e.g. \citet{menon2020pulse}), a cropped image in image inpainting (e.g. \citet{richardson2021encoding}), or a sentence in text-to-image synthesis (e.g. \citet{rombach2022high}).  
Since the condition $y$ specifies certain features that should be reflected in the generated samples, it might be too restrictive to assess the fairness of a generative model in terms of sample diversity. 
Doing so would ignore the coherence between the condition and the generated samples. 
To address this, \citet{jalal2021fairness} extend traditional group-fairness \citep{dwork2012fairness} measures for conditional generative models to define \emph{representation demographic parity} (RDP), \emph{proportional representation} (PR), and \emph{conditional proportional representation} (CPR). 
In particular, let $\mathcal{C}=\{C_1, \dots, C_k\}$ be a partition\footnote{Note, the metrics also generalize to non-disjoint covers of $\mathcal{X}$.} 
of the data space $\mathcal{X}$, defining $k$ classes. 
Then, RDP, PR, and CPR are satisfied if and only if 
\begin{align}
     && \myP(\hat{x} \in C_i \vert \, x \in C_i) &= \myP (\hat{x} \in C_j\vert \, x \in C_j) \enspace   \forall i,j \in [k], \, \forall x \in \mathcal{X}  ,   
    \label{eq:rdp} 
    \\
     && \myP(\hat{x} \in C_i) &= \myP (x  \in C_i) \enspace  \forall i\in [k] \enspace , 
    \label{eq:pr} && 
    \\
    && \myP (\hat{x} \in C_i \vert \, y) &= \myP (x \in C_i \vert \, y)  \enspace \forall i \in [k], \, \forall  y \in \mathcal{Y} \enspace , && 
    \label{eq:cpr} 
\end{align}
respectively, where $\mathcal{Y}$ denotes the space of possible conditions and $[k]:=\{1,\dots, k\}$. We provide further explanation and synthetic examples highlighting the difference between these definitions in Section~\ref{subsec:further_intuition}.

For instance, in face image upsampling we could have the high-resolution image manifold $\mathcal{X}$, the low-resolution image manifold $\mathcal{Y}$, and a partition $\mathcal{C}:=\{C_{\text{glasses}}, \, C_{\text{no-glasses}}\}$ into a set of people that do or do not wear glasses. 
In this specific example, \citet{tanjim2022debiasing} approximates equation~\eqref{eq:rdp} by measuring the performance of a classifier that predicts whether or not a given image depicts a person wearing glasses.

\subsection{Image Upsampling Methods} \label{sec:upsampling_methods}
In image upsampling, the goal is to create a high-resolution image given a low-resolution input. To make the notation more illustrative in this case, we denote the condition by $\lr$, the true high-resolution image by $\hr$, and the conditionally generated output by $\hathr$. 
A myriad of image upsampling methods have been introduced in recent years; we focus on five carefully selected methods representative of most previously proposed methods. To cover a diverse set of approaches, we select PULSE~\citep{menon2020pulse}, which performs gradient-based latent-code optimization of a pretrained GAN~\citep{jabbar2021survey}, Pixel2Style2Pixel~\citep{richardson2021encoding}, which utilizes an encoder to obtain latent-codes of a pretrained GAN, and denoising diffusion restoration models~\citep{kawar2022denoising} based on diffusion models~\citep{yang2023diffusion}. Furthermore, we include two models that were specially designed for fair image upsampling: Posterior Sampling~\citep{jalal2021fairness} and a fairness-improved version of Pixel2Style2Pixel~\citep{tanjim2022debiasing}, which we refer to as fair-pSp. 
In the following, we provide a brief overview of these methods. 

Upsampling in PULSE is conducted by fixing a pretrained generative model $G$ and then optimizing\footnote{The optimization procedure is accompanied by several regularization terms.} for a latent code $z^\ast:= \argmin_{z} \Vert \lr - DS(G(z))\Vert$, where $\lr$ is the low-resolution input and $DS$ is a downscaling operator. The resulting high-resolution input is given by $\hathr :=G(z^\ast)$. 
In contrast, Pixel2Style2Pixel (pSp) learns an encoder $E$ that can directly map from $\lr$ to the latent code $z^\ast = E(\lr)$ and returns $\hathr:=G(z^\ast)$. 
To improve the fairness of pSp during its training, \citet{tanjim2022debiasing} propose to use a stratified sampling scheme, contrastive learning, and a cross-entropy regularization based on the predicted fairness labels.  
Posterior Sampling generates samples by sampling from the posterior $p(\hr \vert \lr) \propto p(\lr \vert \hr) p(\hr)$. \citet{jalal2021fairness} implement Posterior Sampling by leveraging a NCSNv2 score-based unconditional model~\citep{song2020improved}, which gives access to $\nabla_{\hr} \log p(\hr)$. They model $p(\lr \vert \hr)$ by fixing a linear downscaling operation $A$ and modeling $\lr = A \hr + \varepsilon$, where $\varepsilon \sim \mathcal{N}(0, \sigma I )$ for some $\sigma > 0$. This allows to model the score function of $p(\hr \vert \lr)$ by $\nabla_{\hr} \log p(\hr \vert \lr) = \nabla_{\hr} \log p(\lr \vert \hr) + \nabla_{\hr} \log p(\hr)$ and to sample using Langevin dynamics~\citep{welling2011bayesian}.
Denoising diffusion restoration models (DDRM) also attempt to sample from $p(\hr\vert \lr)$, but, instead of running the diffusion process in pixel space, they propose running it in the spectral space of the downscaling operator $A$. This makes DDRM remarkably efficient, requiring only around 20 diffusion steps in the reported experiments.  

\section{Benchmarking Fairness of Conditional Generative Models}\label{sec:framework}
In this section, we introduce a collection of metrics that allow us to quantify the performance and fairness of conditional generative models. %
Although evaluating a model's performance is standard practice, investigating its fairness and diversity is often neglected. 
Hence, our framework expands conventional evaluation methodologies by systematically examining the fairness and diversity of conditional generative models.  
The following metrics are specifically designed for image data but can readily be generalized to other data types.   

\subsection{Performance} 
\label{subsec:metrics:upsampling-performance}
We categorize the performance criteria into three types: expected reconstruction losses, referenceless quality losses, and expected attribute reconstruction losses. While the first two groups are common, the expected attribute reconstruction loss is rarely considered. 

Formally, the expected reconstruction loss is defined as 
\begin{displaymath}
    \mathbb{E}_{x, y \sim \Pxy, \, \hat{x}\sim \Phatxy}  \bigl[ L_{\operatorname{rec}}(x , \hat{x} ) \bigr] \approx \frac{1}{n} \sum_{i=1}^n L_{\operatorname{rec}} \Bigl(x^{(i)} , \hat{x}^{(i)}\Bigr) \enspace , 
\end{displaymath}
where $x^{(i)}, y^{(i)}$ are samples from $\Pxy$, $\hat{x}^{(i)}$ are samples from $\Phatxyi$, and $L_{\operatorname{rec}}$ is a reconstruction loss.
For image data, we choose $L_{\operatorname{rec}}$ to be either the LPIPS distance~\citep{zhang2018perceptual}, or the structural dissimilarity 
\begin{displaymath}
    DSSIM(x, \hat{x}):= \frac{1 - SSIM(x, \hat{x})}{2} \in [0,1] \enspace , 
\end{displaymath}
where $SSIM$ denotes the structural similarity index measure \citep{wang2004image}. 

In certain conditional generation tasks, multiple outputs $x$ align with a given condition $y$. For instance, in text-to-image synthesis multiple images $x$ may depict images that correspond to an input text $y$, rendering a reconstruction loss inappropriate. Therefore, to account for such settings, we measure the expected referenceless quality loss
\begin{displaymath}
   \mathbb{E}_{y\sim \Py, \, \hat{x}\sim \Phatxy} \bigl[ L_{\operatorname{qual}}(\hat{x}) \bigr] \approx \frac{1}{n} \sum_{i=1}^n L_{\operatorname{qual}} \bigl(\hat{x}^{(i)} \bigr) \enspace , 
\end{displaymath}
where $y^{(i)}$ are samples from $\Py$,  $\hat{x}^{(i)}$ are samples from $\Phatxyi$, and $L_{\operatorname{qual}}$ is a quality assessment metric. For image data, we set $L_{\operatorname{qual}}$ to be either the NIQE score~\citep{mittal2012making} or the negative of the blurriness index introduced by~\citet{blur}, for which in both cases a lower score indicates a better image quality.

The last performance metric focuses on the reconstruction quality of certain categorical attributes. 
Given a classifier $f_{\operatorname{att}}$ of these attributes, we define the expected attribute reconstruction loss 
\begin{displaymath}
     \mathbb{E}_{x, y\sim \Pxy, \, \hat{x}\sim \Phatxy} \bigl[ L_{\operatorname{att}}(x, \hat{x}; f_{\operatorname{att}}) \bigr] \approx 
    \frac{1}{n} \sum_{i=1}^n L_{\operatorname{att}}\bigl( x^{(i)}, \hat{x}^{(i)}; f_{\operatorname{att}}\bigr) \enspace , 
\end{displaymath}
where $L_{\operatorname{att}}$ is an attribute-related loss. One choice for $L_{\operatorname{att}}$ is the 
binary attribute prediction loss, i.e., 
\begin{equation} \label{eq:att-0-1}
    L_{\operatorname{att}}^{\text{0--1}}\bigl(x, \hat{x}; f_{\operatorname{att}}\bigr) := \begin{cases}
        1 \enspace , \enspace &\text{if } f_{\operatorname{att}}(x) \neq f_{\operatorname{att}}(\hat{x}) \\ 
        0 \enspace, \enspace &\text{if } f_{\operatorname{att}}(x) = f_{\operatorname{att}}(\hat{x})
    \end{cases} \enspace . 
\end{equation}
To relax the hard 0--1 penalty in the binary prediction loss, we also compute the cosine similarity of the latent representations of $x$ and $\hat{x}$, that is,  
\begin{equation}
    \label{eq:att-cos}
    L_{\operatorname{att}}^{\operatorname{cos}}\bigl(x, \hat{x}; f_{\operatorname{att}}\bigr) := \frac{ \langle l_x, l_{\hat{x}}  \rangle }{\Vert l_x \Vert \cdot \Vert l_{\hat{x}} \Vert} \enspace, 
\end{equation}
where $l_x$ and $l_{\hat{x}}$ denote the last activations\footnote{Note that by incorporating activations, we implicitly assume $f_{\operatorname{att}}$ to be a neural network classifier.} of $f_{\operatorname{att}}(x)$ and $f_{\operatorname{att}}(\hat{x})$, respectively. 
While attribute reconstruction can be related to LPIPS, using an attribute-specific classifier allows for a more precise evaluation with respect to the labeled attributes. 

\subsection{Fairness and Diversity} \label{subsec:fairness_diversity}
There are no unified metrics for assessing the amount of fairness in unconditional models. For instance, \citet{tanjim2022debiasing} use the expected attribute reconstruction loss as a heuristic for fairness, while~\citet{jalal2021fairness} evaluate fairness by visual inspection of confusion matrices. 
Therefore, to formalize the amount of fairness in a principled way, we extend the definitions of RDP~\eqref{eq:rdp} and PR~\eqref{eq:pr} through divergence measures between probability distributions.   
Let $\Prdp$ be the distribution quantifying the probability that the conditional sample $\hat{x}$ and the original sample $x$ belong to the class $c$. That is, for a classifier $f_{\operatorname{att}}$ that predicts the class  $c \in \{c_1,\dots, c_k\}$ of $x$, we define 
\begin{eqnarray}
  \Prdp(c=c_j) &:=& \frac{\myP (\hat{x} \in C_j \vert \, x \in C_j)}{\sum_{l=1}^k \myP (\hat{x} \in C_l \vert \, x \in C_l)} \nonumber\\ &\approx& 
    \frac{ \frac{1}{n_j} \sum_{i \in I_j}  L_{\operatorname{att}}^{\text{0--1}}\bigl(x^{(i)}, \hat{x}^{(i)}; f_{\operatorname{att}}\bigr)}{
    \sum_{l=1}^k \frac{1}{n_l} \sum_{i \in I_l}  L_{\operatorname{att}}^{\text{0--1}}\bigl(x^{(i)}, \hat{x}^{(i)}; f_{\operatorname{att}}\bigr)
    }
    \enspace \forall j \in [k] \enspace, \label{eq:Prdp}\nonumber\\ 
\end{eqnarray} 
where $C_l$ is the set of samples corresponding to class $c_l$, $I_l:= \{ i \in [n]: \; x^{(i)} \in C_l\}$,  and $n_l=\vert I_l\vert$. 
According to~\eqref{eq:rdp}, an algorithm satisfies RDP if and only if $\Prdp(c=c_j)=\Prdp(c=c_i)$ for all $i,j\in [k]$, or equivalently, if and only if $\Prdp=\mathcal{U}([k])$, i.e., if $\Prdp$ is uniformly distributed over $[k]$.
Hence, to quantify the amount of violation of fairness, we introduce 
\begin{equation}
    \label{eq:rdp_div}
    \Delta_{\operatorname{RDP}} := D\bigl(\Prdp\, \Vert \, \mathcal{U}([k]) \bigr) \enspace , 
\end{equation}
where $D$ is a divergence measure. 
We set $D$ to the Pearson \\$\chi^2$-divergence or the Chebyshev-distance to obtain 
\begin{align}
    \Delta_{\operatorname{RDP-}\chi^2} :&= \sum_{j=1}^k
    \frac{(\Prdp(c=c_j) - 1/k)^2}{1/k} \notag \\ 
    &= k \sum_{j=1}^k \biggl(\Prdp(c=c_j) - \frac{1}{k}\biggr)^2 \enspace , \label{eq:delta_rdpchi} \\ 
    \Delta_{\operatorname{RDP-Cheb}} :&= \max_{j\in[k]} \; \biggl\vert \, \Prdp(c=c_j) - \frac{1}{k} \biggr\vert  \enspace .  \label{eq:delta_rdpcheb} 
\end{align}
By empirically estimating~\eqref{eq:Prdp}, we obtain plug-in estimates of $\Delta_{\operatorname{RDP-}\chi^2}$ and $\Delta_{\operatorname{RDP-Cheb}}$. 
One could think of other divergences, such as KL-divergence or Total-Variation-divergence but we decided on the $\chi^2$-divergence to relate the scores to the test statistics of a Pearson's $\chi^2$-test. This test allows us to statistically test the hypothesis $\Prdp=\mathcal{U}([k])$. 
Furthermore, we propose to consider the Chebyshev distance, since it can be interpreted as the maximum violation of fairness. 

However, note that considering RDP in isolation might be insufficient for fairness assessment as demonstrated by the following example. Let the conditional generative model be such that $\myP(\hat{x} \in C_j \vert \, x \in C_j)=0.5$ for all $j\in[k]$, and therefore $\Prdp(c=c_j)= 1/k$ , thus satisfying RDP.
Let us further assume that all misclassified samples belong to $C_1$, i.e., $\hat{x} \in \{C_j, C_1\}$ if $x \in C_j\neq C_1$. It follows that $\myP(\hat{x} \in C_1 \vert \, x \in C_j)= 1 - \myP(\hat{x} \in C_j \vert \, x \in C_j) = 0.5$ for all $j\neq 1$ and consequently, it holds that 
\begin{displaymath}
    \myP (\hat{x} \in C_1) = \sum_{j=1}^k \myP (\hat{x} \in C_1 \vert \, x \in C_j) \,  \myP (x \in C_j) = \sum_{j=1}^k 0.5 \cdot  \frac{1}{k} = 0.5 \enspace, 
\end{displaymath}
under the assumption that $\myP (x \in C_j)=1/k$ for all $j\in$ [\textit{k}]. 
Hence, even though the conditional generative model is fair with respect to RDP, it has an unreasonable bias for producing samples from $C_1$ more frequently than samples from $C_j\neq C_1$ if $k>2$. 
We provide further simplified synthetic examples highlighting the differences between RDP and PR in Example~\ref{exa} of the Appendix.
This motivates us to complement RDP with quantitative measures of PR. As above in~\eqref{eq:Prdp}, we define a discrete distribution $\Ppr$ over classes $\{c_1, \dots, c_k\}$ by 
\begin{displaymath}
    \Ppr(c=c_j):=  \myP (\hat{x} \in C_j) \approx \frac{1}{n} \sum_{i=1}^n  
    \bm{1}
    \bigl[ \hat{x}^{(i)} \in C_j \bigr] \enspace \forall j\in [k] \enspace ,
\end{displaymath}
where $\bm{1}$ is the indicator function. In practice, we use the attribute classifier $f_{\operatorname{att}}$ to quantify whether $\hat{x}^{(i)} \in C_j$.  
Once again, PR is satisfied if and only if $\Ppr=\mathcal{U}([k])$, which motivates to introduce 
$\Delta_{\operatorname{PR-}\chi^2}$ and $\Delta_{\operatorname{PR-Cheb}}$ analogous to their RDP-versions in~\eqref{eq:delta_rdpchi} and \eqref{eq:delta_rdpcheb}, respectively. 

\label{subsec:metrics:diversity}
A different viewpoint of fairness is that of diversity in the presence of conditional arguments that are uninformative for certain attributes. 
For illustration, let us consider the low-resolution sample for image upsampling depicted in Figure~\ref{fig:white_avg}. 
This sample does not contain any information regarding its race, and hence, a diverse upsampling algorithm should produce high-resolution samples without favoring any race over the other. 
We quantify that intuition by building upon CPR~\eqref{eq:cpr} and defining $\Pcprcondy$ by 
\begin{equation*} %
    \Pcprcondy( c=c_j):=  \myP ( \hat{x} \in C_j \vert \, y) \approx \frac{1}{n_{x}} \sum_{i=1}^{n_{x}}
    \bm{1} \bigl[  \hat{x}^{(i)} \in C_j \bigr]  
    \enspace \forall j\in[k] \enspace 
\end{equation*} 
for $n_{x}$ conditional samples $\hat{x}^{(i)}$ based on a fixed condition $y$. %
Certainly, we are not interested in the diversity of $\Pcprcondy$ for any $y$ but only for non-informative conditions $y$.  Hence, we define the \emph{uninformative conditional proportional representation} (UCPR) distribution $\Pucpr$ by taking the expectation of $\Pcprcondy$ over all uninformative $y$, that is, 
\begin{eqnarray} 
    \Pucpr( c=c_j)&:=& \mathbb{E}_{y \sim U} \bigl[  \Pcprcondy( c=c_j)\bigr]   \nonumber\\ &\approx& \frac{1}{n_{y}} \sum_{m=1}^{n_{y}} \frac{1}{n_{x}} \sum_{i=1}^{n_{x}}
    \bm{1} \bigl[  \hat{x}^{(i)}\bigl(y^{(m)} \bigr) \in C_j \bigr] \enspace \forall j\in[k] \enspace , \nonumber\\
\end{eqnarray} 
where $U$ is a random variable that generates uninformative $y$, $y^{(1)}, \dots , y^{(n_{y})}$ are $n_{y}$ samples from $U$, and $ \hat{x}^{(i)} (y^{(m)})$ denotes the $i$th output of the conditional generative model given the condition $y^{(m)}$. 
We say that the conditional generative model satisfies UCPR if and only if $\Pucpr( c=c_j) = \Pucpr( c=c_i)$ for all $i,j \in [k]$.
As above, we define the plug-in estimators $\Delta_{\operatorname{UCPR-}\chi^2}, \,\Delta_{\operatorname{UCPR-Cheb}}$ to measure the diversity of the conditional model in the presence of uninformative inputs.
Generally, the generation of conditions that are uninformative regarding the classes $\mathcal{C}$ depends on the application. In the following section, we explain how we obtain uninformative samples in the specific case of image upsampling. 

Conveniently, since the introduced fairness metrics are evaluations of a divergence (see e.g.~\eqref{eq:rdp_div}), they satisfy $D(\mathbb{P} \, \Vert \, \mathbb{Q})=0$ if and only if $\mathbb{P}=\mathbb{Q}$. Consequently, we conclude that RDP, PR, and UCPR of a conditional generative model can be verified using our proposed metrics. 
\begin{corollary}
    A conditional generative model satisfies 
    \begin{enumerate}
        \item RDP if and only if $\Delta_{\operatorname{RDP}-\chi^2}=\Delta_{\operatorname{RDP-Cheb}}=0$; 
        \item PR if and only if $\Delta_{\operatorname{PR}-\chi^2}=\Delta_{\operatorname{PR-Cheb}}=0$;
        \item UCPR if and only if $\Delta_{\operatorname{UCPR}-\chi^2}=\Delta_{\operatorname{UCPR-Cheb}}=0
        $.
    \end{enumerate}
\end{corollary}

\section{Introducing \dataset{}}\label{sec:data}

In the further course of this work, we apply the framework derived in Section~\ref{sec:framework} to evaluate the performance, fairness, and diversity of a particular type of conditional generative model, namely for the application of image upsampling on human face images. 
Specifically, we empirically analyze the upsampling performance, but also evaluate the fairness and diversity of the reconstructions using race labels.
Unfortunately, standard face datasets used for generative models---such as CelebA~\citep{liu2015faceattributes} and FFHQ~\citep{karras2019style}---do not come with race labels. On the other hand, datasets that come with race labels like PPB~\citep{buolamwini2018gender} and UTKFace~\citep{zhang2017age} are often too small to be used to train a (conditional) generative model. 
Suited for our experiments are the BUPT~\citep{bupt} and FairFace~\citep{karkkainen2021fairface} datasets, which both provide sufficient examples to train a generative model and include race labels. However, the BUPT database is not consistent in the shape and size of the images and contains only celebrities. Therefore, we decided to base our experiments on the FairFace dataset, which consists of around 87k training samples of resolution $224\times 224$. The samples are labeled by race (7~categories), which are approximately uniformly distributed across the races, see Figure~\ref{fig:fairface_prop}. 

While it is often desirable to train on a balanced dataset, having access to such a dataset is rare in practice. Instead, existing public face datasets are usually constructed from online sources, which are biased towards faces that have a light skin tone~\citep{merler2019diversity}. In this study, we aim to investigate the effect of such a dataset bias on image upsampling algorithms. 
Hence, to mimic the bias apparent in large-scale datasets while having access to race labels, we subsample from the FairFace dataset such that it imitates the racial distribution of CelebA\footnote{The approximate racial distribution is taken from Figure~1 in \citet{karkkainen2021fairface}.} (see Figure~\ref{fig:unfairface_prop}). The resulting dataset, which we refer to as \textit{\dataset}, consists of around 20k samples of which more than $80\%$ are labeled as ``White''.

For testing, we take a subset of the test set of FairFace consisting of $200$ samples per race, totaling a test-set size of $1400$ images. 
We base our expected attribute reconstruction loss on race reconstruction, i.e., $f_{\operatorname{att}}$ in~\eqref{eq:att-0-1} and~\eqref{eq:att-cos} is a race classifier. 
Note, that we do not have access to an unambiguous, ground-truth race classifier; 
instead, we leverage the pretrained race classifier provided by~\citet{karkkainen2021fairface}. To diminish the effect of the imperfect classifier on the evaluations, we select only the first $200$ samples per race that are correctly labeled by the classifier. 
To emphasize the fact that in the following we focus on race reconstruction, we denote the losses in~\eqref{eq:att-0-1} and~\eqref{eq:att-cos} by $L_{\operatorname{race}}^{\text{0-1}}$ and $L_{\operatorname{race}}^{\text{cos}}$, respectively and henceforth use the notation $\hr, \, \hathr,\, \lr$ introduced in Section~\ref{sec:upsampling_methods}.

\section{Experiments}\label{sec:exp}
\begin{figure*}[t]
    \centering
     \subfigure[Models 
     trained on \dataset.]{\label{fig:black_unfair}\includegraphics[width=\textwidth]{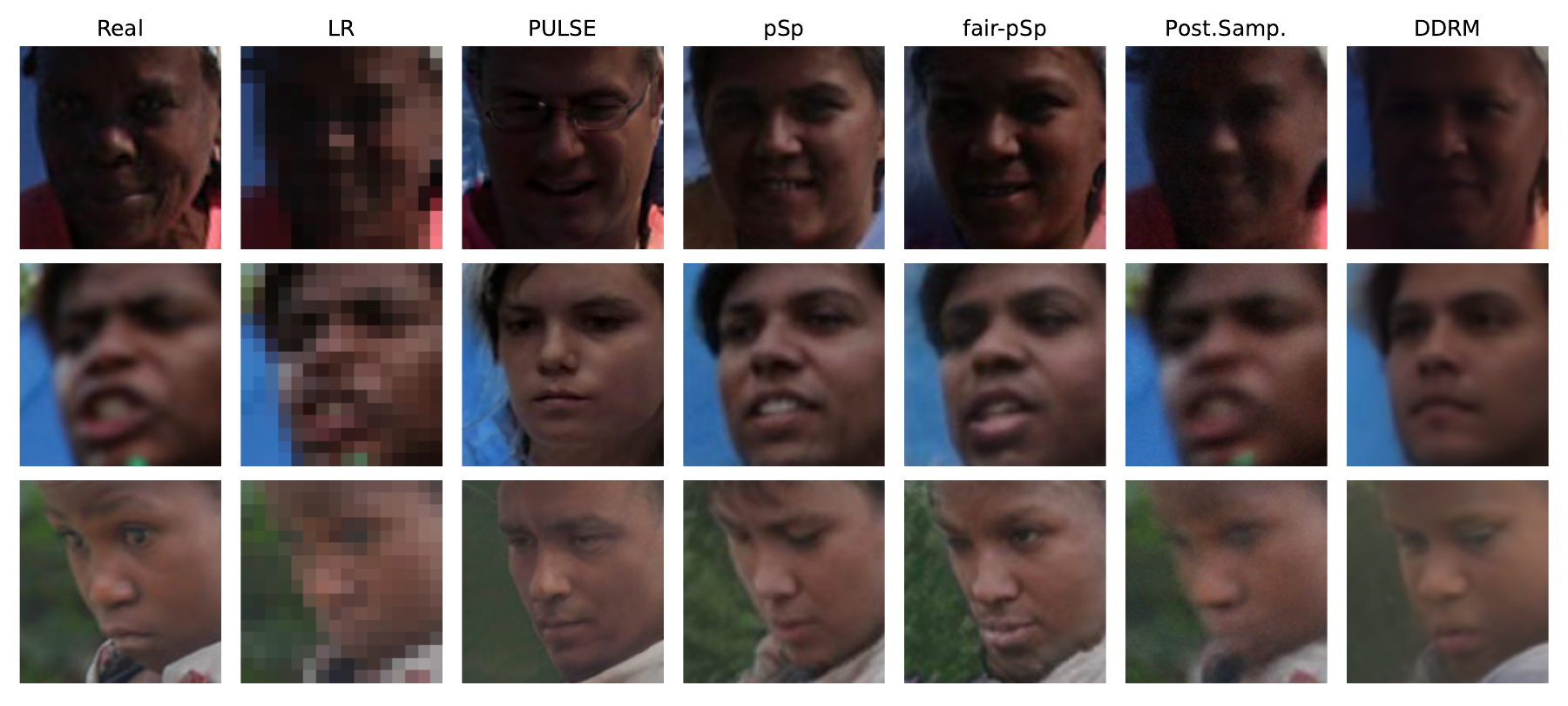}}
  \subfigure[Models trained on FairFace.]{\label{fig:black_fair}\includegraphics[width=\textwidth]{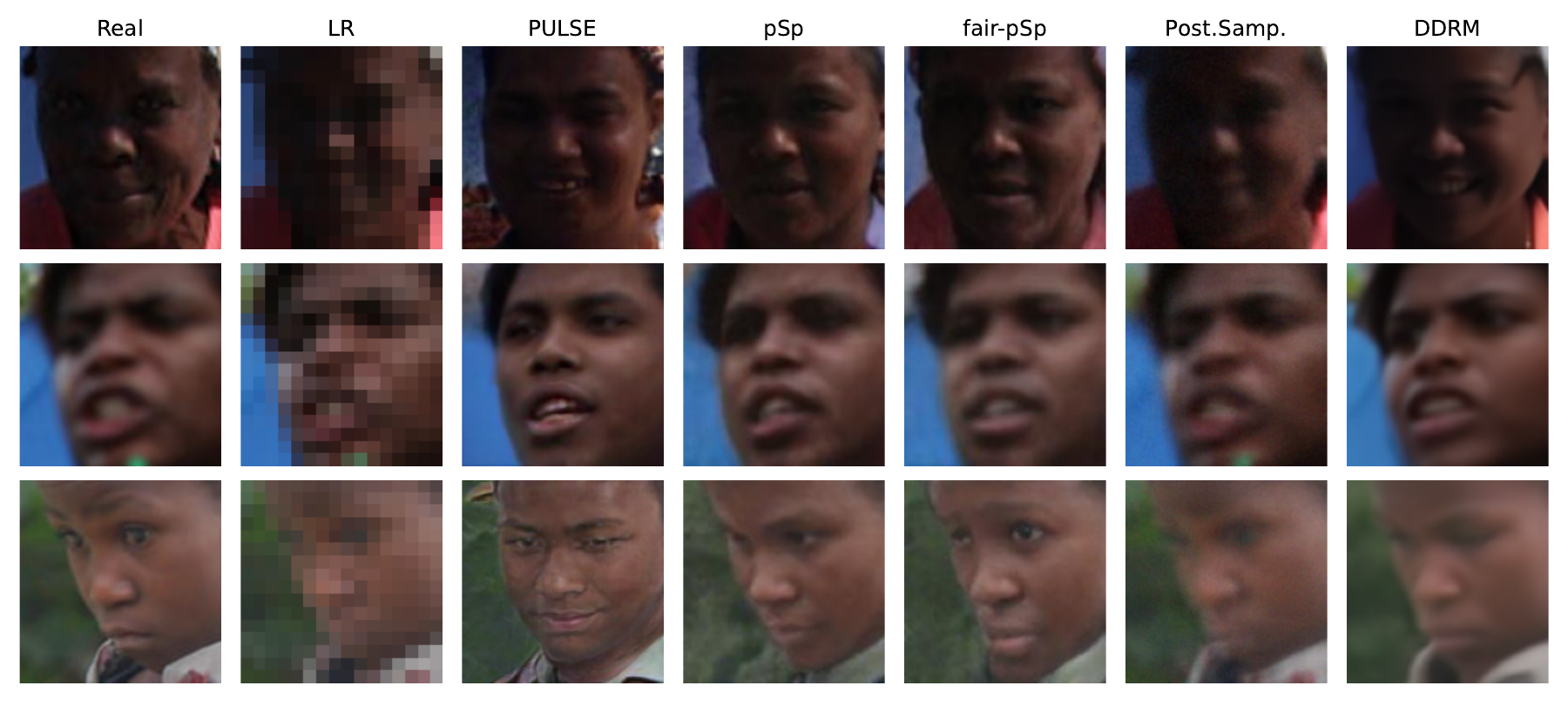}}
  \caption[]{Upsampling results for models trained on \dataset{} and FairFace using test samples categorized as ``Black''.}
  \label{fig:black}
  \Description{Reconstructions based on several image upsampling algorithms if the input is classified as ``Black''.}
\end{figure*}
In this section, we apply our framework derived in Section~\ref{sec:framework} for the evaluation of image upsampling algorithms. 
To assess the bias of training on an unbalanced population with multiple underrepresented races, we compare upsampling methods trained on the UnfairFace dataset to the same methods trained on the original FairFace dataset with balanced subpopulations. All models are evaluated on the same holdout set with balanced races. The experiments can be reproduced using the \href{https://github.com/MikeLasz/Benchmarking-Fairness-ImageUpsampling}{official code repository}.

\paragraph{Experimental setup}
Many architectures, such as StyleGAN~\citep{karras2019style, karras2020analyzing, karras2020training, karras2021alias} are designed for outputs of resolution $2^k \times 2^k$ for some $k\in \mathbb{N}$. Since FairFace contains images of resolution $224\times 224$, we trained all image upsampling models to generate outputs of the next-smallest power of two, which is $128\times 128$. 
For testing, we first downsample each test sample to resolution $128 \times 128$ using bilinear downscaling with antialiasing to obtain $\hr$. To get $\lr$, we downsample $\hr$ using the same downscaling. 
For the experiments evaluating performance and fairness, we set $\lr$ to resolution $16 \times 16$. For evaluating the diversity, we employ a two-step procedure to produce uninformative samples. First, we average over all test samples of the same race. Second, we downsample to a resolution of $4\times 4$ to obtain uninformative samples. In our experiments, we use $7$ uninformative conditions, for which we generate $100$ reconstructions each, resulting in a total of $700$ samples. Note, that we first average to wipe out sample-specific biases apparent in a single image. 
In \Cref{subsec:supplement:model-descriptions}, we describe the individual models and specify the hyperparameters used for training and evaluation, respectively.

\subsection{Qualitative Results}
We start with the qualitative evaluation of upsampling methods, which provides a first intuition and valuable insights into the strengths and weaknesses of the individual methods. 

For underrepresented subpopulations, we observe that the upsampling results show faces of a considerably lighter skin tone compared to the original images if the models are trained on \dataset{}; for example, in \Cref{fig:black}(a), we show the upsampling results for samples categorized as ``Black''. The racial bias appears especially pronounced for models that are not designed for fair image upsampling, i.e., PULSE, pSp, and DDRM.
While the diffusion-based models exhibit a less severe bias, the reconstructions appear more blurry. Notably, the blurriness appears to be reduced when reconstructing samples categorized as ``White'' (see Figure~\ref{fig:black}(b) in the Appendix). As a reference, in Figure~\ref{fig:black}(b), we present the results for the same methods trained on the original FairFace dataset with balanced subpopulations.
Further examples provided in \Cref{fig:teaser} and in~\Cref{subsec:supplement:additional-qualitative-results} in the Appendix demonstrate that the models fail to reconstruct particular phenotypes related to ethnicity, such as bindis and headscarves. 
When trained on FairFace, the reconstructions match the original samples better across all races. 
Unsurprisingly, all methods faithfully reconstruct faces categorized as ``White''  if trained on \dataset{} (see \Cref{fig:white_app} in the Appendix).
In summary, the qualitative results highlight the potential biases in image upsampling and motivate a thorough quantitative analysis of the performance and fairness of different methods.

\subsection{Upsampling Performance}
\begin{table*}
  \caption[]{Performance metrics for each algorithm trained on \dataset{} (UFF) and FairFace (FF). Values are highlighted in bold if the Null hypothesis that the results on the UFF and FF datasets coincide is not rejected. Lower scores indicate a better performance.}
  \label{tab:performance}
    \begin{tabular}{lrrcrrcrrcrrcrrcrr}
\toprule
    & \multicolumn{2}{c}{LPIPS} && \multicolumn{2}{c}{DSSIM} &&\multicolumn{2}{c}{$L_{\operatorname{race}}^{\operatorname{cos}}$} && \multicolumn{2}{c}{$L_{\operatorname{race}}^{\text{0-1}}$} &&\multicolumn{2}{c}{NIQE} && \multicolumn{2}{c}{BLUR} \\ 
    \cline{2-3} \cline{5-6} \cline{8-9} \cline{11-12} \cline{14-15} \cline{17-18}
    & UFF & FF && UFF & FF && UFF & FF && UFF & FF && UFF & FF  && UFF & FF 
    \\ 
\midrule
PULSE & \textbf{0.24} & \textbf{0.24} && \textbf{0.21} & \textbf{0.21} && 0.25 & 0.22 && 0.65 & 0.59 && 3.87 & 3.80 && -2.85 & -3.07 \\
pSp & 0.18 & 0.16 && 0.18 & 0.16 && 0.16 & 0.11 && 0.56 & 0.37 && 4.76 & 4.73 && -0.60 & -0.62 \\
fair-pSp & 0.17 & 0.16 && 0.17 & 0.16 && 0.14 & 0.11 && 0.46 & 0.30 && 4.62 & 4.74 && -0.62 & -0.61 \\
Post.Samp. & 0.27 & 0.29 && 0.16 & 0.16 && 0.09 & 0.08 && \textbf{0.33} & \textbf{0.31} && 5.84 & 5.45 && -0.89 & -1.08 \\
DDRM & 0.23 & 0.20 && 0.16 & 0.12 && 0.23 & 0.12 && 0.68 & 0.36 && 6.38 & 6.78 && -0.24 & -0.24 \\
\bottomrule
\end{tabular}
\end{table*}
\begin{figure}[htbp]
  \centering
  \subfigure[Models trained on \dataset.]{\label{fig:0-1_per_eth-unfairface}\includegraphics[width=0.48\textwidth]{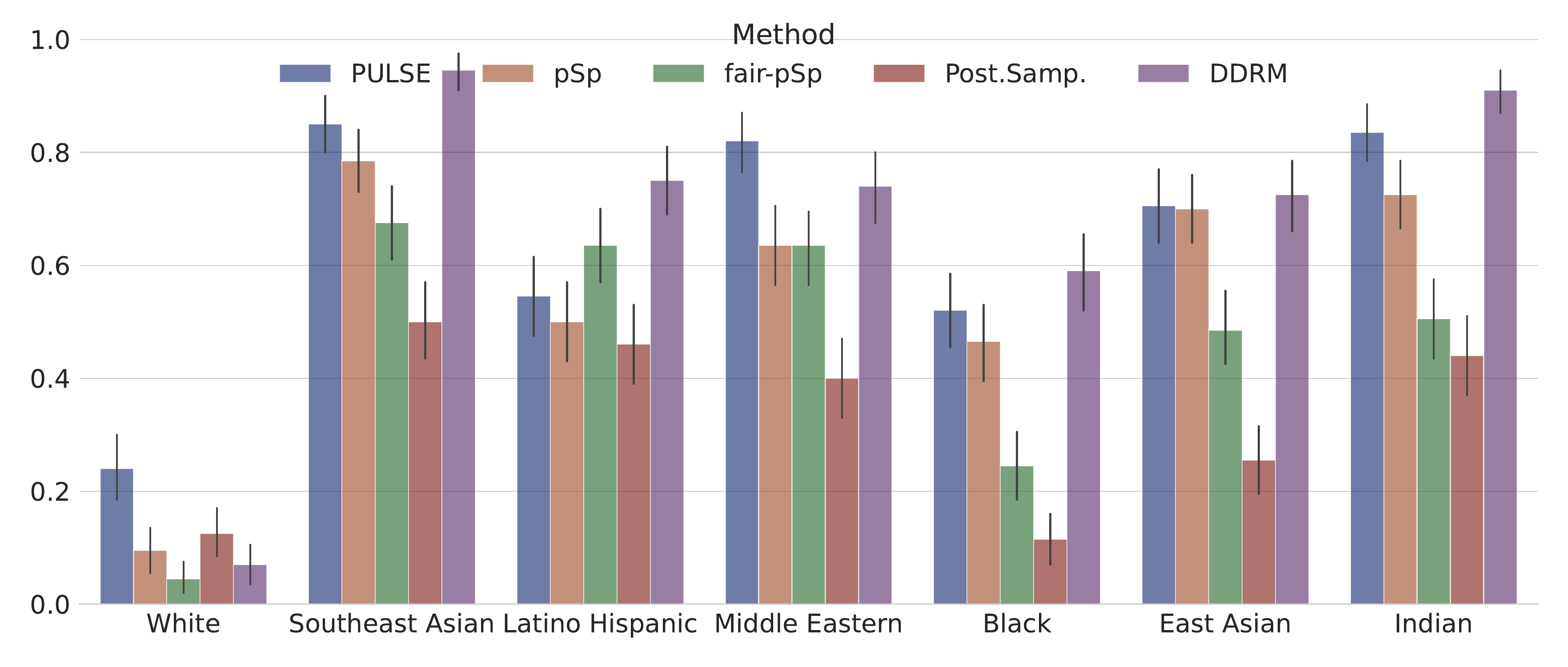}}
  \hfill%
  \subfigure[Models trained on FairFace.]{\label{fig:0-1_per_eth-fairface}\includegraphics[width=0.48\textwidth]{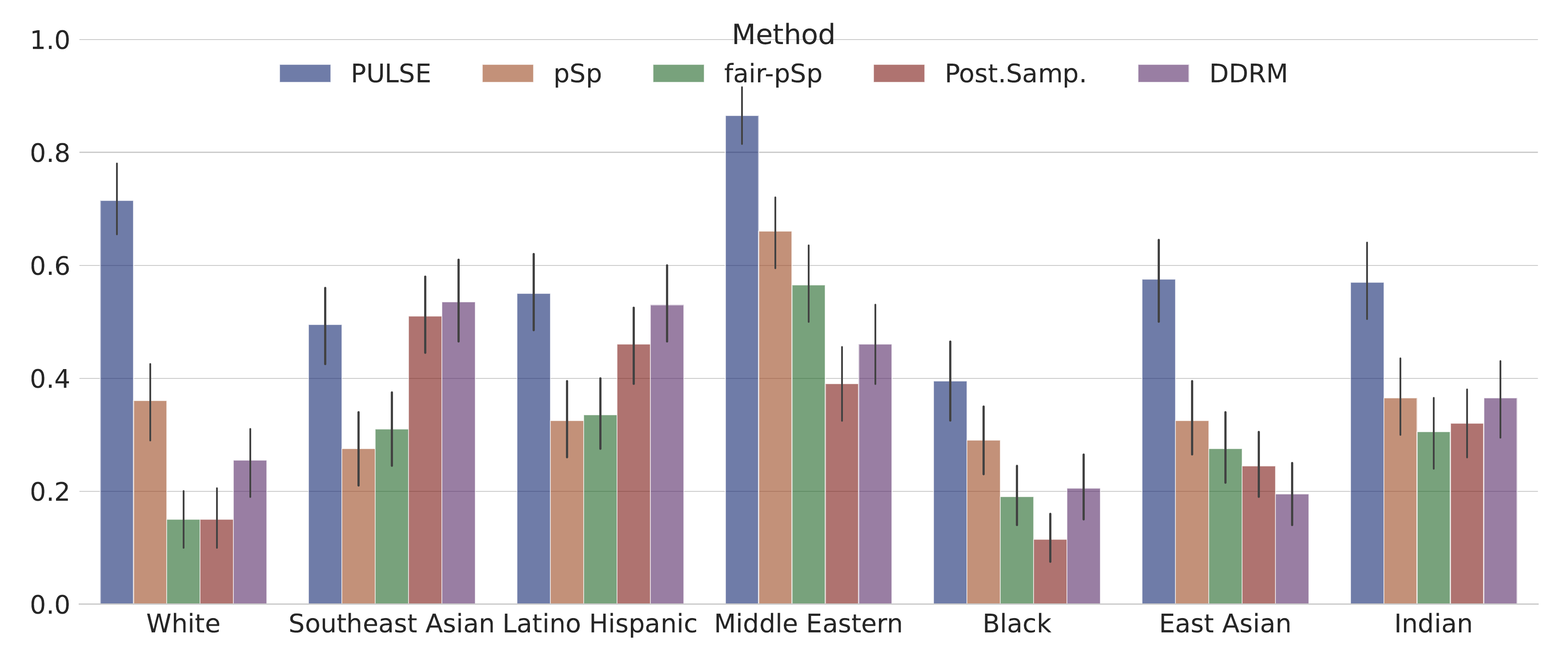}}
  \caption[]{Comparing the race reconstruction loss $L_{\operatorname{race}}^{\text{0-1}}(\hr, \hathr)$ if $\hr \in C_j$ for varying $C_j$. Lower scores indicate a better reconstruction. }
  \label{fig:0-1_per_eth}
  \Description{Two plots that compare the race reconstruction accuracy loss for inputs of varying races for each upsampling algorithm. The left plot shows the results of the algorithms trained on UnfairFace and the right plot shows the results of the algorithms trained on FairFace.}
\end{figure}
Next, we present the quantitative evaluation of the upsampling performance based on the metrics described in \Cref{subsec:metrics:upsampling-performance}. In \Cref{tab:performance}, we compare the results across all considered methods trained on the \dataset{} and FairFace datasets, respectively. Based on these results, we perform a Wilcoxon signed-rank test~\citep{wilcoxon}
to test the Null hypothesis that the distributions of metrics derived from a model trained on \dataset{} and FairFace coincide.%
\footnote{
    We first run an Anderson-Darling test~\citep{anderson} to test for normality of the differences in the losses between the UFF and FF models. We find that only NIQE is close to a normal distribution, which prevents us from using a paired t-test to test whether the means coincide.
   Since $L_{\operatorname{race}}^{\text{0-1}}$ is a binary value, we employ a Pearson's $\chi^2$-test in that case.
}
In bold, we indicate if the null hypothesis cannot be rejected based on a significance level of $\alpha=0.05$.
In most cases, we see a significant difference between models trained on the \dataset{} and FairFace datasets. In the case of PULSE, we notice that it is less affected by the racial distribution, as it exhibits more similar results for both datasets. For DDRM, we observe a notable improvement in the race reconstruction loss when trained on FairFace. 
Additionally, in \Cref{subsec:supplement:additional-quantitative-results}, we report the p-values of the Wilcoxon signed-rank test to assess how significant the difference in performance is for each method. 

Overall, we find that StyleGAN-based %
models perform better in terms of image quality.
We observe that almost all models are sensitive to the training data bias when it comes to race reconstruction loss. Posterior Sampling constitutes the only exception, which indicates that it has a smaller racial bias when trained on unbalanced data.

\subsection{Fairness and Diversity}\label{subsec:eval_fairness}
\begin{table*}[t]
    \centering
    \caption[]{Evaluating the fairness discrepancy $D(\Prdp \, \Vert \,  \mathcal{U}([k]))$ and $D(\Ppr \, \Vert \,  \mathcal{U}([k]))$ for different divergences $D$
    for each algorithm trained on \dataset{} (UFF) and FairFace (FF). Lower scores indicate more fairness.
    The \xmark{} illustrates that the Null hypothesis $\Prdp=\mathcal{U}([k])$ or $\Ppr=\mathcal{U}([k])$ is rejected.
    }
    \label{tab:fairness}
    \begin{tabular}{lrrcrrcrrcrrcrrcrr}
    \toprule
    & \multicolumn{2}{c}{$\Delta_{\operatorname{RDP-}\chi^2}$} && \multicolumn{2}{c}{$\Delta_{\operatorname{RDP-Cheb}} 
$} && \multicolumn{2}{c}{$\Prdp=\mathcal{U}([k])$}&& 
\multicolumn{2}{c}{$\Delta_{\operatorname{PR-}\chi^2}$} && \multicolumn{2}{c}{$\Delta_{\operatorname{PR-Cheb}} 
$} && \multicolumn{2}{c}{$\Ppr=\mathcal{U}([k])$} \\  
\cline{2-3} \cline{5-6} \cline{8-9} \cline{11-12} \cline{14-15} \cline{17-18} 
    & UFF & FF && UFF & FF && UFF & FF && UFF & FF && UFF & FF && UFF & FF \\ 
    \midrule
    PULSE & 0.34 & 0.12 && 0.16 & 0.10&& \xmark & \xmark && 0.86 & 0.21 && 0.29 & 0.11 && \xmark & \xmark\\

    pSp & 0.24 & 0.04 && 0.15 & 0.07 && \xmark & \xmark && 0.71 & 0.13 && 0.24 & 0.10 &&\xmark & \xmark  \\
    fair-pSp & 0.16 & 0.03 && 0.11 & 0.05 && \xmark & \xmark&& 0.53 & 0.07 && 0.24 & 0.06 && \xmark & \xmark\\
    Post.Samp. & 0.05 & 0.04 && 0.05 & 0.04 && \xmark & \xmark&& 0.11 & 0.07 && 0.10 & 0.08 && \xmark & \xmark\\
    DDRM & 0.70 & 0.05 && 0.27 & 0.04&& \xmark & \xmark && 1.49 & 0.04 && 0.41 & 0.06&& \xmark & \xmark \\
    \bottomrule
    \end{tabular}
\end{table*}
Before evaluating the fairness metrics introduced in Section~\ref{subsec:metrics:diversity}, let us first break down parts of the evaluation from Table~\ref{tab:performance} and compare how the performance varies across the races. 

In Figure~\ref{fig:0-1_per_eth}, we visualize $L_{\operatorname{race}}^{\text{0-1}}(\hr, \hathr)$ conditioned on $\hr\in C_j$, i.e., it shows a rescaled version of $\Prdp$, which we provide in Figure~\ref{fig:rdp} of the Appendix. Remember, we quantified RDP (discrepancy) by the divergence between the scaled race-conditioned performance and the uniform distribution (see \Cref{eq:rdp_div}). The results in Figure~\ref{fig:0-1_per_eth-unfairface} illustrate that all models trained on \dataset{} have a comparably low race reconstruction loss for images labeled as ``White''. For all other races, we observe high reconstruction losses, especially for PULSE and DDRM. Interestingly, we find that ``Southeast Asian'' and ``Indian'', which represent the least frequent races in \dataset{} (see Figure~\ref{fig:unfairface_prop}), have the largest reconstruction loss. 
When trained on FairFace, we observe that the reconstruction losses approach uniformity. As before, we find that DDRM is most sensitive with respect to the training data. 
A similar plot that visualizes $\Ppr$ is provided in Figure~\ref{fig:pr} in \Cref{subsec:supplement:additional-quantitative-results} of the Appendix. 
In \Cref{tab:fairness}, we summarize these results by evaluating the divergences proposed in~\Cref{subsec:metrics:diversity}. 
If the models are trained on \dataset{}, we find that Posterior Sampling achieves the highest degree of fairness. 
Generally, training on FairFace has a large influence on the fairness of the models; all fairness scores improve. In that setting, DDRM, Posterior Sampling, and fair-pSp obtain approximately similar scores in all metrics, while PULSE performs worst in all metrics. 
To test whether $\Prdp=\mathcal{U}([k])$ and $\Ppr=\mathcal{U}([k])$, we run a Pearson's $\chi^2$-test with significance $\alpha=0.05$. 
We find that the Null hypothesis is rejected for all models, even when they are trained on FairFace. 
This means that even though some methods are more fair than others, the statistical evidence suggests that $\Prdp \neq \mathcal{U}([k])$ and $\Ppr \neq \mathcal{U}([k])$, i.e., no method can be considered fair. 
Once again, we want to highlight that the fairness of DDRM spikes from the worst score when trained on \dataset{} to the best when trained on FairFace. 

\begin{figure*}[t]
    \centering
     \subfigure[Models 
     trained on \dataset.]{\label{fig:unfairface_avg}\includegraphics[width=\textwidth]{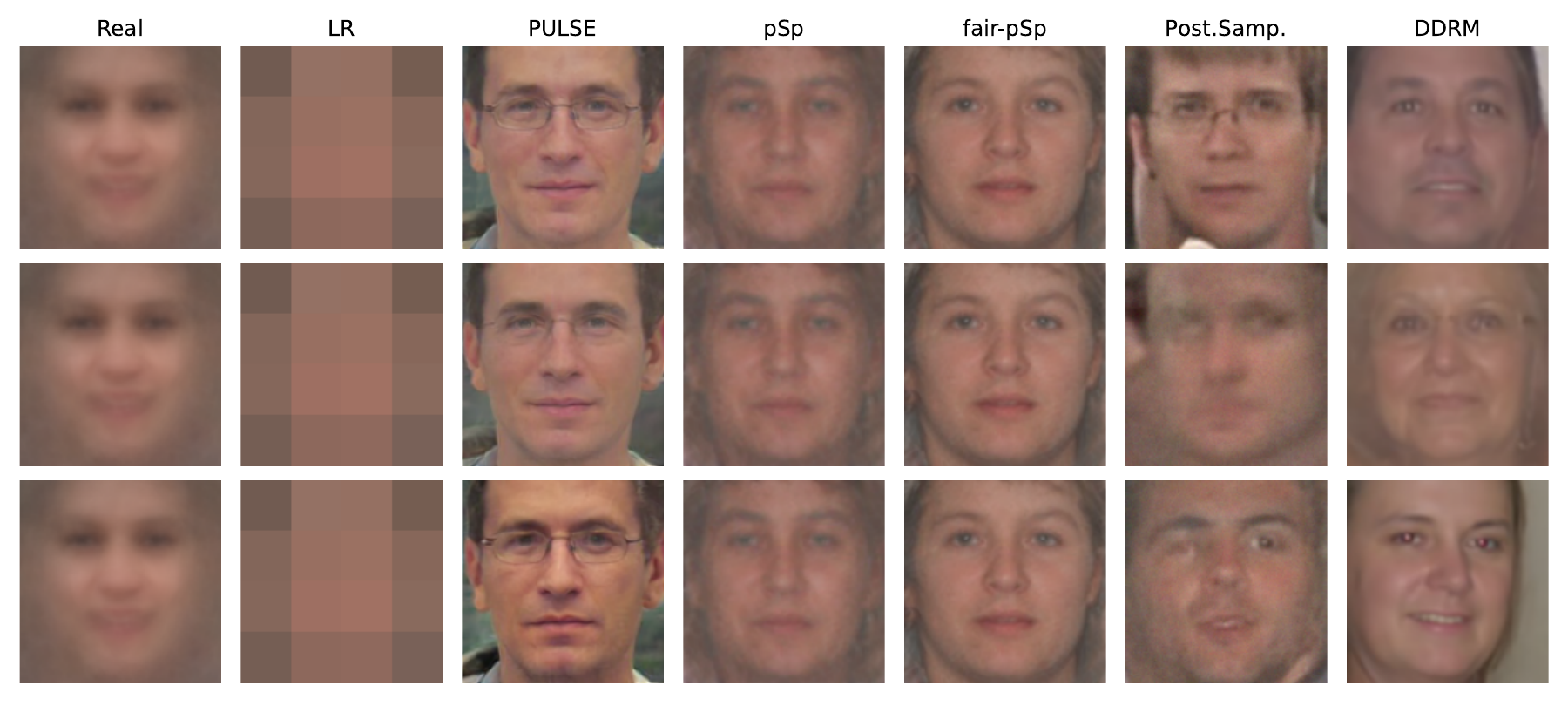}}
  \subfigure[Models trained on FairFace.]{\label{fig:fairface_avg}\includegraphics[width=\textwidth]{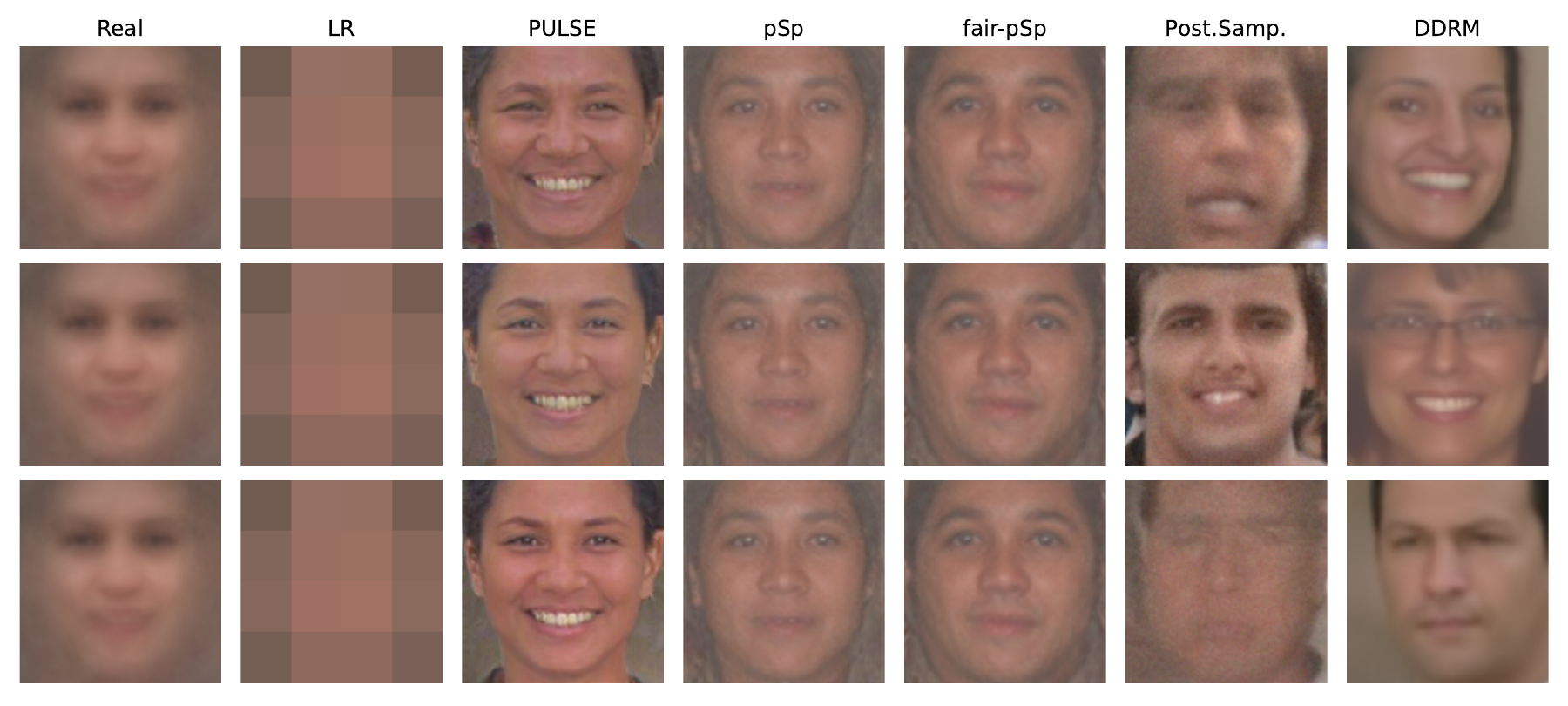}}
  \caption[]{Upsampling results for models trained on \dataset{} and FairFace using uninformative test samples. The real image is an average over images classified as ``White''.}
  \Description{Reconstructions based on several image upsampling algorithms if the input is an uninformative $4\times 4$ image. The input is an average over images classified as ``White''.}
  \label{fig:white_avg}
\end{figure*}

\begin{figure}[htbp]
  \centering
  \subfigure[Models trained on \dataset.]{\label{fig:ucpr-unfairface}\includegraphics[width=0.48\textwidth]{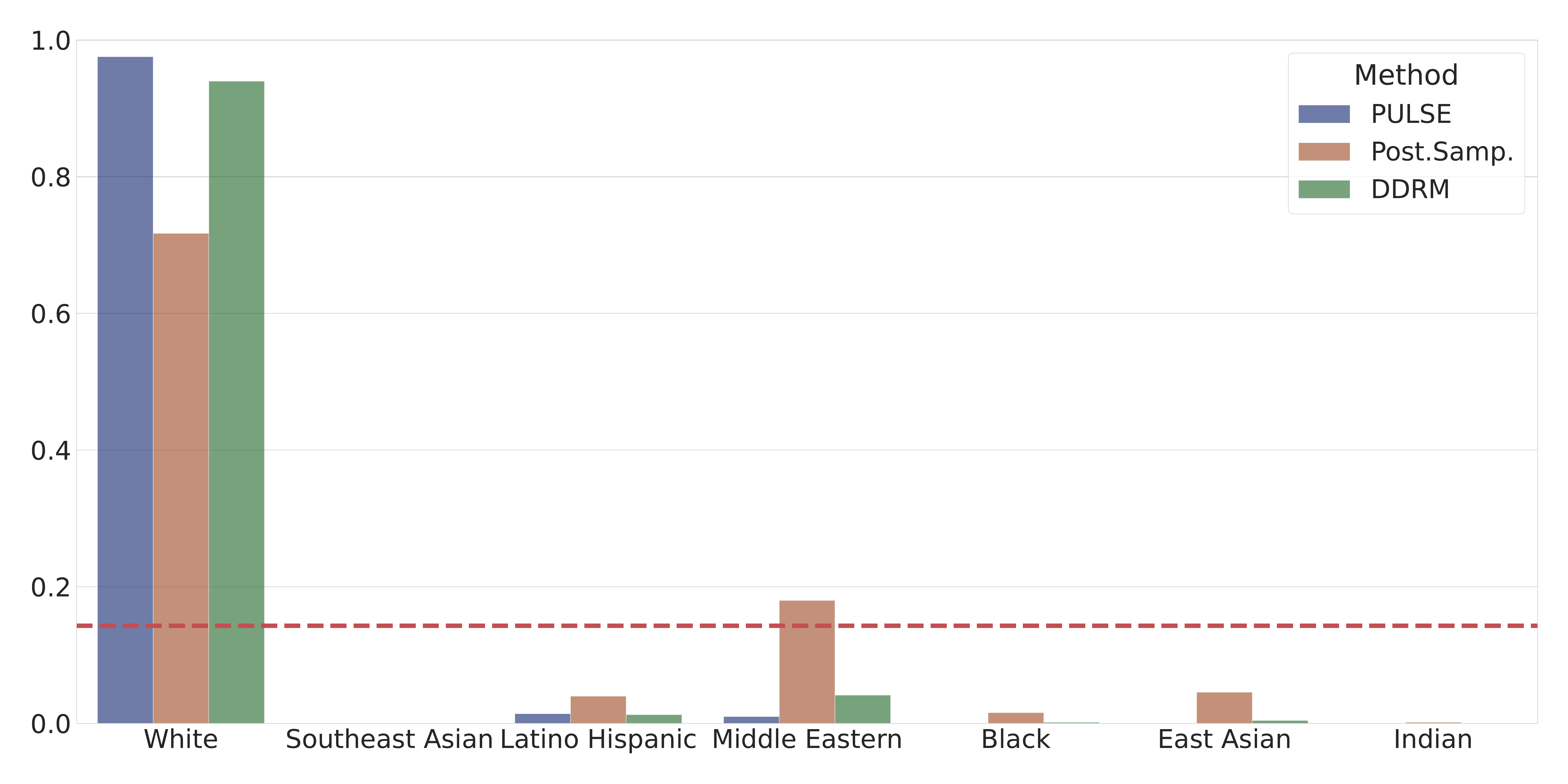}}
  \hfill %
  \subfigure[Models trained on FairFace.]{\label{fig:ucpr-fairface}\includegraphics[width=0.48\textwidth]{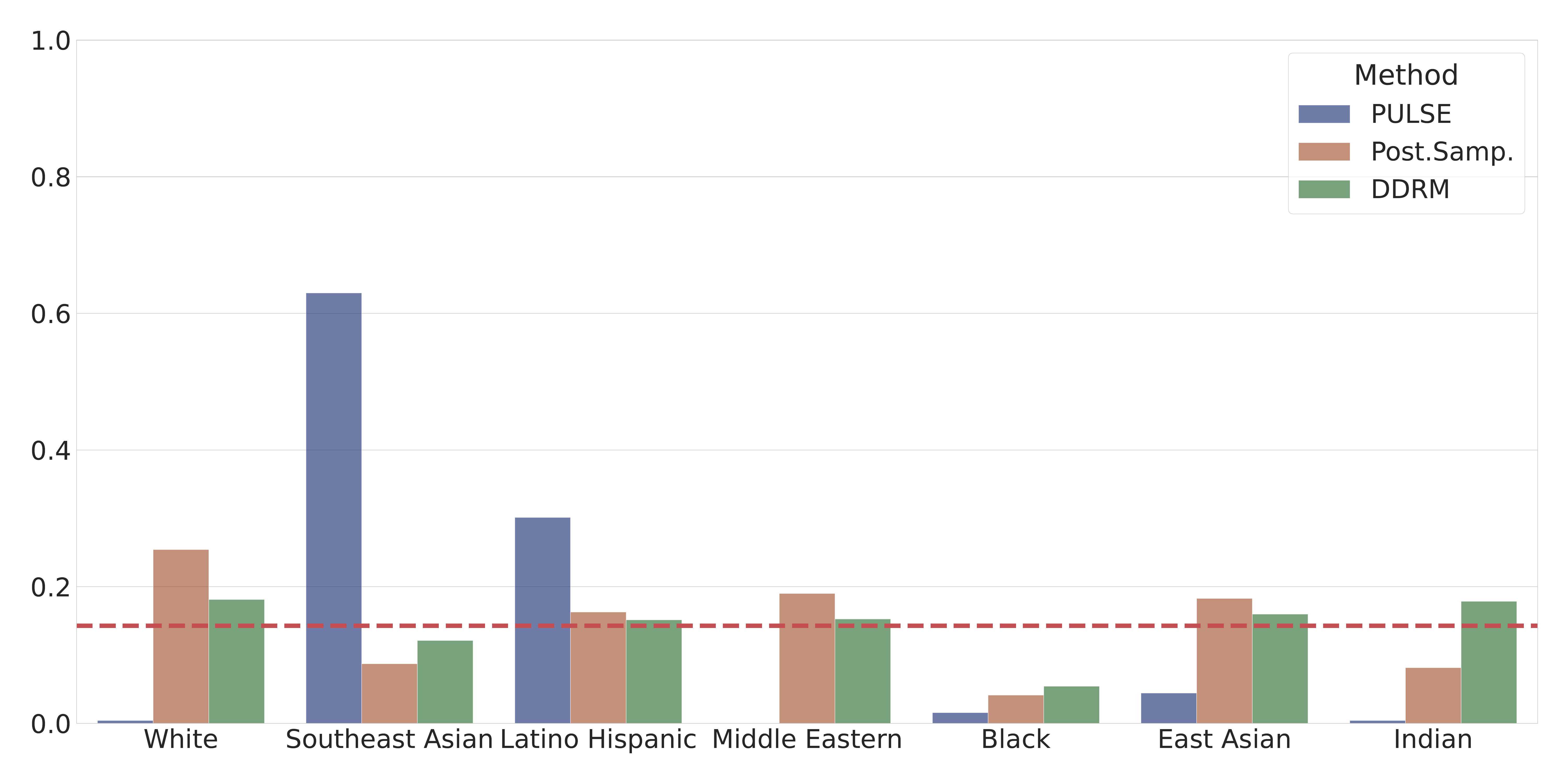}}
  \caption[]{Comparing the uninformative conditional proportional representation distribution $\Pucpr$ of models trained on \dataset{} and FairFace. The horizontal dashed line indicates the bar height corresponding to a uniform distribution.}
  \Description{Two plots that show the uninformative conditional proportional representation distribution for each upsampling algorithm. The left plot shows the results of the algorithms trained on UnfairFace and the right plot shows the results of the algorithms trained on FairFace.}
  \label{fig:ucpr}
\end{figure}

To measure the diversity of the methods, we propose to upsample uninformative samples multiple times (see~\Cref{fig:white_avg}). 
Qualitatively, we observe a clear bias towards ``White'' reconstructions in PULSE and DDRM.\footnote{Further reconstructions can be found in \Cref{subsec:supplement:additional-qualitative-results} in the Appendix.} 
This bias becomes even more evident when examining $\Pucpr$ in Figure~\ref{fig:ucpr}. When trained on \dataset{}, almost all reconstructions are labeled as ``White''. When trained on FairFace, Posterior Sampling, and DDRM 
generate samples such that the resulting racial distribution is close to uniformity for all races but ``Black'', which is still highly underrepresented. Surprisingly, PULSE generates no faces labeled as ``White'' anymore but almost exclusively generates samples labeled as ''Southeast Asian'' and ``Latino Hispanic''. Therefore, PULSE is subject to a strong racial bias even when trained on FairFace. 
We provide the quantitative evaluation based on $\Delta_{\operatorname{UCPR}-\chi^2 }$ and $\Delta_{\operatorname{UCPR-Cheb}}$ in \Cref{tab:diversity}. Note, that since the generation process in pSp and fair-pSp is deterministic,\footnote{In Section~\ref{subsec:supplement:additional-quantitative-results}, we provide an additional experiment on the diversity, which leverages randomly perturbed $4\times 4$ inputs. This setup allows to evaluate pSp and fair-pSp with respect to their diversity.} we cannot generate different values given the same input. Therefore, these models do not apply to our analysis. 
Unsurprisingly, we see that training on FairFace benefits the diversity of the output for all methods. DDRM provides the most diverse outputs when trained on FairFace. We run a similar Pearson's $\chi^2$-test and find that no method provides significant diversity in the sense that $\Pucpr=\mathcal{U}([k])$.

\begin{table}[]
    \centering
    \caption[]{
    Evaluating the diversity discrepancy $D(\Pucpr  \,  \Vert \, \mathcal{U}([k]) )$ for different divergences $D$
    for each algorithm trained on \dataset{} (UFF) and FairFace (FF). Lower scores indicate more diversity. The \xmark{} illustrates that the Null hypothesis $\Pucpr=\mathcal{U}([k])$ is rejected.}
    \label{tab:diversity}
    \begin{tabular}{lrrcrrcrr}
\toprule
& \multicolumn{2}{c}{$\Delta_{\operatorname{UCPR-}\chi^2}$} && \multicolumn{2}{c}{$\Delta_{\operatorname{UCPR-Cheb}} 
$} && \multicolumn{2}{c}{$\Pucpr = \mathcal{U}([k])$} \\  
\cline{2-3} \cline{5-6} \cline{8-9} \
    & UFF & FF && UFF & FF && UFF & FF  \\ 
\midrule
PULSE & 5.67 & 2.43 && 0.83 & 0.49 && \xmark & \xmark \\
Post.Samp. & 2.85 & 0.24 && 0.57 & 0.11&& \xmark & \xmark \\
DDRM & 5.20 & 0.08 && 0.80 & 0.09 && \xmark & \xmark \\
\bottomrule
\end{tabular}
\end{table}

\subsection{Summary} 
Our qualitative results reveal a discernible racial bias when models are trained on \dataset{}. This holds for $16\times 16$ inputs and uninformative $4\times 4$ inputs. The effect is most prominent in PULSE, while being least apparent in Posterior Sampling. 
Yet, this comes with a drawback, as Posterior Sampling tends to generate blurry images.
The quantitative measurements of the performance metrics highlight that it is significantly influenced by the dataset choice, particularly pronounced in DDRM. Additionally, the fairness performance of DDRM is very volatile to the training dataset, which should be taken into consideration for practical applications. When trained on FairFace, fair-pSp, Posterior Sampling, and DDRM achieve comparable levels of fairness and diversity. But still, statistical tests reject the hypothesis that RDP, PR, and UCPR are satisfied. Finally, it is worth highlighting the advantages of fair-pSp over pSp. Stratification, contrastive learning, and regularizing with a race classifier could, in principle, also be used to improve other techniques regarding their fairness. 

\section{Limitations and Future Work}
    \paragraph{$128\times 128$-Resolution}While our provided framework (\Cref{sec:framework}) is generally applicable, our evaluation of upsampling methods considers $128\times 128$-resolution images. For comparison, PULSE~\citep{menon2020pulse} was evaluated on CelebA-HQ~\citep{karras2018progressive} images of resolution up to $1024 \times 1024$. However, we also want to emphasize that this choice is driven by the fact that large-scale datasets, such as CelebA-HQ and FFHQ, do not come with fairness-relevant labels. Therefore, we believe that the research community would highly benefit from a high-quality dataset with labels comparable to those in FairFace, which we consider as one valuable future research direction.
    \paragraph{Fairness-relevant Labels} There is a general concern in choosing the fairness-relevant labels. For instance, the exact partitioning and its granularity can lead to different results~\citep{race_multidimensional}.
    Additionally, the boundaries between races can be intricate, given that phenotypic features might vary within races~\citep{race_multidimensional2}.  
    As an alternative to race labels, we experimented with ITA-based skin tone estimations~\citep{kinyanjui2019estimating}. However, we found skin tone estimates to be inconsistent and unreliable in FairFace (see e.g. Figure~3 in~\citet{karkkainen2021fairface}). From a technical perspective, the labels should be easy and reliable to predict, since the empirical evaluation is based on predicted labels. 
    For reference, the FairFace race classifier \citep{karkkainen2021fairface} used in our experiments has an average prediction accuracy of around $71\%$ on FairFace. To diminish the effect of the imperfect classifier, we select only real test samples that are correctly classified. Although a favored approach would involve utilizing a more powerful label classifier. 
    \paragraph{Model Parameters} We found that upsampling models pretrained on FFHQ and CelebA do not yield good reconstructions on FairFace samples. This is due to differently shaped image crops and camera angles apparent in FairFace, which is in contrast to the clean images in CelebA. 
    This is why we must retrain all models on FairFace and \dataset{}. Our hyperparameter choice (\Cref{subsec:supplement:model-descriptions}) was guided by the settings reported in the respective works, which might be suboptimal in this setting. 
    All utilized models can be downloaded from our \href{https://github.com/MikeLasz/Benchmarking-Fairness-ImageUpsampling}{open-source repository}.

\section{Conclusion}
This work advances toward a principled assessment of %
conditional generative models. 
Our contribution lies in the formulation of a comprehensive evaluation framework that extends beyond conventional performance criteria as it incorporates novel fairness and diversity metrics.
To validate our framework, we introduce \dataset{}, a dataset derived from the original FairFace dataset, and perform an in-depth evaluation of the fairness of image upsampling methods. 
Our empirical analysis underscores the sensitivity of the fairness of all considered methods to the underlying training data. Furthermore, our findings reveal that, while some models exhibit a closer alignment with fairness ideals, no model achieves statistically significant fairness. This outcome emphasizes the pressing need for further research in this domain.
In light of these insights, we encourage researchers and practitioners in this field to embrace our proposed framework in the evaluation of conditional generative models beyond image upsampling methods.

\newpage 
\section{Ethical Considerations}\label{sec:ethical}
Various definitions of race and ethnicity have been developed over time and a unified definition remains elusive.  
In previous works, both terms are often used interchangeably~\citep{grosfoguel2004race, hunt2008ambiguous}.

We adopted the terminology employed by~\citet{balestra2018diversity} and the Canadian Institute for Health Information~\citep{canadian2022guidance} to describe ``ethnicity'' as a community belonging to a common cultural group, and ``race'' as a group defined by similarities of physical phenotypes. 
It is important to note that contemporary scientific understanding supports the view that there is no biological basis for a categorization into distinct racial groups~\citep{fujimura2008introduction, race_multidimensional, race_multidimensional2}.
We are aware of the controversies surrounding these definitions, specifically that they are imprecise and do not capture the full heterogeneity of human societies and cultures.
For instance,~\citet{schaefer2008encyclopedia} describes race as an interplay between physical similarities, social similarities, and self-identification as such.
Emphasizing the socially constructed nature of the terms ``race'' and ``ethnicity'', we recognize their definitions and connotations as subject to variation across time and context.

In our specific example of image upsampling, the evaluation is restricted to visible phenotypes expressed in images with labels adopted from previous work~\citep{karkkainen2021fairface}.
Strictly speaking, the adopted categorization violates the nuanced meaning of race and ethnicity, which are multifaceted concepts that might vary over time, even for a single individual. For instance, an Asian immigrant can be of American ethnicity; 
however, a single image cannot express the underlying social and cultural characteristics in their full diversity, complexity, and variability. 
Consequently, assessing whether a reconstruction accurately represents a sample categorized as a specific race becomes a highly debatable endeavor. 
In fact, even the evaluation of reconstruction performance of specific phenotypes, such as a headscarf, may prove insufficient, as ethnicity can never be reduced to sole phenotypes. Thus, being aware of the limitations of the adopted categorization, in this work, we merely use it as a proxy to quantify the potential biases of upsampling methods in a relevant context to highlight the potential issues in real-world applications. We encourage future work to remain aware of these ethical considerations and advocate for developing a more nuanced evaluation methodology.

\begin{acks}
We thank Ri\v cards Marcinkevi\v cs for encouraging us to enhance contemporary evaluation procedures, which eventually resulted in this work, and Mohamed Hebiri for his insightful comments. 
This work was supported by the Deutsche Forschungsgemeinschaft (DFG, German Research Foundation) under Germany’s Excellence Strategy – EXC 2092 CASA – 390781972.
\end{acks}
\bibliographystyle{ACM-Reference-Format}
\bibliography{literatur_cr}

\newpage 
\section{Appendix}
\subsection{Further Intuitions and Recommendations for Practitioners}
\label{subsec:further_intuition}
The proposed fairness metrics in Section~\ref{subsec:fairness_diversity} are based on the notion of RDP, which is satisfied if 
\begin{equation} \label{eq:app_rdp}
    \myP(\hat{x} \in C_i \vert \, x \in C_i) = \myP (\hat{x} \in C_j\vert \, x \in C_j) \enspace   \forall i,j \in [k], \, \forall x \in \mathcal{X}
\end{equation} 
and PR, which is satisfied if 
\begin{equation} \label{eq:app_pr}
    \myP(\hat{x} \in C_i) = \myP (x  \in C_i) \enspace  \forall i\in [k] \enspace 
\end{equation}
for a partition $\mathcal{C}=\{C_1, \dots, C_k\}$ of the dataset $\mathcal{X}$. 
In this section, we aim at sheding more light on these definitions and providing the reader some more intuition. 

Note that RDP~\eqref{eq:app_rdp} focuses on equalizing the class error rates of the reconstructions $\hat{x}$ irrespective the actual class of $x$. Intuitively speaking, the conditional generative model should not perform better at correctly reconstructing one class over another. 
In contrast, PR aims at retaining the true class distribution. 
Specifically, in the case $\myP (x  \in C_i)= \myP (x  \in C_j)$ for all $i, j \in [k]$, PR enforces a balance of the overall class distribution of the reconstructions, which has a direct relation to fairness. Therefore, PR does not relate to correct class predictions and hence captures different goals than RDP.

In certain applications, we may favor one fairness notion over another illustrated by the following examples.
For simplicity, let us reconsider the specific task of image upsampling.
For the sake of illustration, let us assume that we use image upsampling for lossy data decompression, that is, the low-resolution images constitutes the compressed image and the decompression is conducted my applying an upsampling algorithm.\footnote{This is just a hypothetical non-practical usecase that exchanges data storage with computing power.}
Since data compression typically targets retaining the semantics of the image---and therefore the predicted class---equalizing the error rates is crucial.
This goal is well-captured by enforcing RDP. 

In contrast, if image upsampling is used for generating a novel synthetic dataset under coarse guidance provided by low-resolution images, we may not necessarily prioritize reconstructing the exact classes. Instead, we may focus on generating an unbiased synthetic dataset in which each class is equally represented. This is captured by PR. 
Ultimately, we recommend practitioners to be aware of which facets of fairness are covered by each definition. It is rarely recommended to discard one definition completely. 
Also, while numerical scores can only tell us whether fairness is violated, figures akin to Figure~\ref{fig:0-1_per_eth} and Figure~\ref{fig:pr} allow tracing down the underrepresented class causing the violation. Hence, we recommend practicioners to also analyzing these figures in their fairness assessment. 

Finally, we present two extreme cases, illustrating that RDP does not induce PR and vice versa. 
\begin{example}\label{exa}
    Consider the image upsampling scenario tackled in Section~\ref{sec:exp} with just three races, ``White'', ``Black'', and ``Asian''. 
    Mathematically, we define 
    \begin{align*}
        C_1&:=\{ x \in \mathcal{X}: \; x \text{ is ``White''}\} \enspace , \\ 
        C_2&:=\{ x \in \mathcal{X}: \; x \text{ is ``Black''}\} \enspace , \\ 
        C_3&:=\{ x \in \mathcal{X}: \; x \text{ is ``Asian''}\} \enspace . 
    \end{align*}
    Let us assume that the conditional class distributions $\myP(\hat{x} \in C_j \vert \, x \in C_i)$ are given by Figure~\ref{fig:cond_class_exa1}.  %
    This means that, for instance, if the low-resolution input corresponds to a real image showing a "Black" person, then the reconstruction is ``White''/``Black'' with probability 50\%/50\% (Figure~\ref{fig:cond_class_exa1_black}). RDP aims at balancing the striped bars in Figure~\ref{fig:cond_class_exa1}, which is satisfied in this concrete example. 
    However, under the assumption that $\myP(x\in C_1) =\myP(x\in C_2) =\myP(x\in C_3)$, the resulting class distribution of the reconstructions $\hat{x}$ is given by 
    \begin{equation*}
        \myP( \hat{x} \in C_1) = \sum_{i=1}^3 \myP(\hat{x} \in C_1 \vert \, x \in C_i) \myP( x\in C_i) = 0.5 
    \end{equation*}
    whereas 
    \begin{equation*} 
 \myP( \hat{x} \in C_2) = \myP( \hat{x} \in C_3) = 0.25 \enspace . 
    \end{equation*}    
    Especially, since $ \myP( \hat{x} \in C_1)> \myP( \hat{x} \in C_j)$ for $j\in \{2, 3\}$, i.e., ``White'' is overrepresented, we observe that PR is violated. 

    Another extreme case arises when the conditional class distributions 
    $\myP(\hat{x} \in C_j \vert \, x \in C_i)$ are as depicted in Figure~\ref{fig:cond_class_exa2}.
    ``White'' faces are 100\% correctly classified (Figure~\ref{fig:cond_class_exa2_white}), whereas ``Black'' and ``Asian'' have 0\% correct class reconstructions (Figure~\ref{fig:cond_class_exa2_black} and Figure~\ref{fig:cond_class_exa2_asian}). Hence, RDP is violated but PR is satisfied under the assumption $\myP(x\in C_1) =\myP(x\in C_2) =\myP(x\in C_3)$: 
    \begin{equation*}
         \myP( \hat{x} \in C_j) = \sum_{i=1}^3 \myP(\hat{x} \in C_j \vert \, x \in C_i) \myP( x\in C_i) = \myP( x \in C_j) \enspace . 
    \end{equation*}
  
    These examples highlight that the utilized fairness definitions are no one-size-fits-all solutions. Instead, PR and RDP capture different aspects of fairness.
    \begin{figure}[htbp]
      \centering
      \subfigure[$\myP(\hat{x} \in C_j \vert \, x \in C_1)$, that is, the low-resolution input corresponds to ``White''.]{
      \label{fig:cond_class_exa1_white}
      \includegraphics[width=0.4\textwidth]{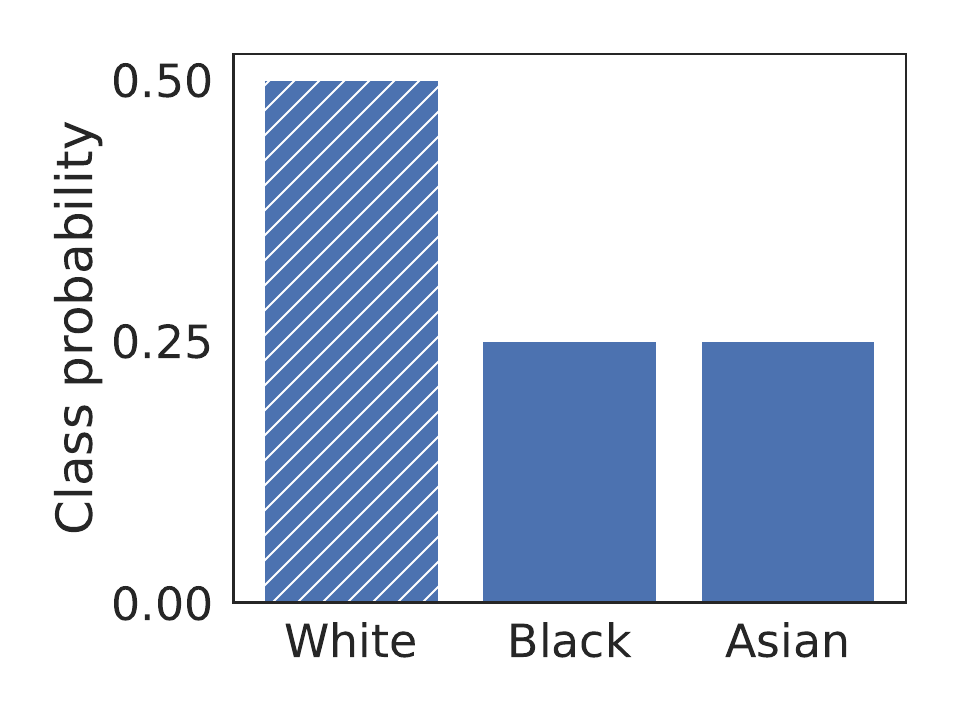}}
      \hfill 
      \subfigure[$\myP(\hat{x} \in C_j \vert \, x \in C_2)$, that is, the low-resolution input corresponds to ``Black''.]
      {
     \label{fig:cond_class_exa1_black}
      \includegraphics[width=0.4\textwidth]{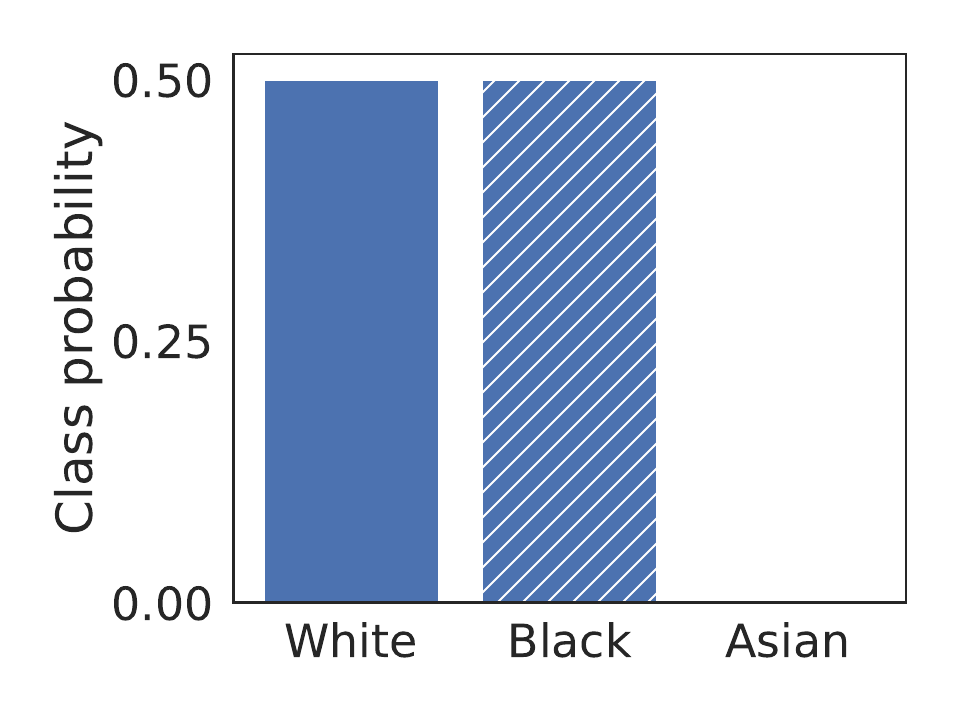}}
      \hfill 
      \subfigure[$\myP(\hat{x} \in C_j \vert \, x \in C_3)$, that is, the low-resolution input corresponds to ``Asian''.]
      {\label{fig:cond_class_exa1_asian}
      \includegraphics[width=0.4\textwidth]{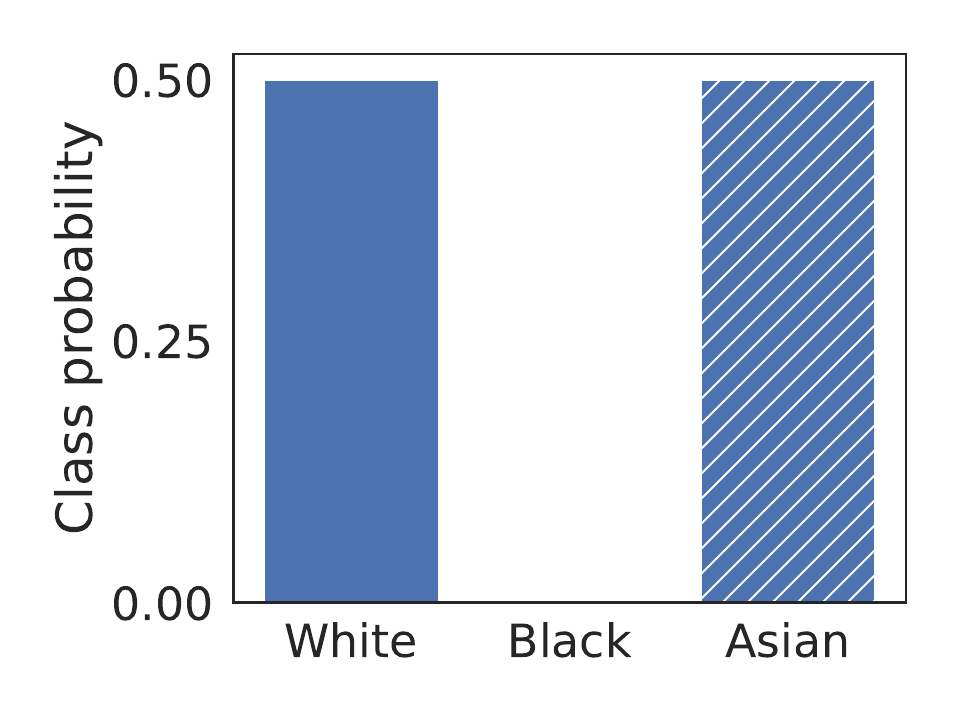}}
    
     \caption[]{Given a low-resolution input (``White'', ``Black'', or ``Asian''),  we assume that the class distributions of the reconstructions are given by the following probability mass functions.}
      \label{fig:cond_class_exa1}
      \Description{A plot that visualizes the conditional class distributions corresponding to the first extreme case in Example~9.1.}
    \end{figure}

    \begin{figure}[htbp]
      \centering
      \subfigure[$\myP(\hat{x} \in C_j \vert \, x \in C_1)$, that is, the low-resolution input corresponds to ``White''.]{
      \label{fig:cond_class_exa2_white}
      \includegraphics[width=0.4\textwidth]{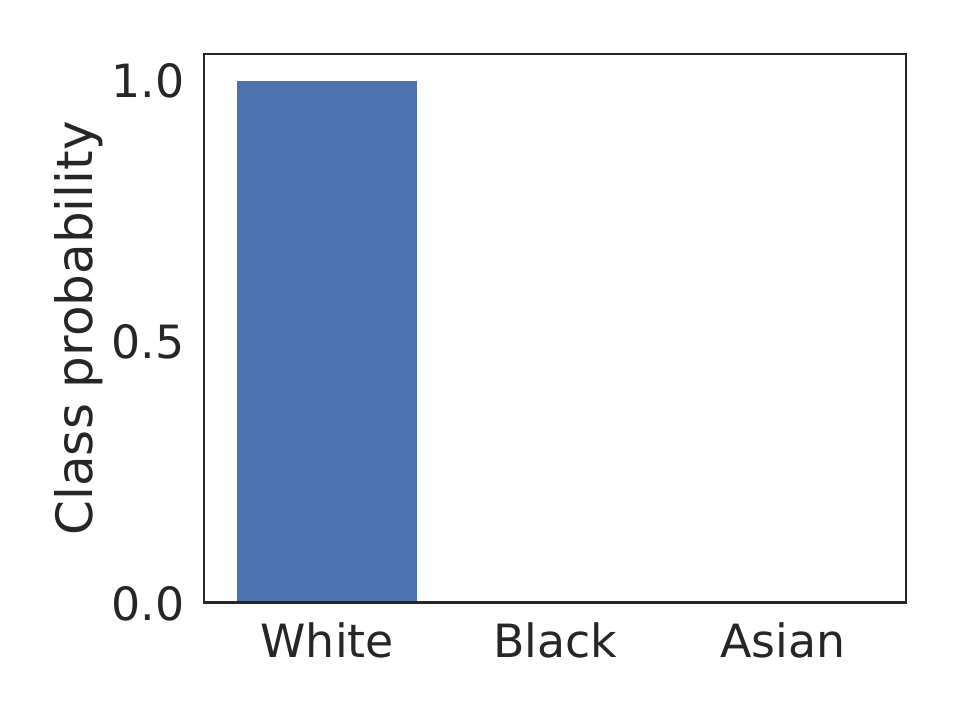}}
      \hfill 
      \subfigure[$\myP(\hat{x} \in C_j \vert \, x \in C_1)$, that is, the low-resolution input corresponds to ``Black''.]
      {
      \label{fig:cond_class_exa2_black}
      \includegraphics[width=0.4\textwidth]{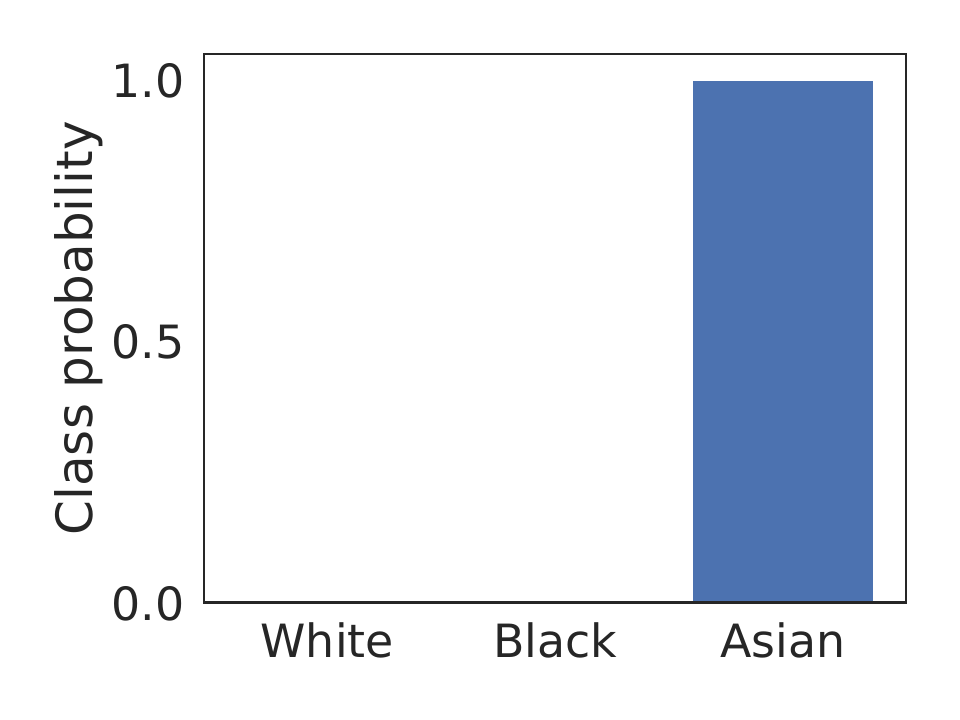}}
      \hfill 
      \subfigure[$\myP(\hat{x} \in C_j \vert \, x \in C_1)$, that is, the low-resolution input corresponds to ``Asian''.]{\label{fig:cond_class_exa2_asian}
      \includegraphics[width=0.4\textwidth]{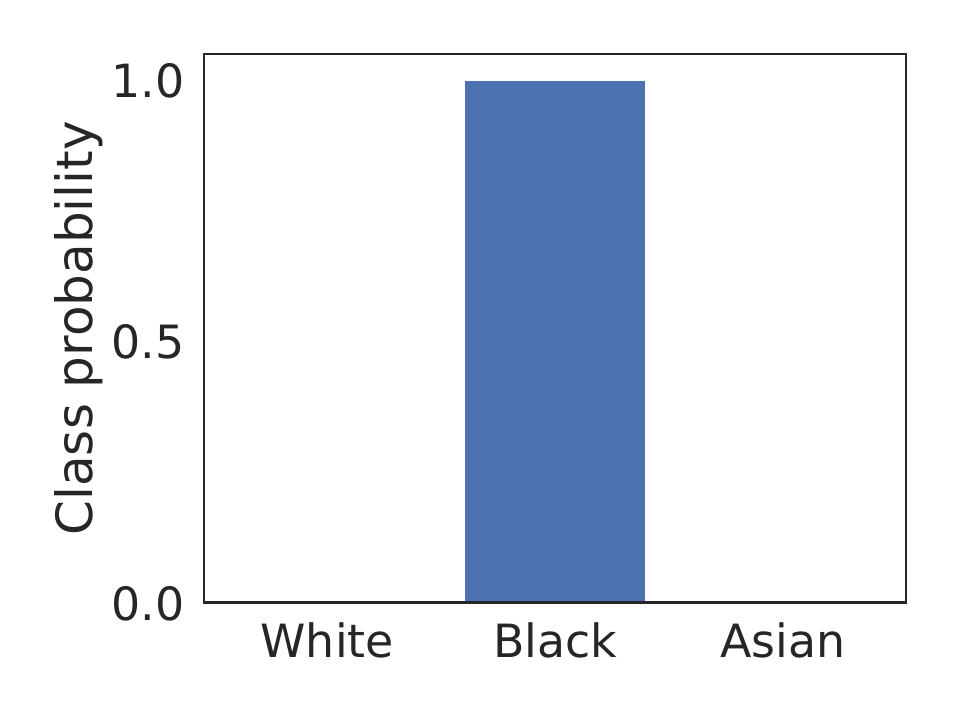}}
    
     \caption[]{Given a low-resolution input (``White'', ``Black'', or ``Asian''), we assume that the class distributions of the reconstructions are given by the following probability mass functions.}
     \label{fig:cond_class_exa2}
     \Description{A plot that visualizes the conditional class distributions corresponding to the second extreme case in Example~9.1.}
    \end{figure}
\end{example}

\subsection{Image Upsampling Models}
\label{subsec:supplement:model-descriptions}

In the following, we describe the models and specify all hyperparameters used for training and evaluation, respectively. 
\paragraph{PULSE}
PULSE upsampling leverages the expressiveness of a pretrained StyleGAN. We trained a StyleGAN2 using the official repository\footnote{\url{https://github.com/NVlabs/stylegan3}} and set $\gamma=0.0128$. The final FID-score is $5.89$ and $5.61$ if trained on \dataset{} and FairFace, respectively.  
To generate the reconstructions, we used the default values provided in the official repository\footnote{\url{https://github.com/adamian98/pulse}} but set the number of \textit{bad noise layers} to $11$ (instead of $17$) because the StyleGAN2 we use has only $12$ layers (instead of $18$).

\paragraph{pSp}
We utilize the same StyleGAN2 backbone as in PULSE. 
The pSp encoder is trained for $300\,000$ steps using the default values as provided in the official repository\footnote{\url{https://github.com/eladrich/pixel2style2pixel}}. Note that its training procedure is based on downsampling a training sample and measuring its reconstruction to the original. We observe that if the training procedure does not contain downsampling to $4\times 4$ resolution, reconstructions of $4\times 4$ inputs---which is the setup in the diversity study---are not meaningful. To compensate for that lack of generalization, we added $4\times 4$-downscaling to the training procedure. 

\paragraph{fair-pSp}
We use the same setting as in pSp but additionally apply the resampling and the curriculum learning scheme as described in~\citet{tanjim2022debiasing}. We found that using an MLP after standardizing the latents leads to bad results. Hence, we standardize the latents to compute the contrastive loss but proceed with the unstandardized latents, i.e., we ignore the MLP originally proposed by~\citet{tanjim2022debiasing}. The different behavior could be attributed to the fact that FairFace has $7$ classes, which is significantly more complex than the binary-class setup considered by~\citet{tanjim2022debiasing}. Other than that, we adopt the hyperparameters from the original paper. Since there is no official implementation, we reimplemented fair-pSp\footnote{\url{https://github.com/MikeLasz/Fair-pSp}}. 

\paragraph{PosteriorSampling}
For training the NCSNv2\footnote{\url{https://github.com/ermongroup/ncsnv2}} backbone model, we select the hyperparameters according to the techniques recommended in~\citet{song2020improved}. For \dataset{} and FairFace, this results in  $L=1022,\, \sigma_1=170, \, T=3, \, \varepsilon=1.86e-6$. We trained the models for $150\, 000$ Iterations. We generate samples from the posterior leveraging the official repository\footnote{\url{https://github.com/ajiljalal/code-cs-fairness}}. 

\paragraph{DDRM}
As a backbone model, we used a DDPM as suggested by~\citet{nichol2021improved} trained using the official repository\footnote{\url{https://github.com/openai/improved-diffusion}}. We set the number of diffusion steps to $1000$ and the channel multiplier of the UNet stages to $1,\, 1, \, 2, \, 3, \, 4$, respectively. Other than that, we stick with the suggested baseline default hyperparameters and trained for $500\,000$ iterations. For computing the reconstructions, we used the default values provided by the official repository\footnote{\url{https://github.com/bahjat-kawar/ddrm}}.

\subsection{Racial Distribution of UnfairFace}
\begin{figure}[t]
  \centering
  \subfigure[Racial distribution in FairFace.]{\label{fig:fairface_prop}\includegraphics[width=0.48\textwidth]{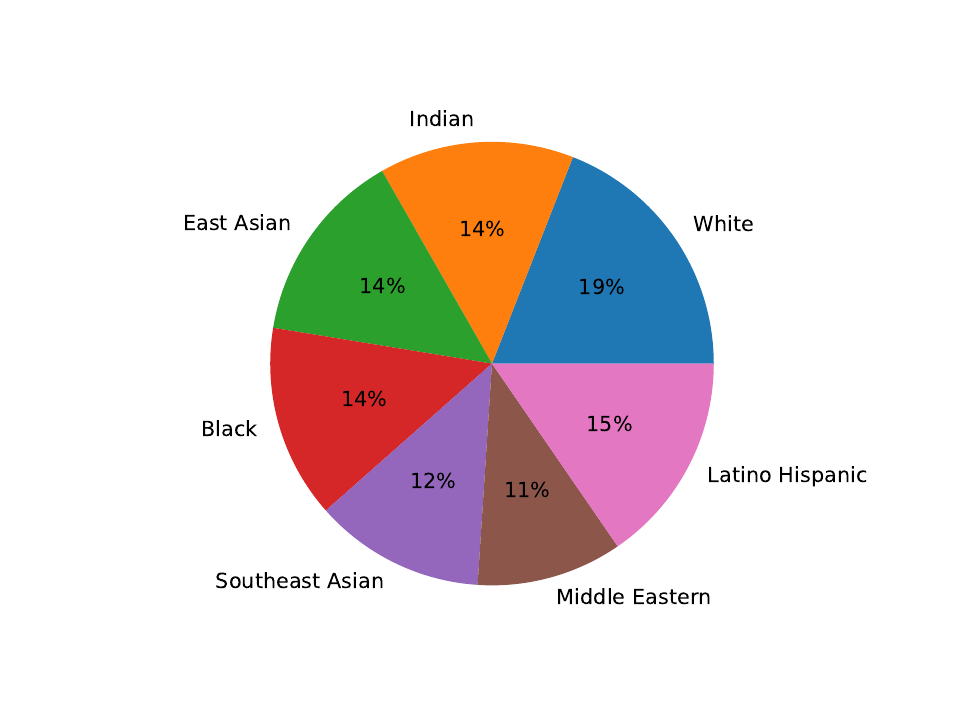}}
  \hfill %
  \subfigure[Racial distribution in \dataset. We leave out the value for Southeast Asian ($0.05\%$) to avoid value overcrowding.]{\label{fig:unfairface_prop}\includegraphics[width=0.48\textwidth]{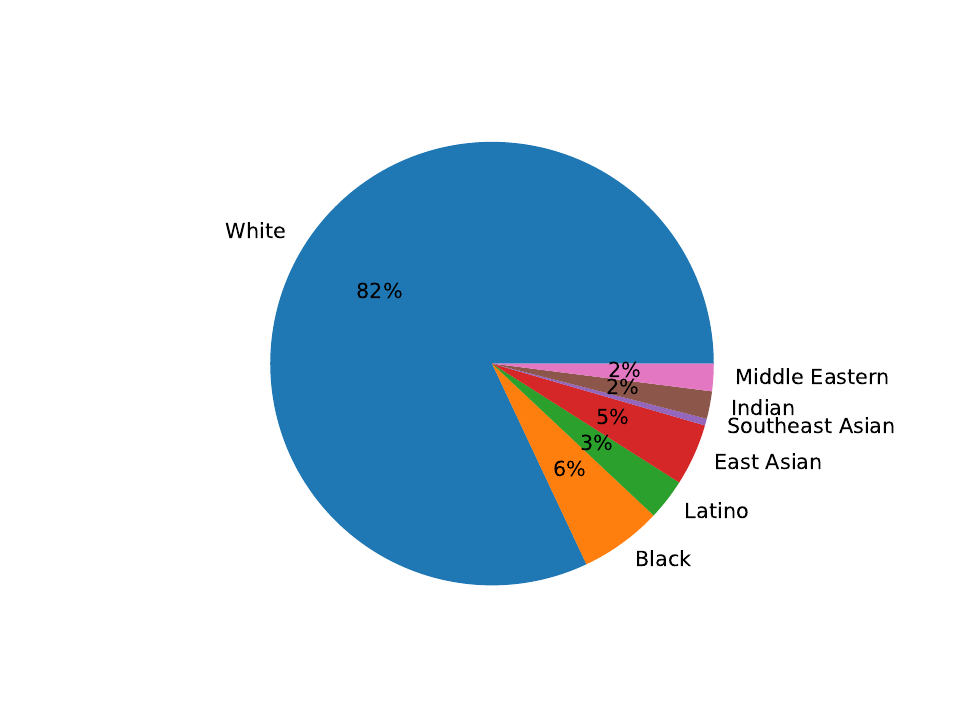}}
  \caption[]{Racial distribution of FairFace and \dataset{} in comparison. }
  \label{fig:proportions}
  \Description{A plot that compares the race distribution of the FairFace dataset with the UnfairFace dataset.}
\end{figure}

In Figure~\ref{fig:proportions}, we compare the racial distribution of FairFace and UnfairFace. 

\subsection{Additional Qualitative Results}\label{subsec:supplement:additional-qualitative-results}
\begin{figure*}
    \centering
     \subfigure[Trained on \dataset.]{\label{fig:white_app_unfairface}\includegraphics[width=\textwidth]{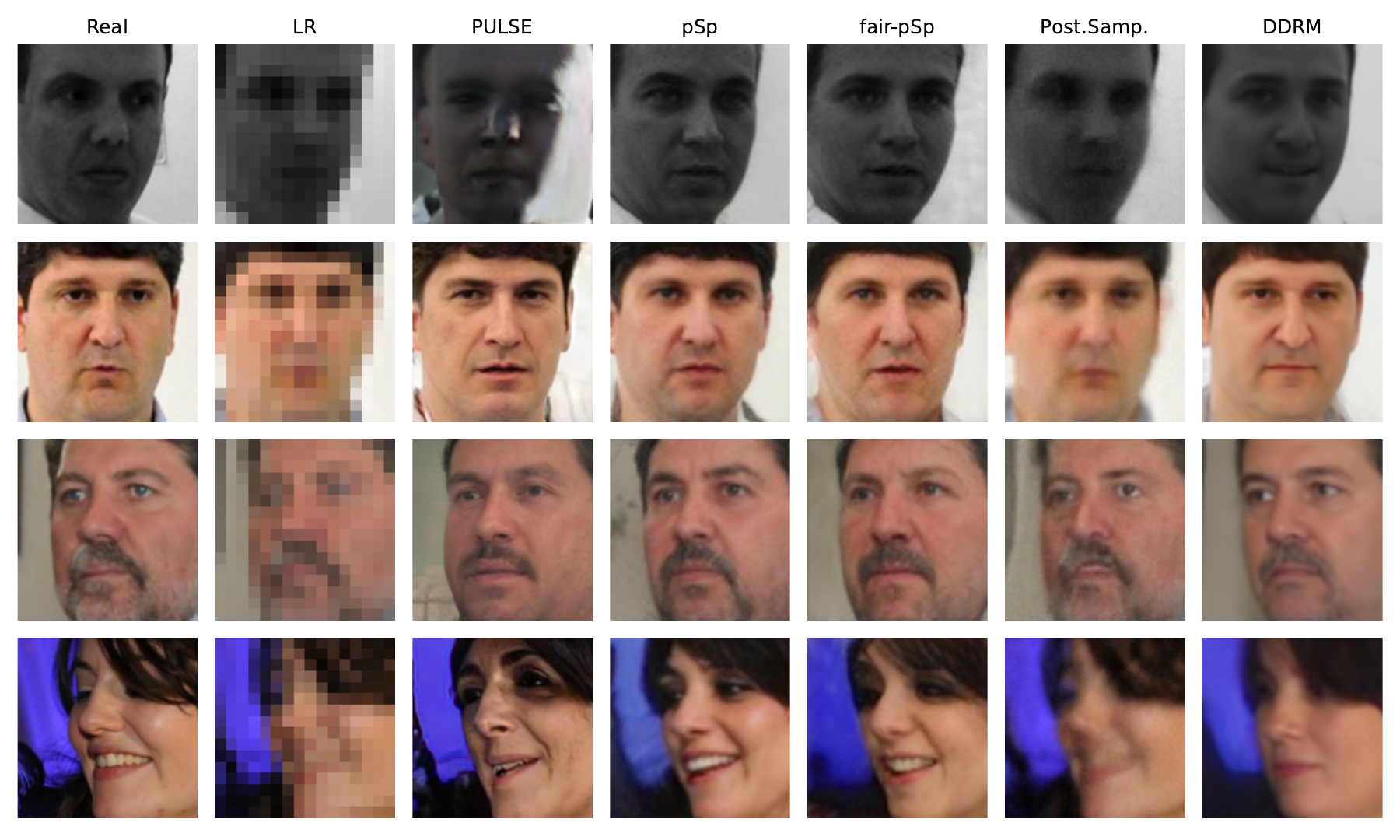}}
  \subfigure[Trained on FairFace.]{\label{fig:white_app_fairface}\includegraphics[width=\textwidth]{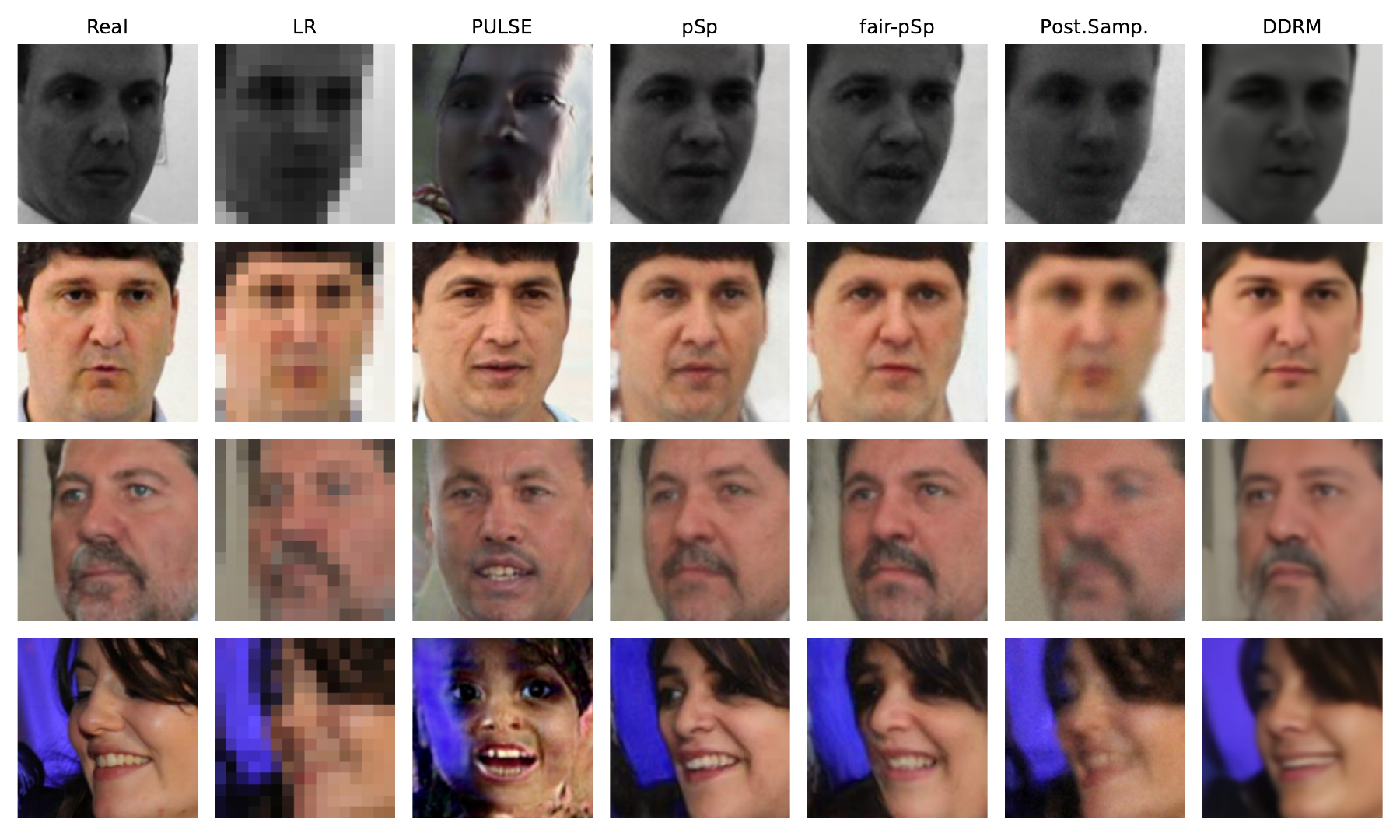}}
  \caption[]{Upsampling results for models using test samples categorized as ``White''.}
  \Description{Reconstructions based on several image upsampling algorithms if the input is classified as ``White''.}
  \label{fig:white_app}
\end{figure*}

\begin{figure*}
    \centering
     \subfigure[Trained on \dataset.]{\label{fig:indian_app_unfairface}\includegraphics[width=\textwidth]{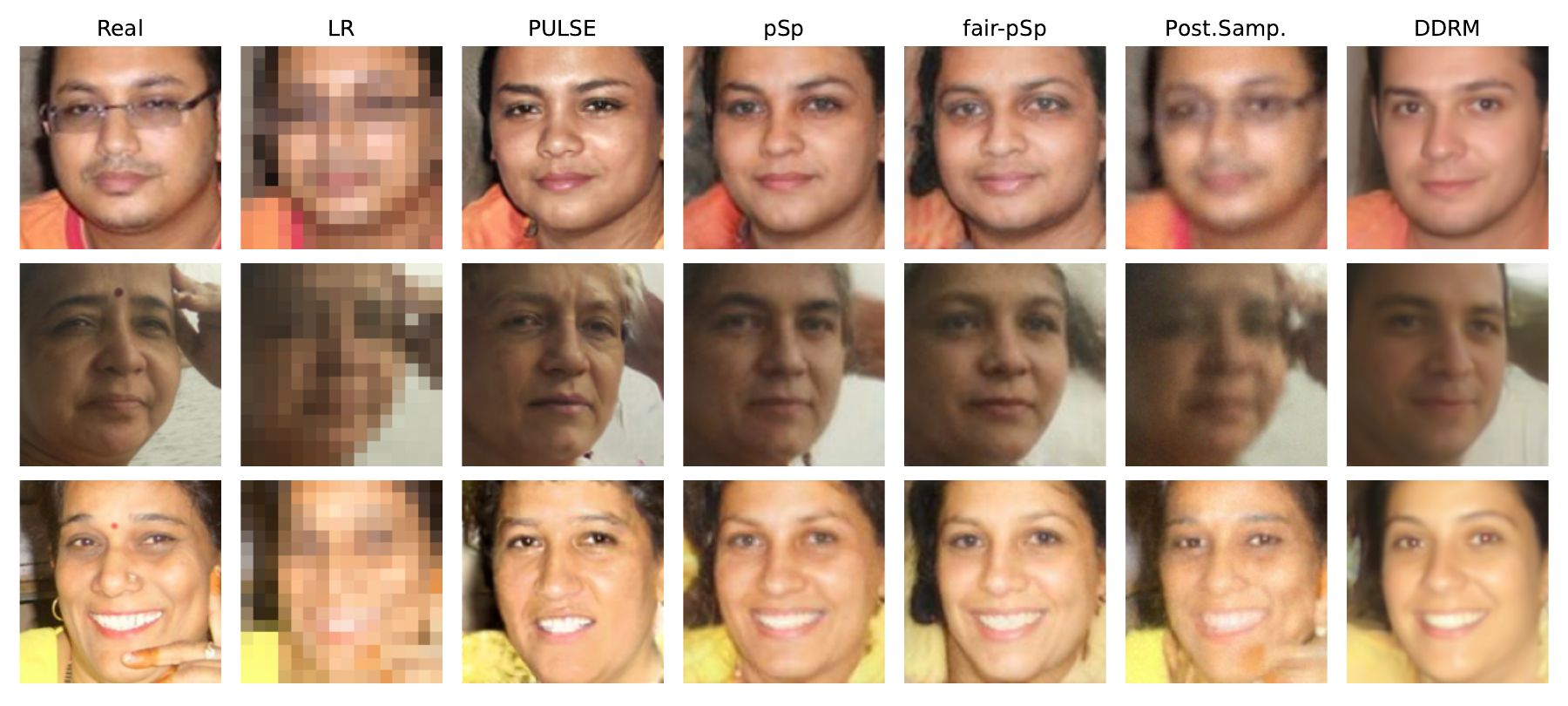}}
  \subfigure[Trained on FairFace.]{\label{fig:indian_app_fairface}\includegraphics[width=\textwidth]{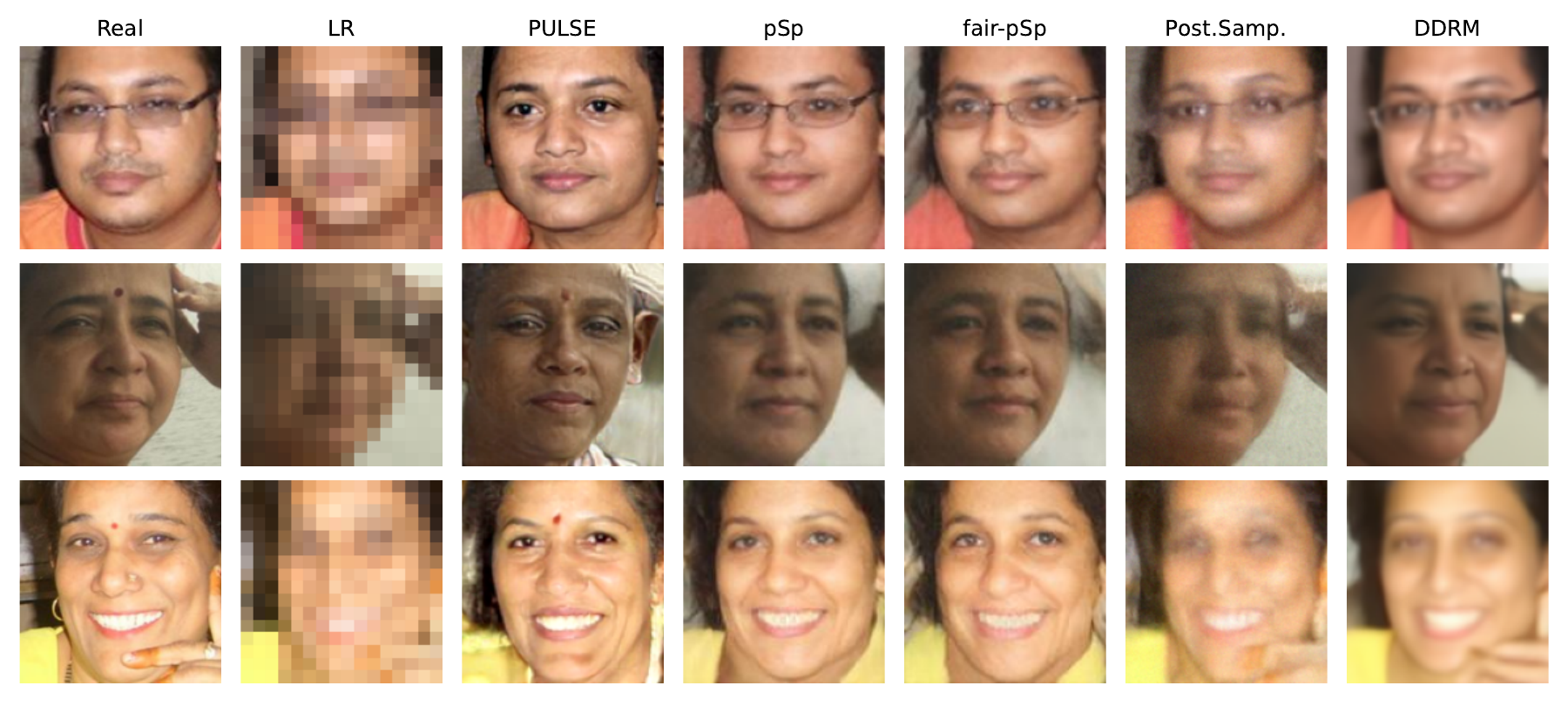}}
  \caption[]{Upsampling results for models using test samples categorized as ``Indian''.}
  \label{fig:indian_app}
  \Description{Reconstructions based on several image upsampling algorithms if the input is classified as ``Indian''.}
\end{figure*}

\begin{figure*}
    \centering
     \subfigure[Trained on \dataset.]{\label{fig:sea_app_unfairface}\includegraphics[width=\textwidth]{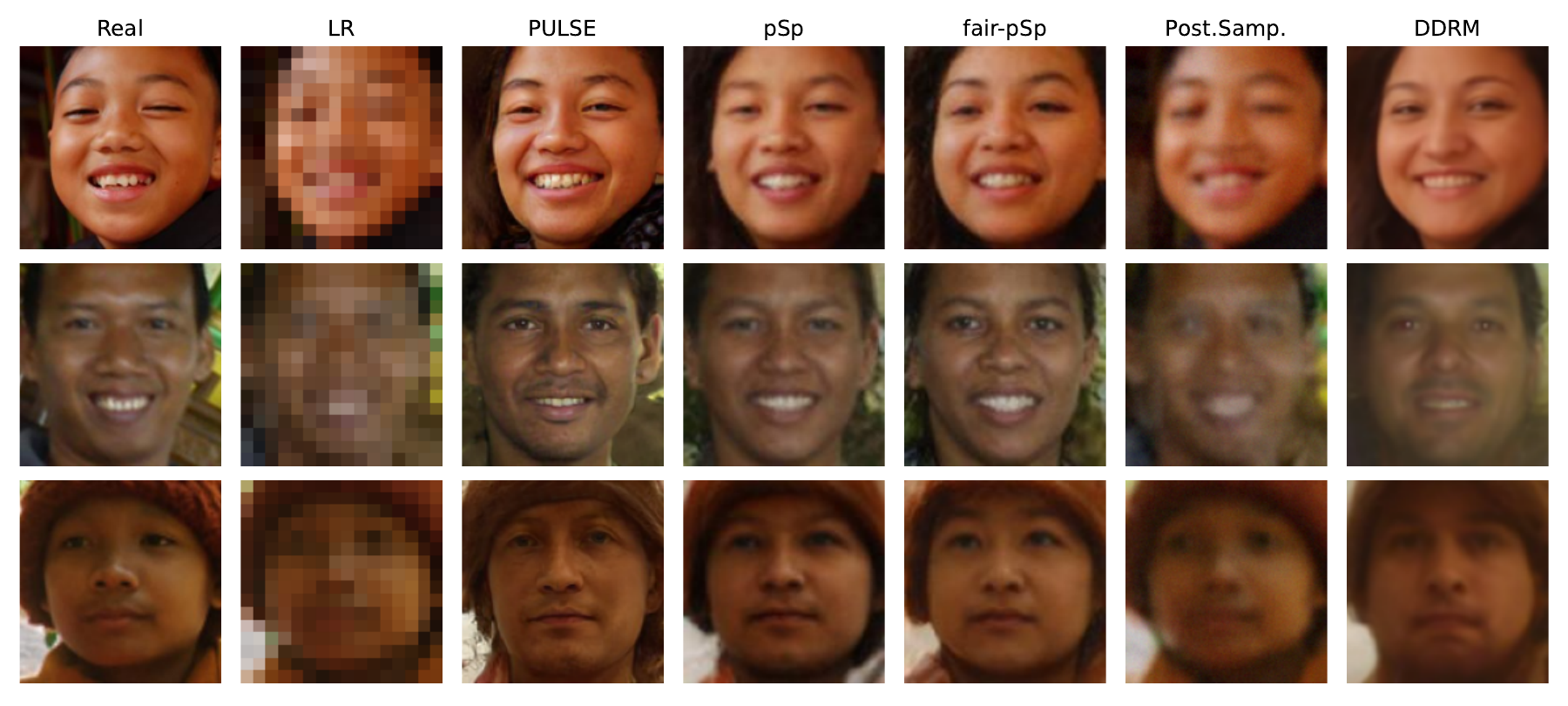}}
  \subfigure[Trained on FairFace.]{\label{fig:sea_app_fairface}\includegraphics[width=\textwidth]{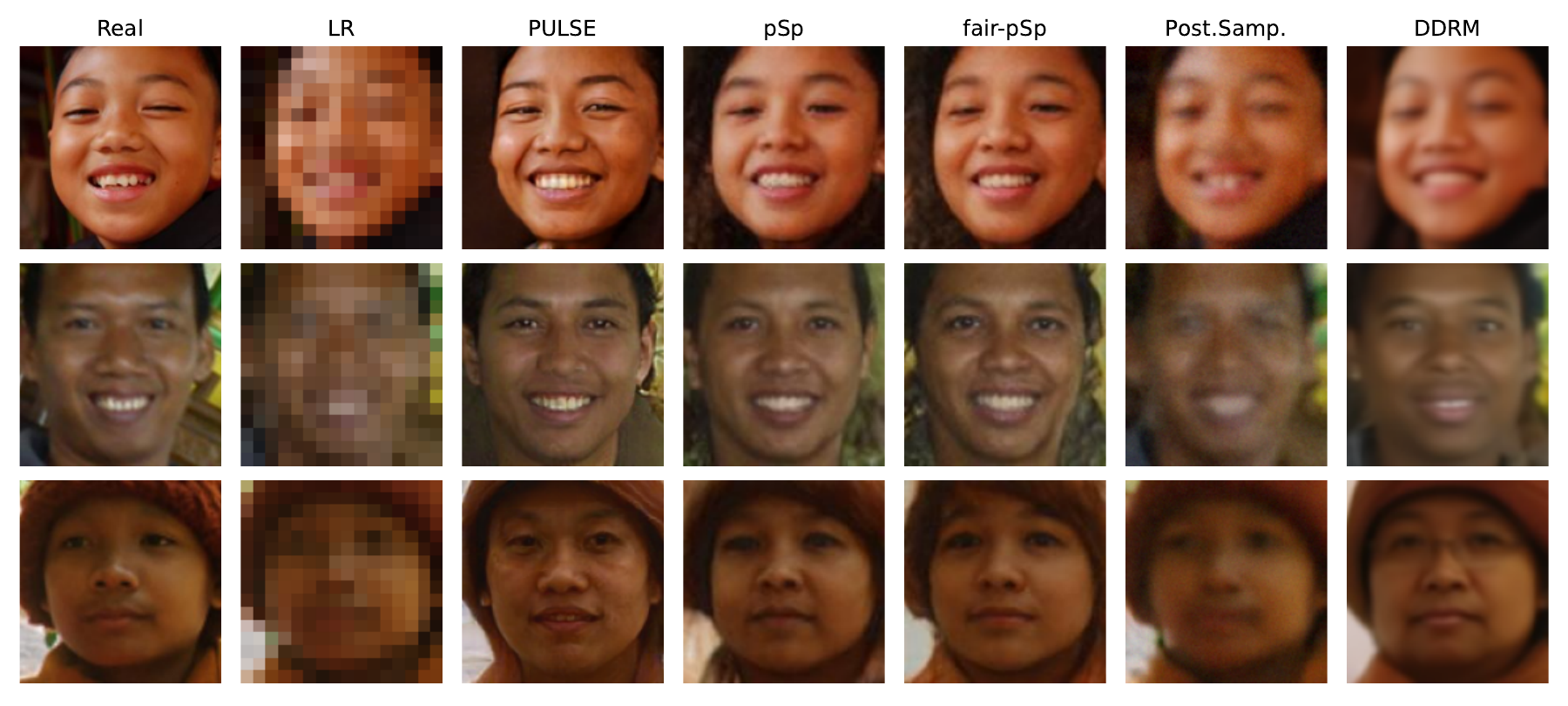}}
  \caption[]{Upsampling results for models using test samples categorized as ``Southeast Asian''.}
  \label{fig:sea_app}
  \Description{Reconstructions based on several image upsampling algorithms if the input is classified as ``Southeast Asian''.}
\end{figure*}

\begin{figure*}
    \centering
     \subfigure[Trained on \dataset.]{\label{fig:ea_app_unfairface}\includegraphics[width=\textwidth]{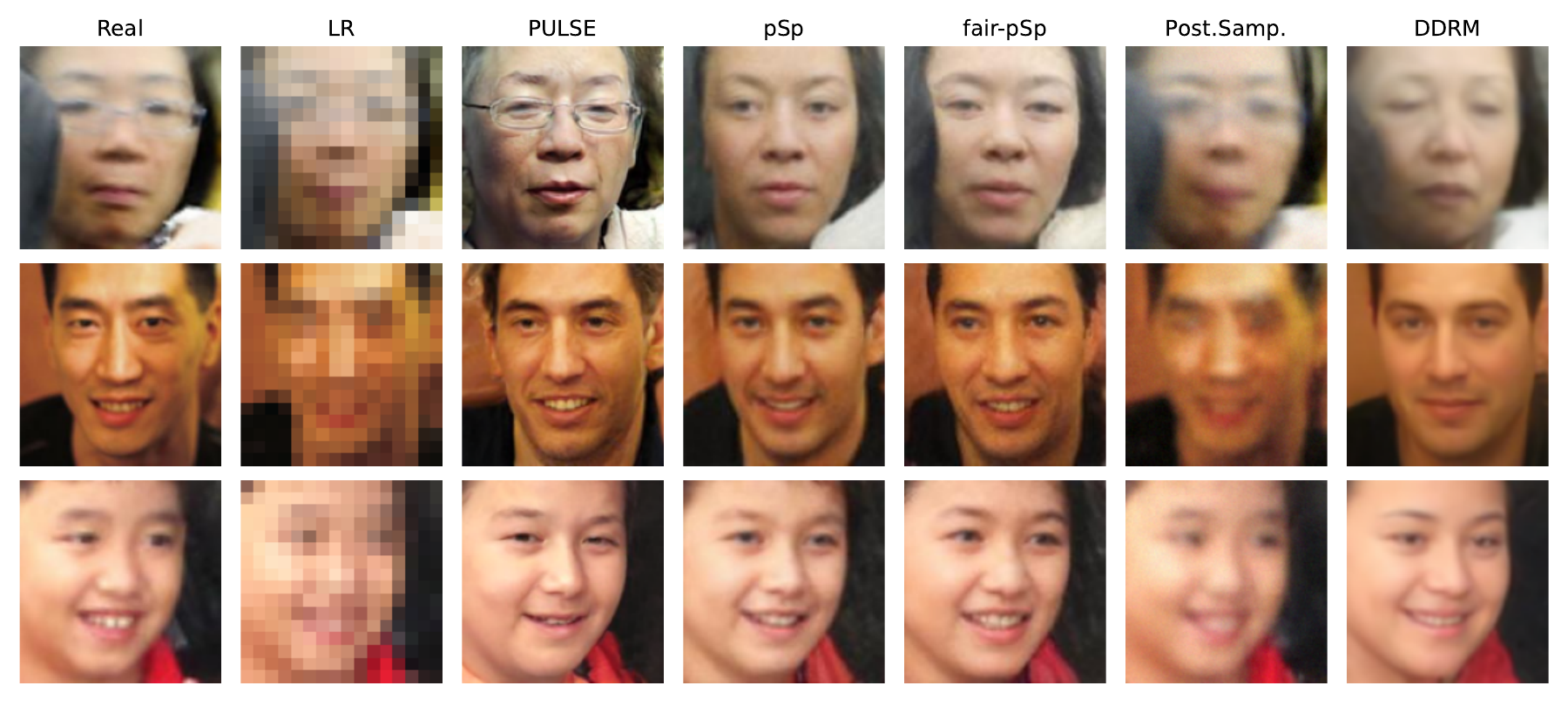}}
  \subfigure[Trained on FairFace.]{\label{fig:ea_app_fairface}\includegraphics[width=\textwidth]{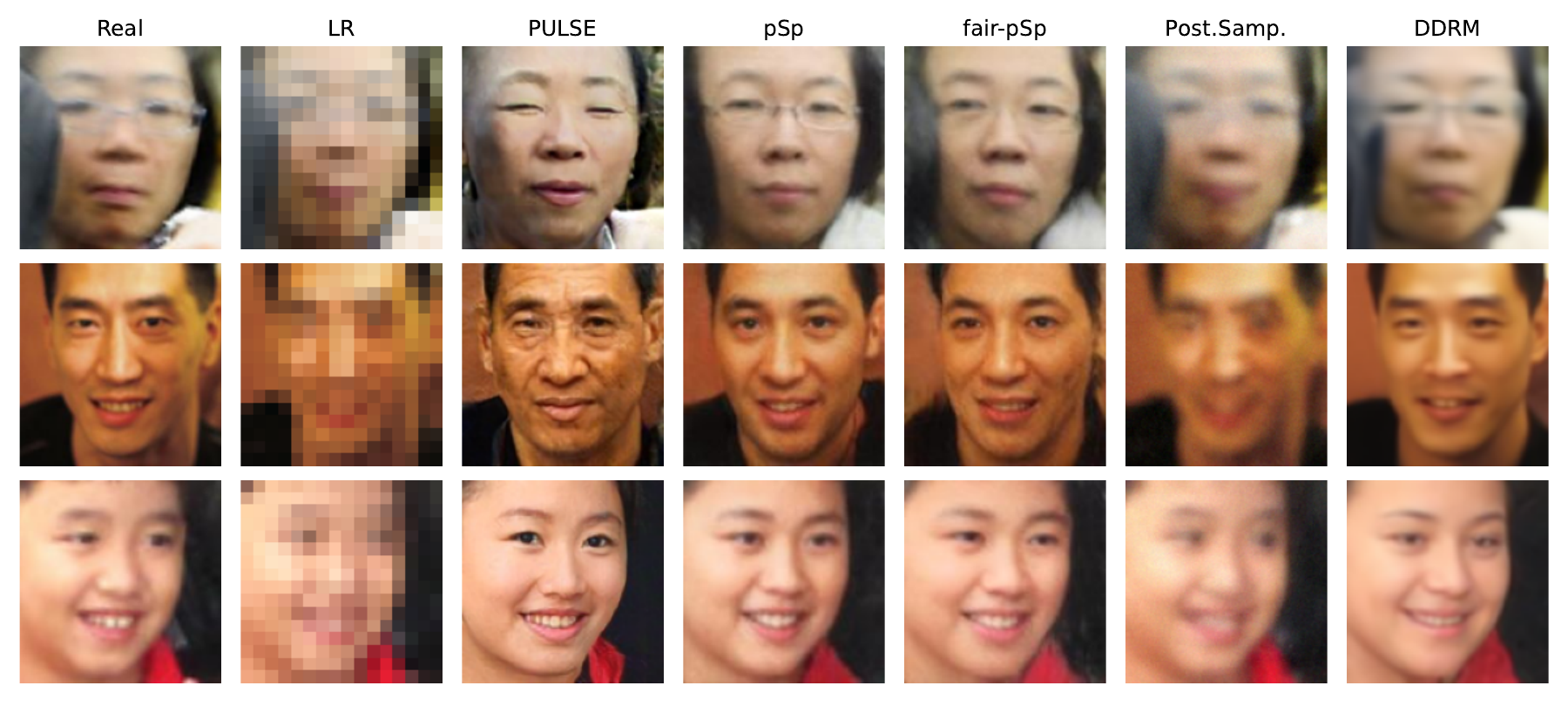}}
  \caption[]{Upsampling results for models using test samples categorized as ``East Asian''.}
  \label{fig:ea_app}
  \Description{Reconstructions based on several image upsampling algorithms if the input is classified as ``East Asian''.}
\end{figure*}

\begin{figure*}
    \centering
     \subfigure[Trained on \dataset.]{\label{fig:me_app_unfairface}\includegraphics[width=\textwidth]{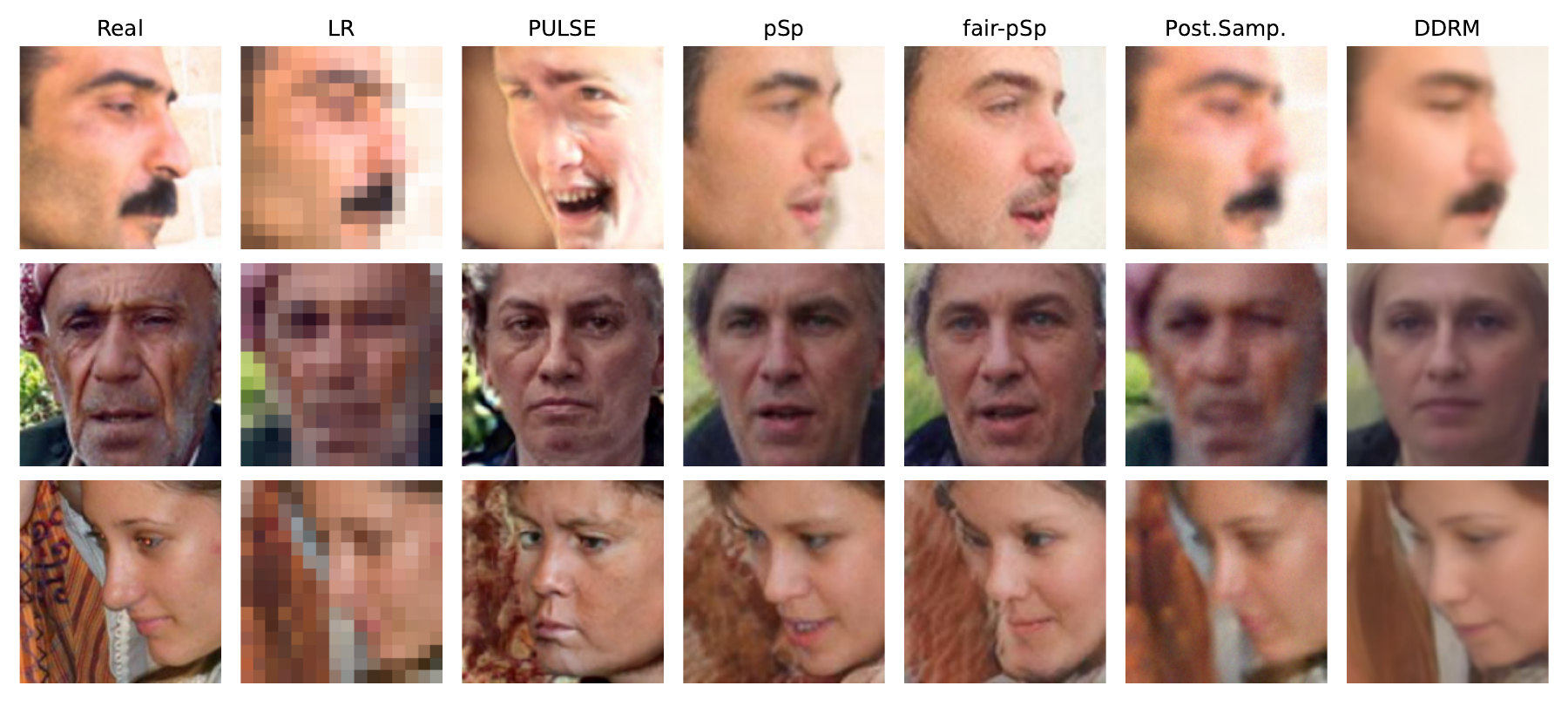}}
  \subfigure[Trained on FairFace.]{\label{fig:me_app_fairface}\includegraphics[width=\textwidth]{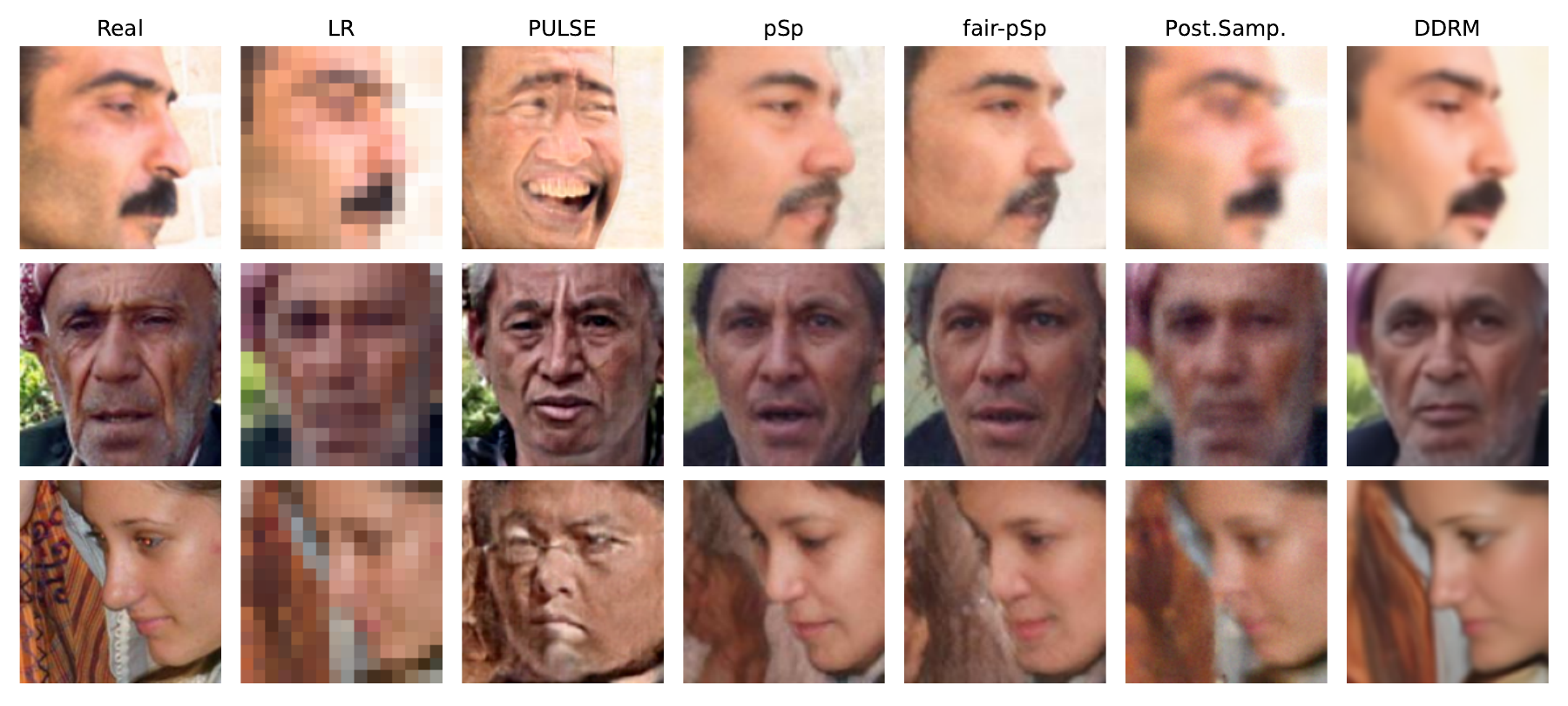}}
  \caption[]{Upsampling results for models using test samples categorized as ``Middle Eastern''.}
  \label{fig:me_app}
  \Description{Reconstructions based on several image upsampling algorithms if the input is classified as ``Middle Eastern''.}
\end{figure*}

\begin{figure*}
    \centering
     \subfigure[Trained on \dataset.]{\label{fig:lh_app_unfairface}\includegraphics[width=\textwidth]{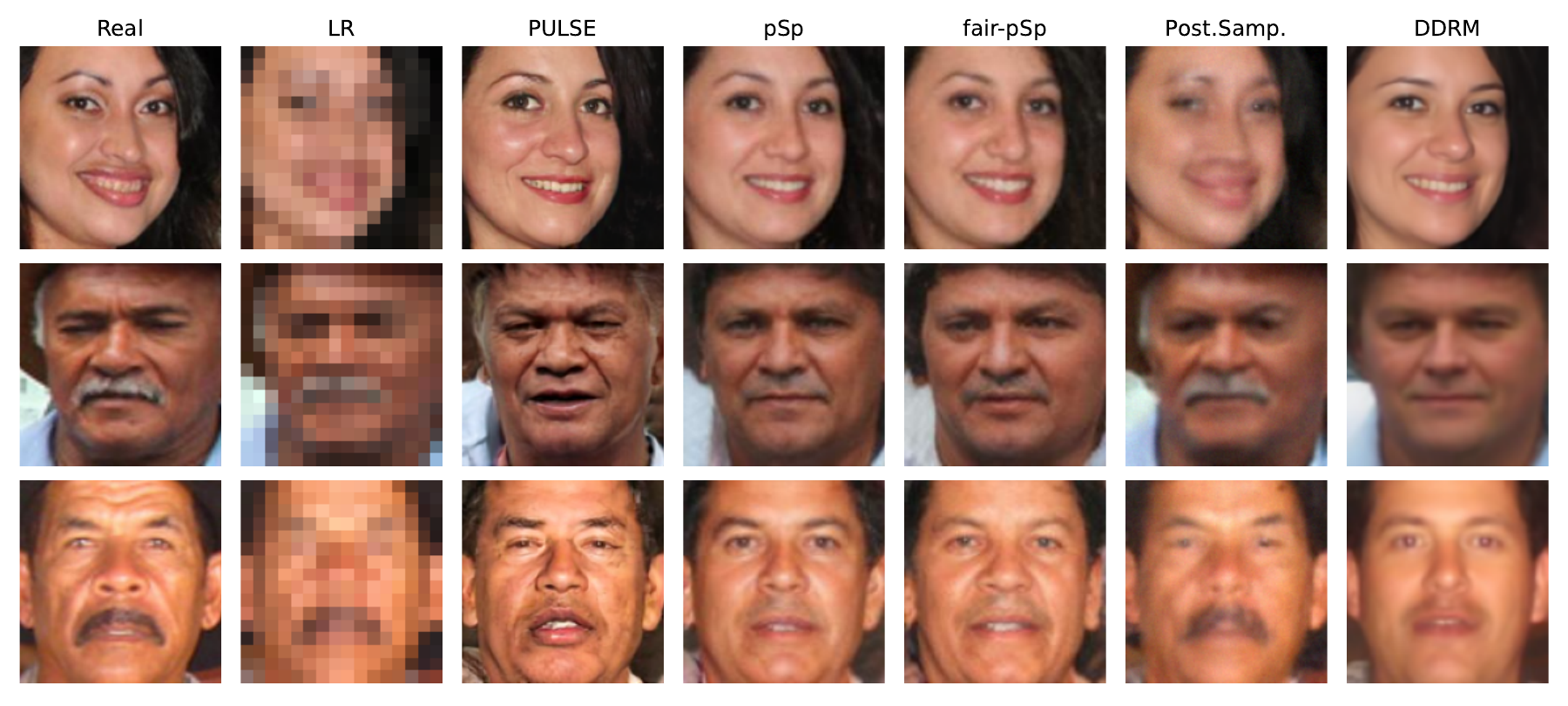}}
  \subfigure[Trained on FairFace.]{\label{fig:lh_app_fairface}\includegraphics[width=\textwidth]{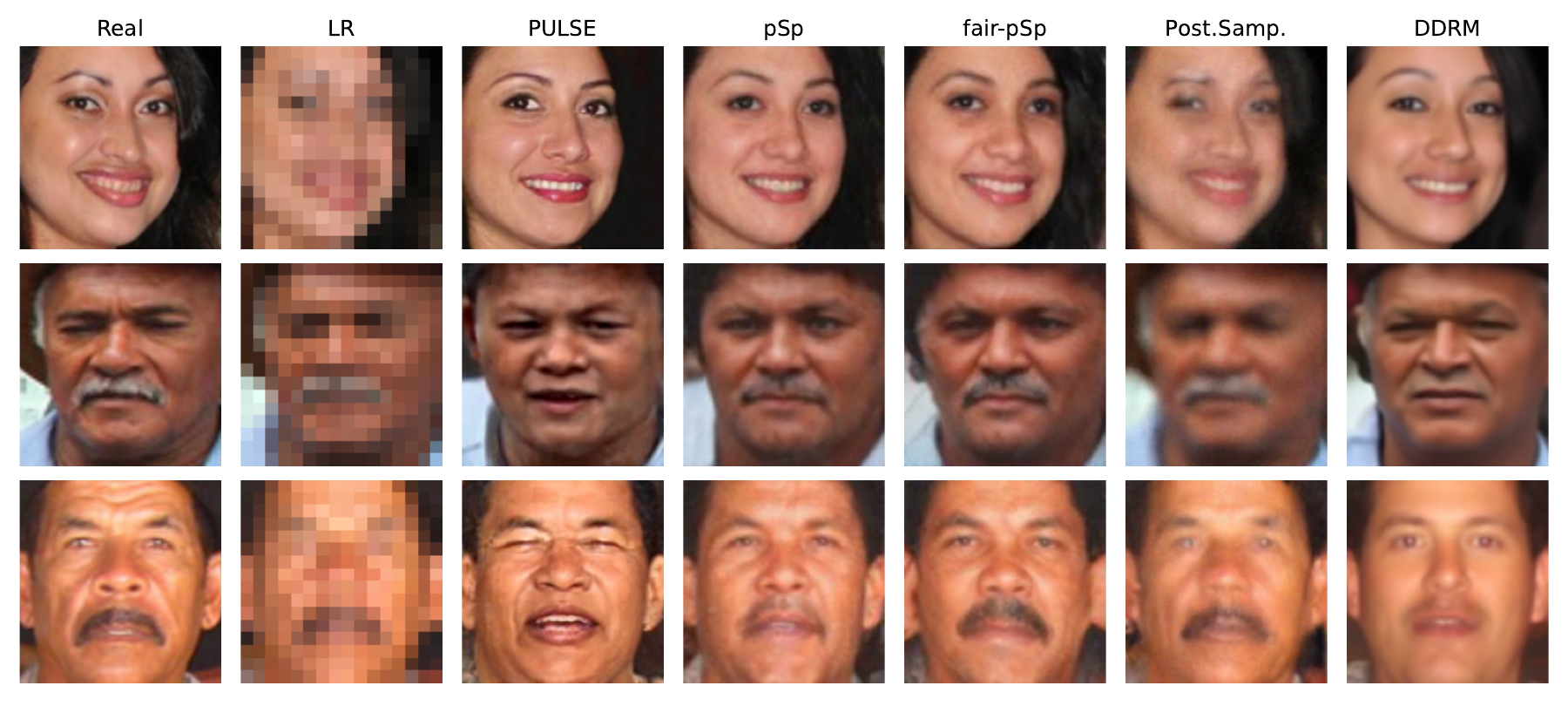}}
  \caption[]{Upsampling results for models using test samples categorized as ``Latino Hispanic''.}
  \label{fig:lh_app}
  \Description{Reconstructions based on several image upsampling algorithms if the input is classified as ``Latino Hispanic''.}
\end{figure*}

\begin{figure*}
  \centering
  \includegraphics[width=\textwidth]{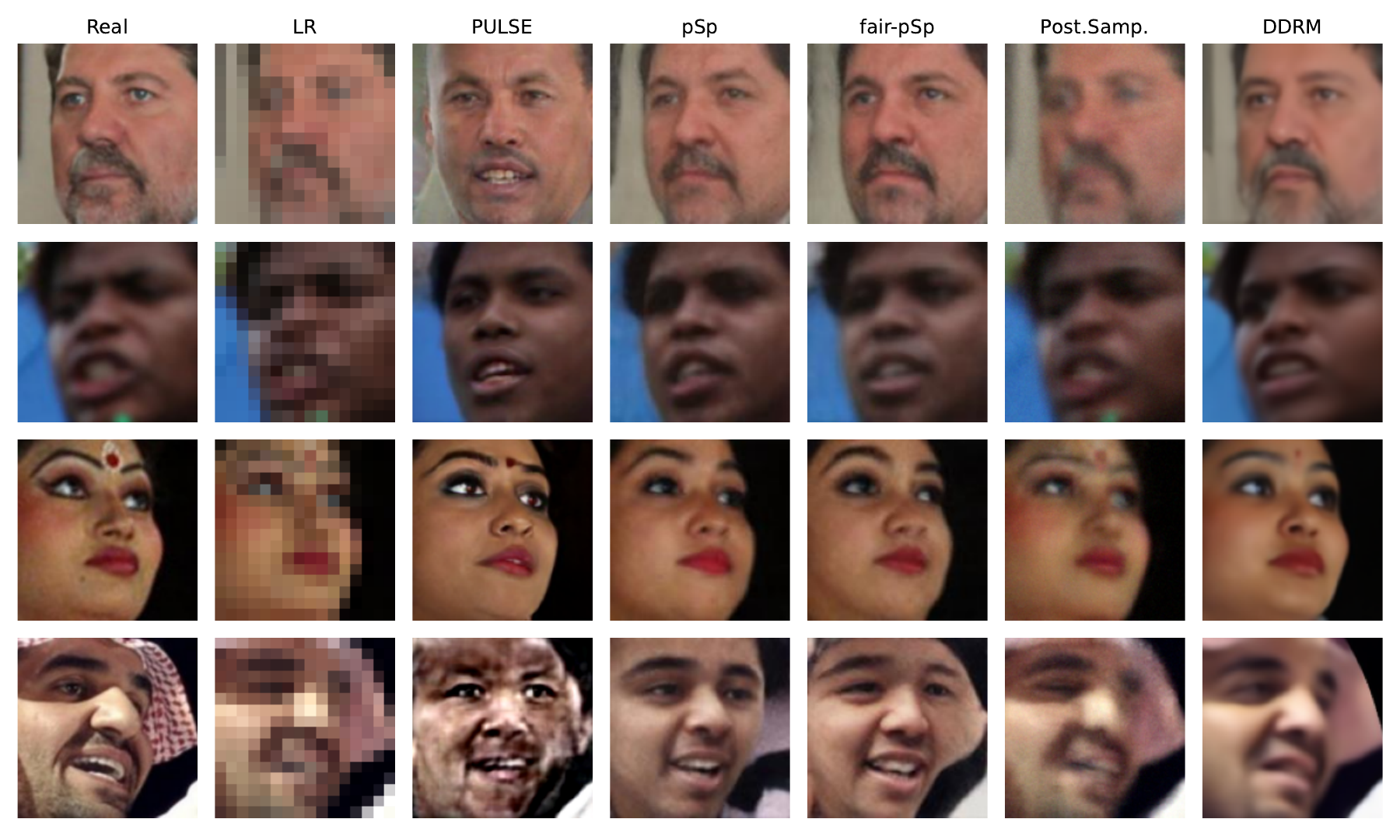}
  \caption[]{
  Upsampling results for the samples provided in Figure~\ref{fig:teaser} using models trained on the FairFace dataset.}
  \label{fig:teaser_fairface}
  \Description{Comparing the reconstructions based on several image upsampling algorithms. The models were trained on FairFace.}
\end{figure*}

\begin{figure*}
    \centering
     \subfigure[Trained on \dataset.]{\includegraphics[width=\textwidth]{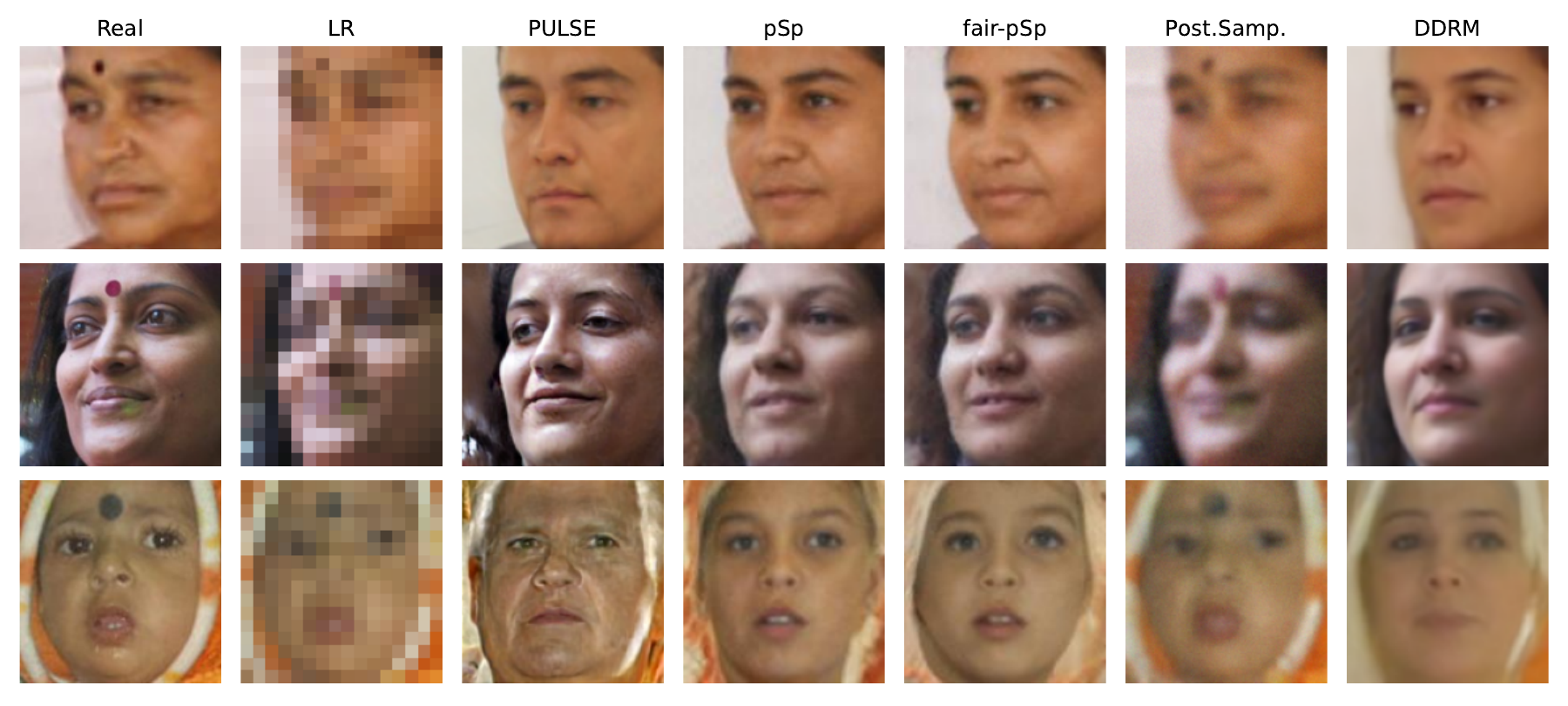}}
  \subfigure[Trained on FairFace.]{\includegraphics[width=\textwidth]{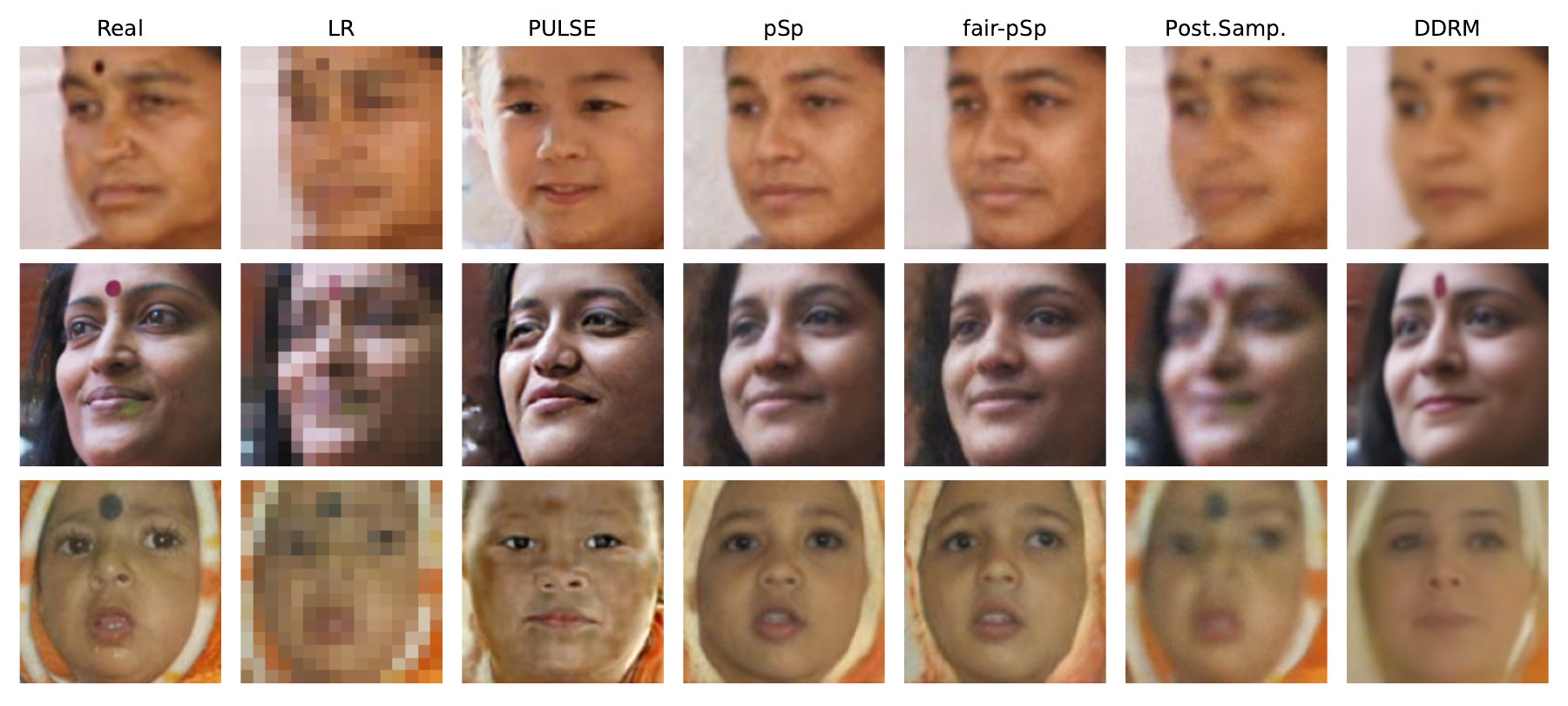}}
  \caption[]{Upsampling results for models using test samples showing a bindi.}
  \label{fig:bindi}
  \Description{Reconstructions based on several image upsampling algorithms if the input is showing a person with a bindi.}
\end{figure*}

\Cref{fig:white_app} to Figure~\ref{fig:lh_app} show additional upsampling results of test samples categorized as the remaining six races, ``White'', ``Indian'', ``Southeast Asian'', ``East Asian'', ``Middle Eastern'', ``Latino Hispanic'', respectively.
\Cref{fig:teaser_fairface} reconstructs the samples provided in Figure~\ref{fig:teaser} when the models are trained on FairFace. We observe that even if the models are trained on FairFace, reconstructing headscarves remains a difficult task, as shown by the blurry reconstructions.
Additional reconstructions of people having bindis, headscarves, and monolid eyes are provided in \Cref{fig:bindi}, Figure~{fig:scarves}, and Figure~{fig:monolid}, respectively.

\begin{figure*}
    \centering
     \subfigure[Trained on \dataset.]{\includegraphics[width=\textwidth]{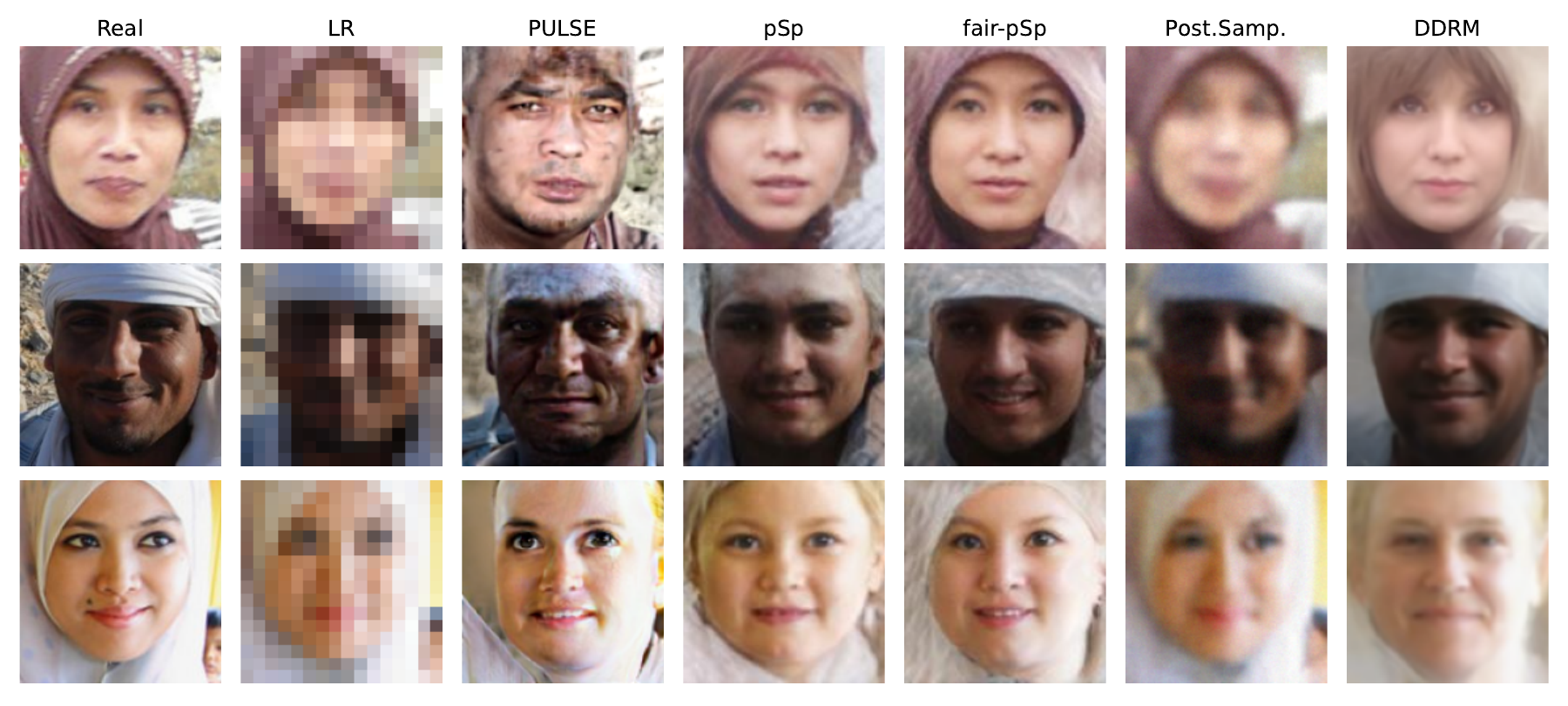}}
  \subfigure[Trained on FairFace.]{\includegraphics[width=\textwidth]{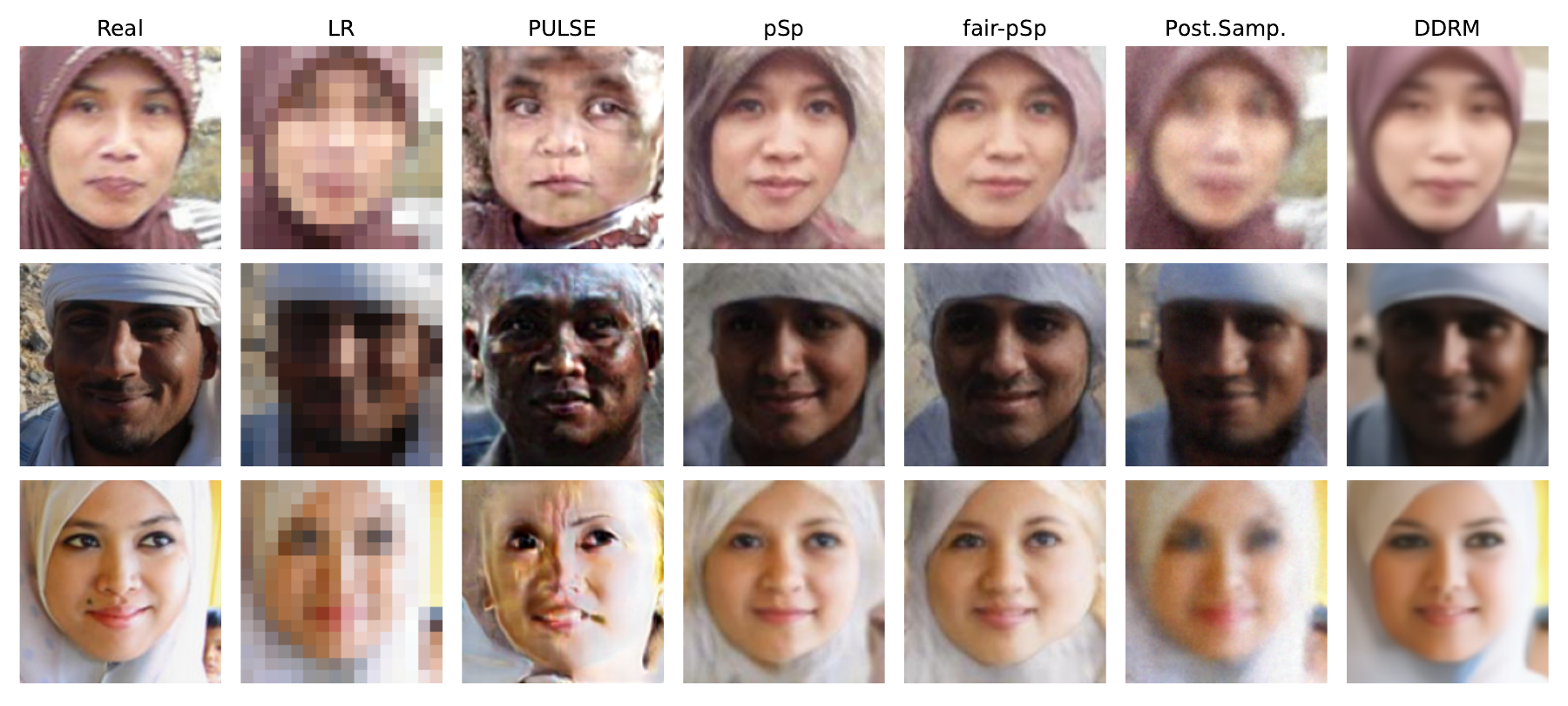}}
  \caption[]{Upsampling results for models using test samples showing a headscarf.}
  \label{fig:scarves}
  \Description{Reconstructions based on several image upsampling algorithms if the input is showing a person wearing a headscarf.}
\end{figure*}

\begin{figure*}
    \centering
     \subfigure[Trained on \dataset.]{\includegraphics[width=\textwidth]{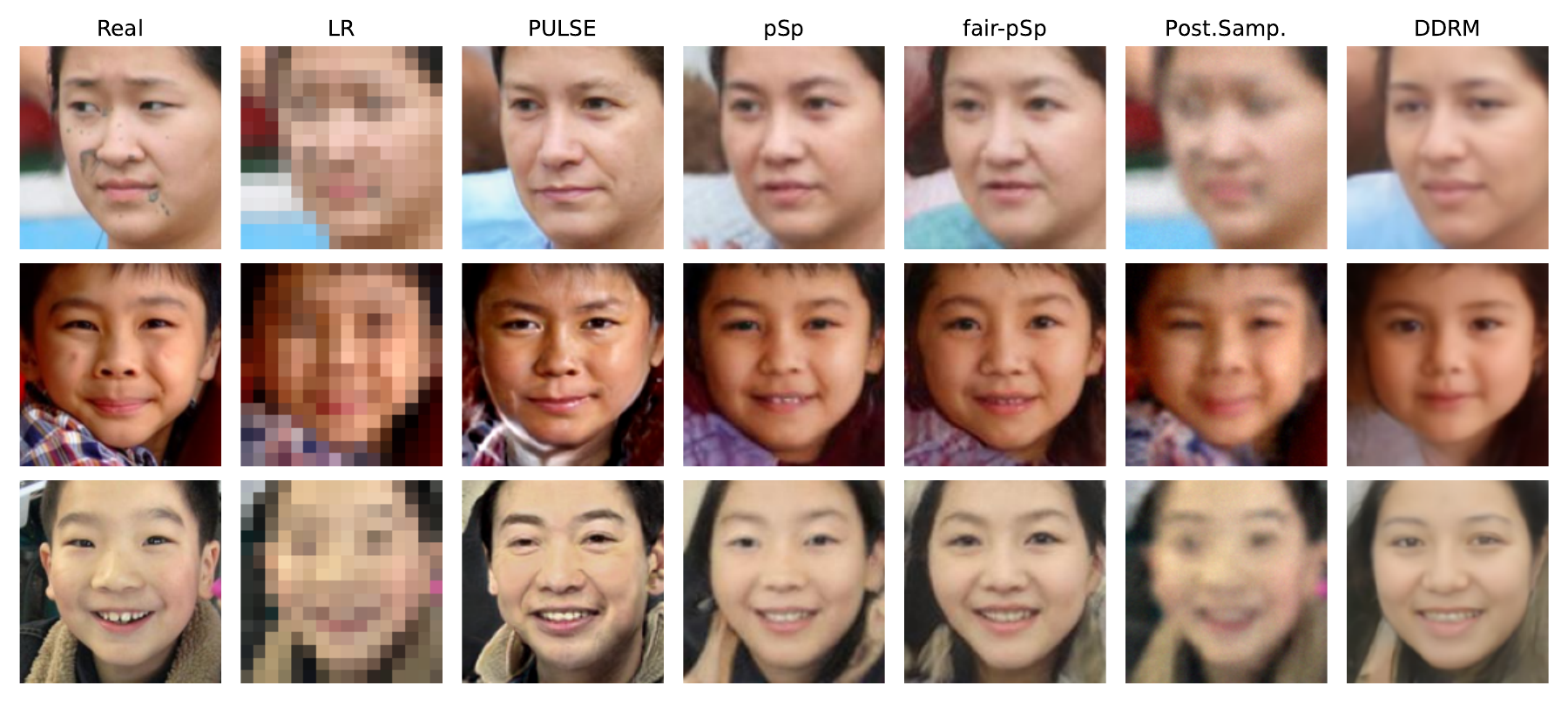}}
  \subfigure[Trained on FairFace.]{\includegraphics[width=\textwidth]{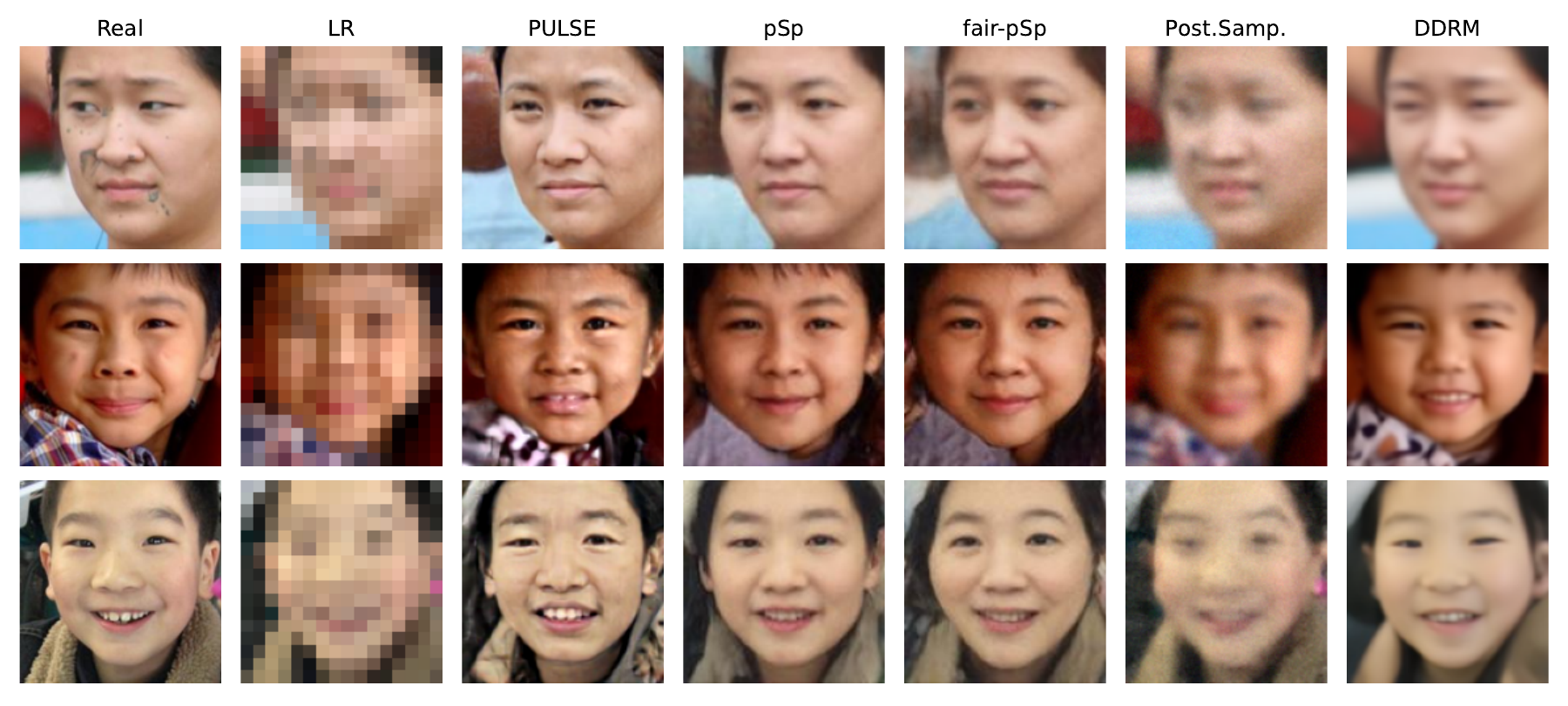}}
  \caption[]{Upsampling results for models using test samples showing monolid eyes.}
  \label{fig:monolid}
  \Description{Reconstructions based on several image upsampling algorithms if the input is showing a person with monolid eyes.}
\end{figure*}

\subsection{Additional Quantitative Results}
\label{subsec:supplement:additional-quantitative-results}
To indicate the significance of the difference of a performance score obtained by a model trained on \dataset{} and FairFace, we provide the corresponding P-values in \Cref{tab:pvalues_perf}. For the reconstruction-based losses (i.e., LPIPS and DSSIM), we find the biggest significance for pSp and DDRM. The training data set has the most significant influence on race reconstruction performance (i.e., $L_{\operatorname{race}}^{\cos}$ and $L_{\operatorname{race}}^{\text{0-1}}$) for DDRM as indicated by the smallest p-value. The image quality differs most significantly for DDRM in terms of the NIQE score and for Posterior Sampling in terms of the blurriness index. 

\begin{figure*}[htbp]%
  \centering
  \subfigure[Models trained on \dataset.]{\label{fig:rdp-unfairface}\includegraphics[width=0.48\textwidth]{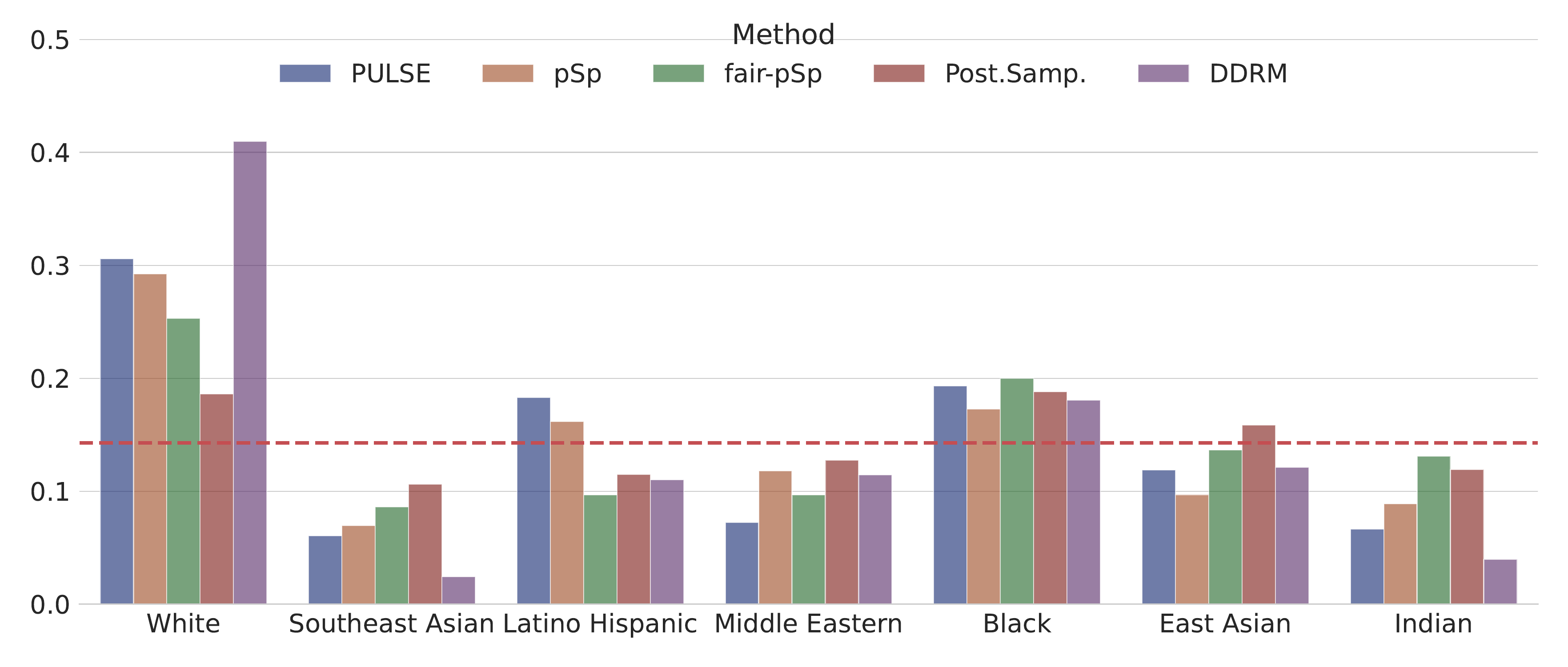}}
  \hfill %
  \subfigure[Models trained on FairFace.]{\label{fig:rdp-fairface}\includegraphics[width=0.48\textwidth]{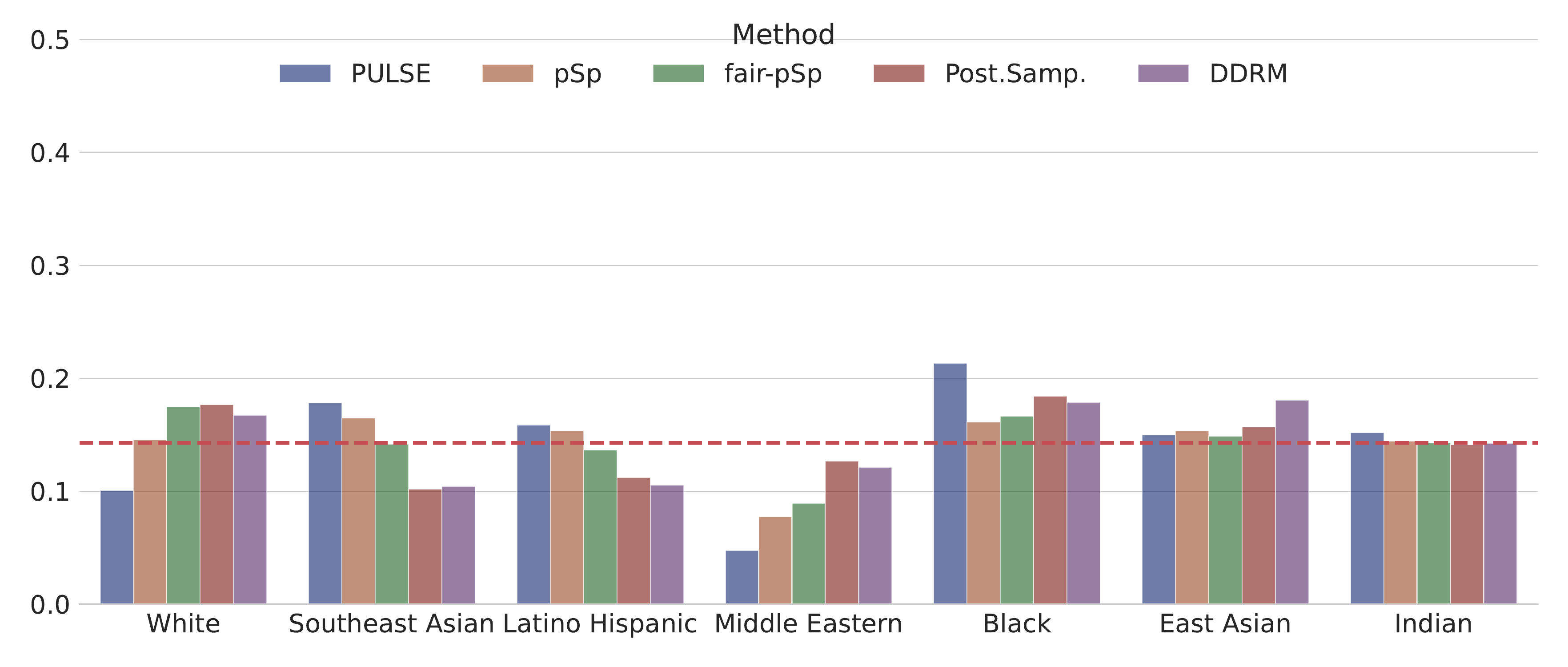}}
  \caption[]{Comparing the representation demographic parity distribution $\Prdp$ for each model trained on \dataset{} and FairFace. The horizontal dashed line indicates the bar height corresponding to a uniform distribution.}
  \Description{Two plots that show the representation demographic parity distribution for each upsampling algorithm. The left plot shows the results of the algorithms trained on UnfairFace and the right plot shows the results of the algorithms trained on FairFace.}
  \label{fig:rdp}
\end{figure*}

In Figure~\ref{fig:rdp}, we visualize $\Prdp$ for each model. Let us highlight two points to clarify the proposed metrics based on RDP. First, this figure presents a rescaled version of Figure~\ref{fig:0-1_per_eth} in which the bars corresponding to a model are normalized such that they add up to one (compare with equation~\ref{eq:Prdp}). Second, the proposed metrics are divergences that measure the discrepancy of $\Prdp$ to the horizontal dashed line representing the uniform distribution in Figure~\ref{fig:rdp}. Specifically, $\Delta_{\operatorname{RDP-}\chi^2}$ (see equation~\ref{eq:delta_rdpchi}) measures the scaled mean-square distance while $\Delta_{\operatorname{RDP-Cheb}}$ (see equation~\ref{eq:delta_rdpcheb}) measures the maximum distance of the bars to the dashed line. The results align with the numbers presented in Table~\ref{tab:fairness}. 

\begin{figure*}[htbp]%
  \centering
  \subfigure[Models trained on \dataset.]{\label{fig:pr_unfairface}\includegraphics[width=0.48\textwidth]{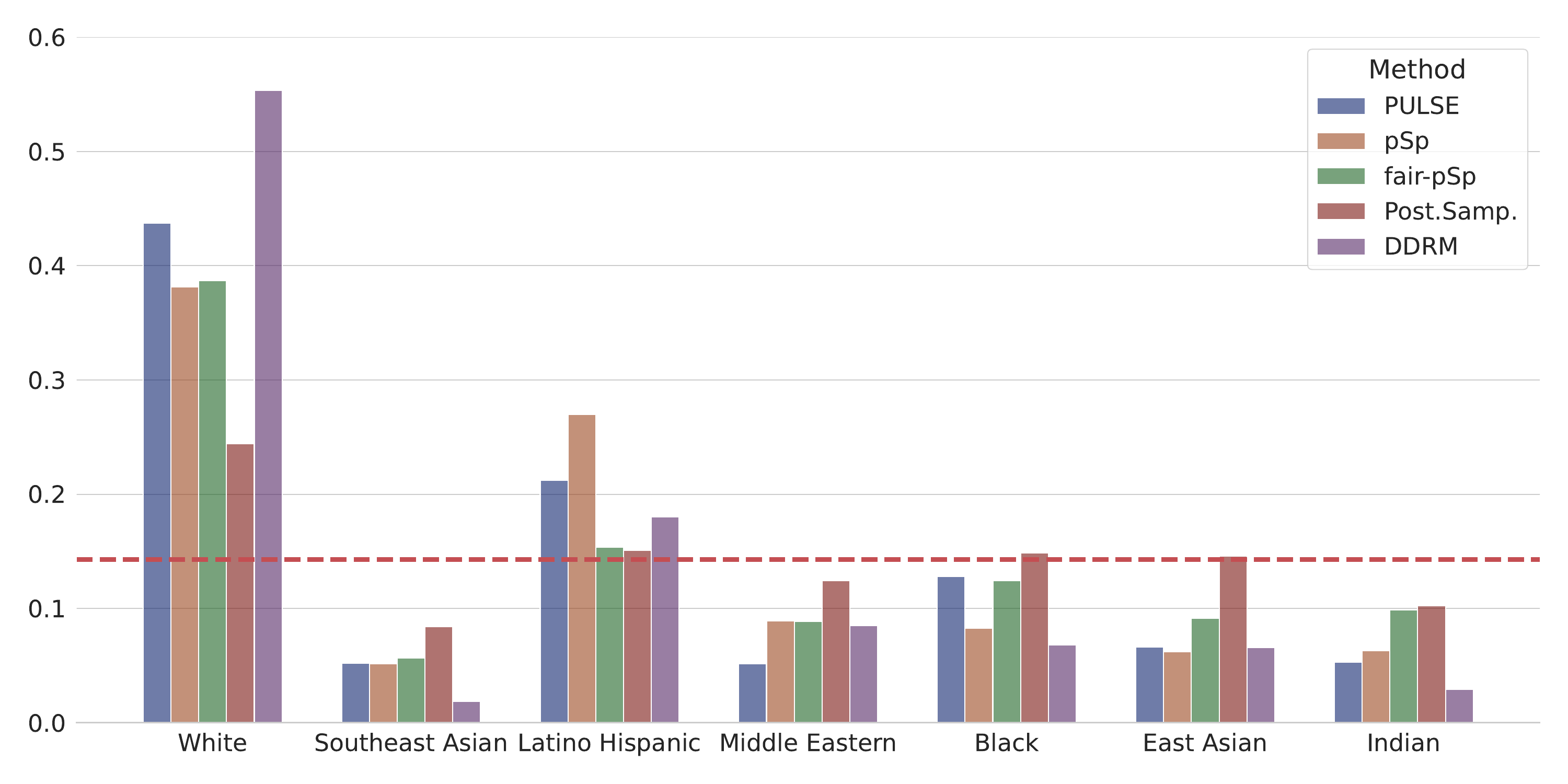}}
  \hfill %
  \subfigure[Models trained on FairFace.]{\label{fig:pr_fairface}\includegraphics[width=0.48\textwidth]{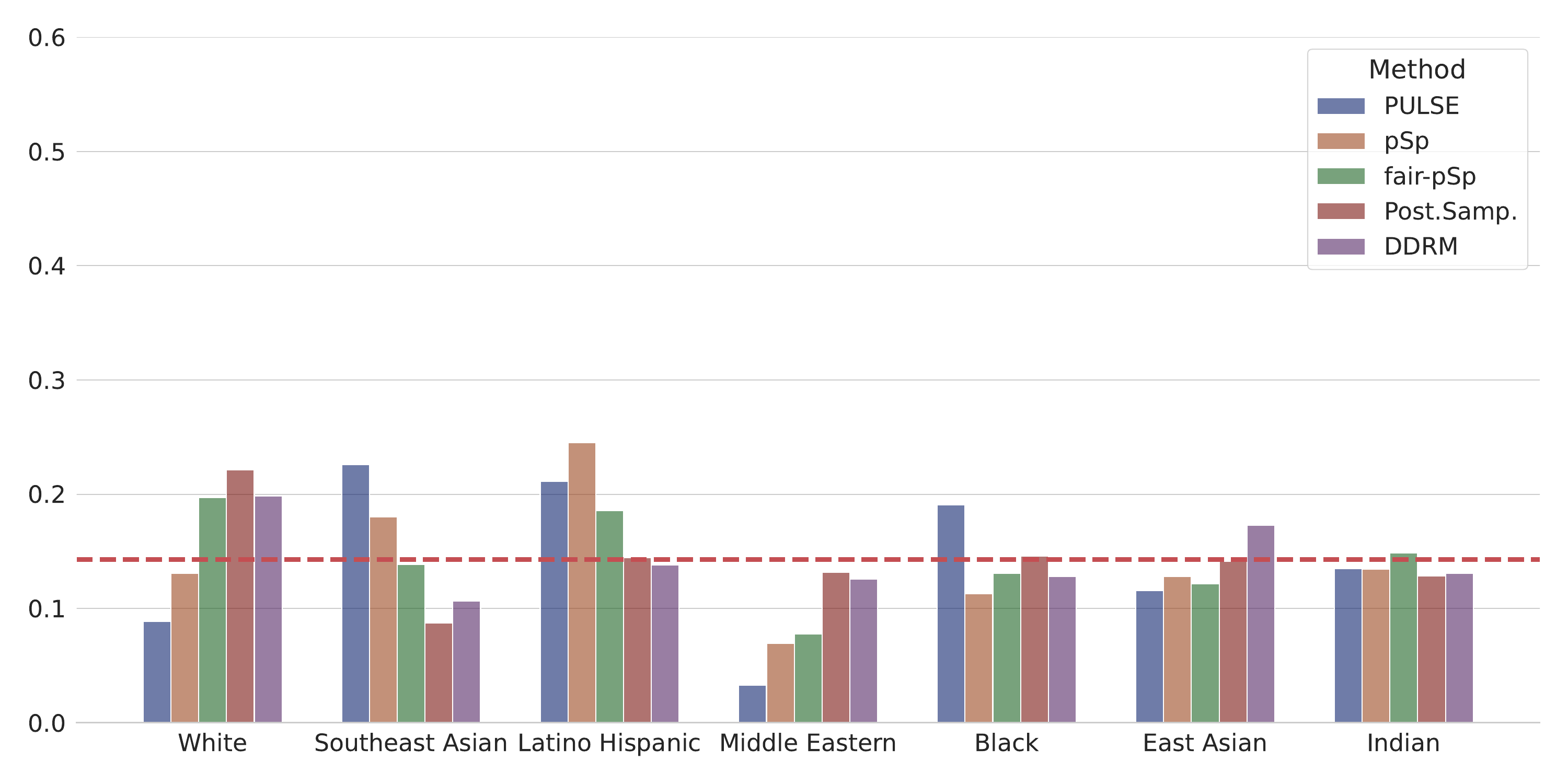}}
  \caption[]{Comparing the proportional representation distribution $\Ppr$ for each model trained on \dataset{} and FairFace. The horizontal dashed line indicates the bar height corresponding to a uniform distribution.}
  \Description{Two plots that show the proportional representation distribution for each upsampling algorithm. The left plot shows the results of the algorithms trained on UnfairFace and the right plot shows the results of the algorithms trained on FairFace.}
  \label{fig:pr}
\end{figure*}

\Cref{fig:pr} visualizes $\Ppr$, which---once again---highlights the influence that a lack of diversity in the training data has on the resulting racial distribution.    
\begin{table*}[t]
    \centering
    \caption[]{P-values for the tests utilized in Table~\ref{tab:performance}. Bold values indicate values that are above the $\alpha$-level.}
    \label{tab:pvalues_perf}
    \begin{tabular}{lrcrcrcrcrcr}
\toprule
    & LPIPS && DSSIM & &$L_{\operatorname{race}}^{\operatorname{cos}}$ & &$L_{\operatorname{race}}^{\text{0-1}}$ && NIQE & &BLUR 
    \\ 
\midrule
    PULSE & \textbf{9.37e-01} && \textbf{3.80e-01} && 6.06e-10 && 7.22e-3  && 7.37e-05 && 7.03e-05 \\
    pSp & 3.71e-190 && 2.76e-192 && 7.39e-59 && 2.31e-17 && 4.20e-02 && 2.51e-03 \\
    fair-pSp & 1.42e-119 && 5.73e-145 && 1.74e-49 && 2.31e-17 && 7.78e-18 && 4.14e-07 \\
    Post.Samp. & 4.60e-08 && 9.51e-31 && 5.81e-06 && \textbf{4.18e-1} && 1.35e-58 && 2.25e-82 \\
    DDRM & 1.35e-64 && 1.66e-217 && 1.92e-159 && 
4.19e-61 &&3.30e-69 && 5.54e-04 \\
\bottomrule
\end{tabular}
\end{table*}

\begin{figure*}[t]
    \centering
     \subfigure[Models 
     trained on \dataset.]{\label{fig:unfairface_noisy_avg}\includegraphics[width=\textwidth]{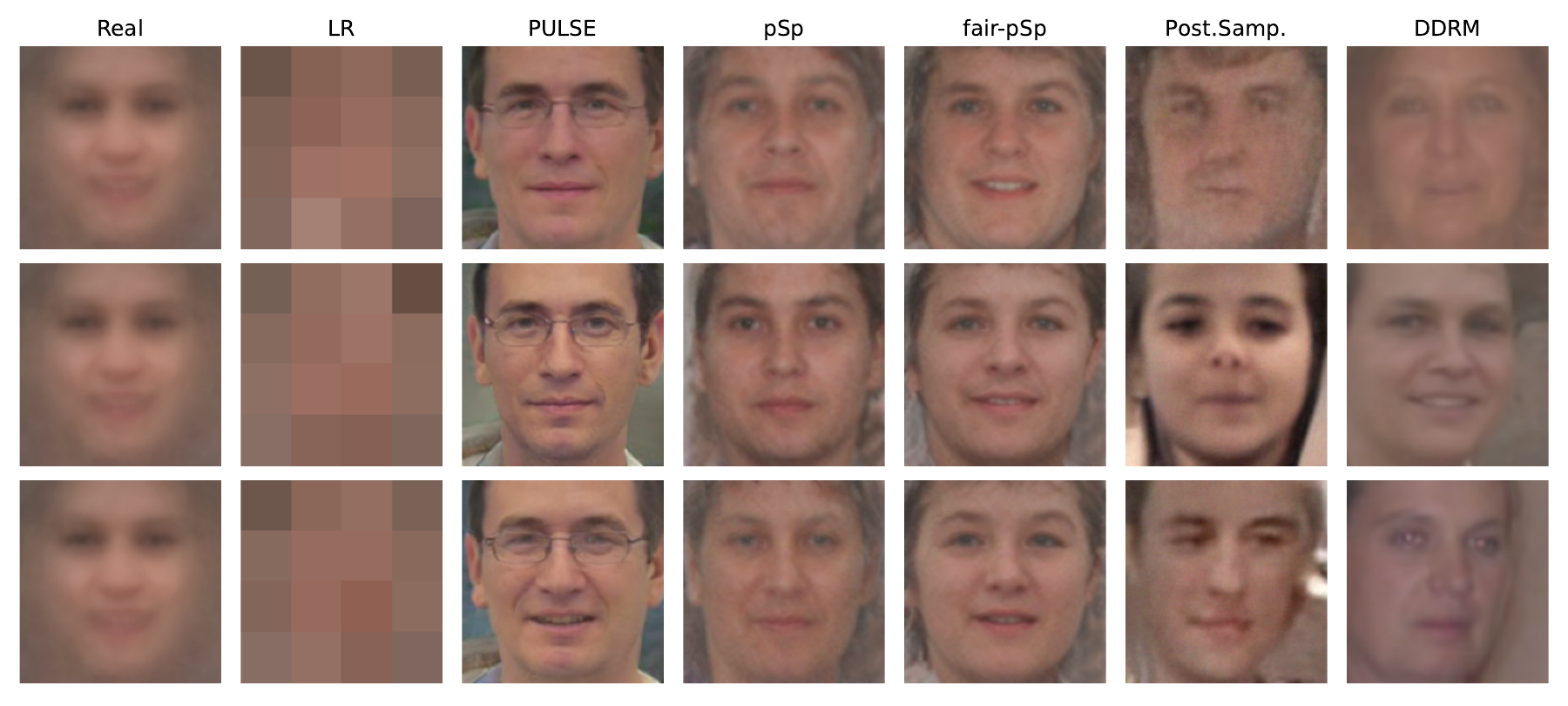}}
  \subfigure[Models trained on FairFace.]{\label{fig:fairface_noisy_avg}\includegraphics[width=\textwidth]{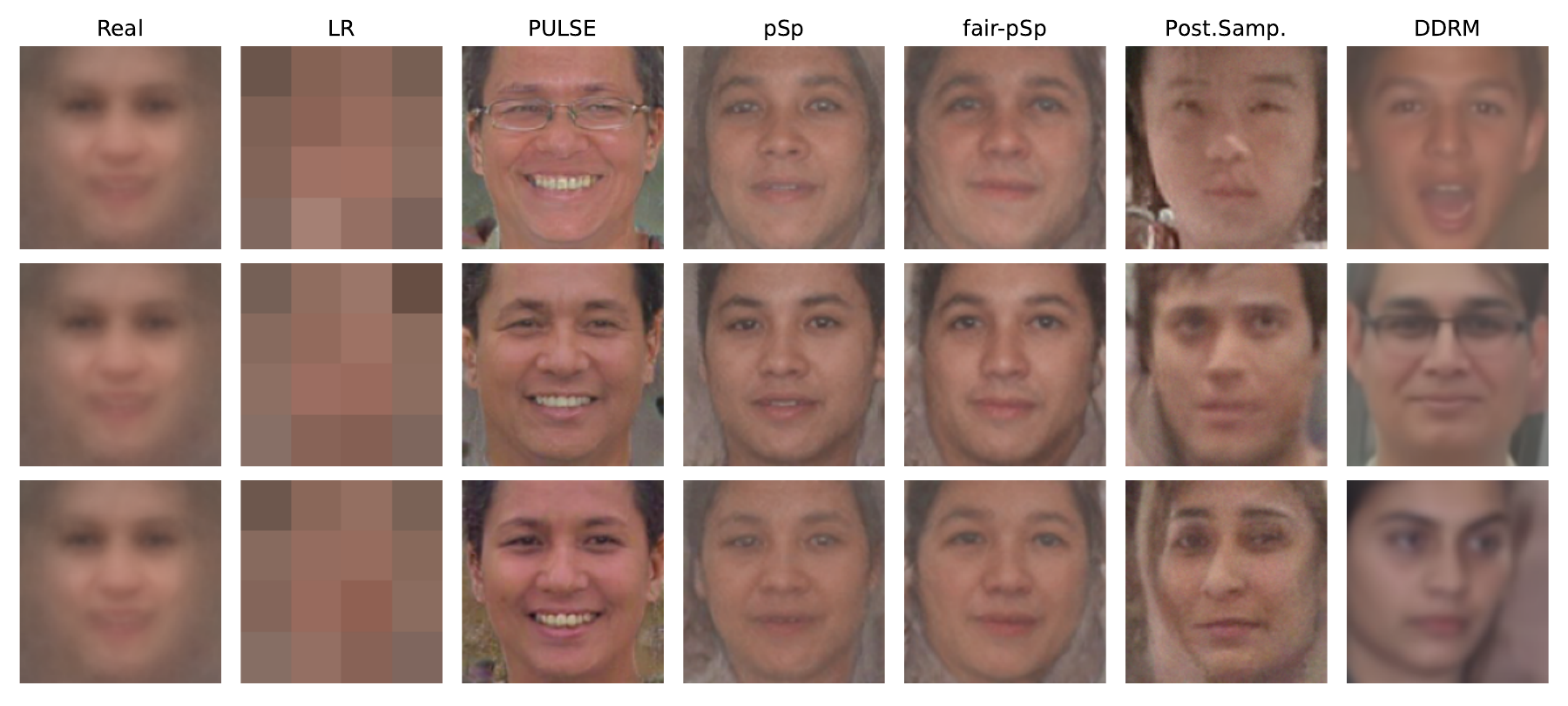}}
  \caption[]{Upsampling results for models trained on \dataset{} and FairFace using uninformative perturbed test samples. The real image is an average over images classified as ``White''.}
  \label{fig:white_noisy_avg}
  \Description{Reconstructions based on several image upsampling algorithms if the input is an uninformative $4\times 4$ image. The inputs are perturbed averages over images classified as ``White''.}
\end{figure*}

\begin{figure*}[t]
    \centering
     \subfigure[Models 
     trained on \dataset.]{\label{fig:unfairface_noisy_avg_black}\includegraphics[width=\textwidth]{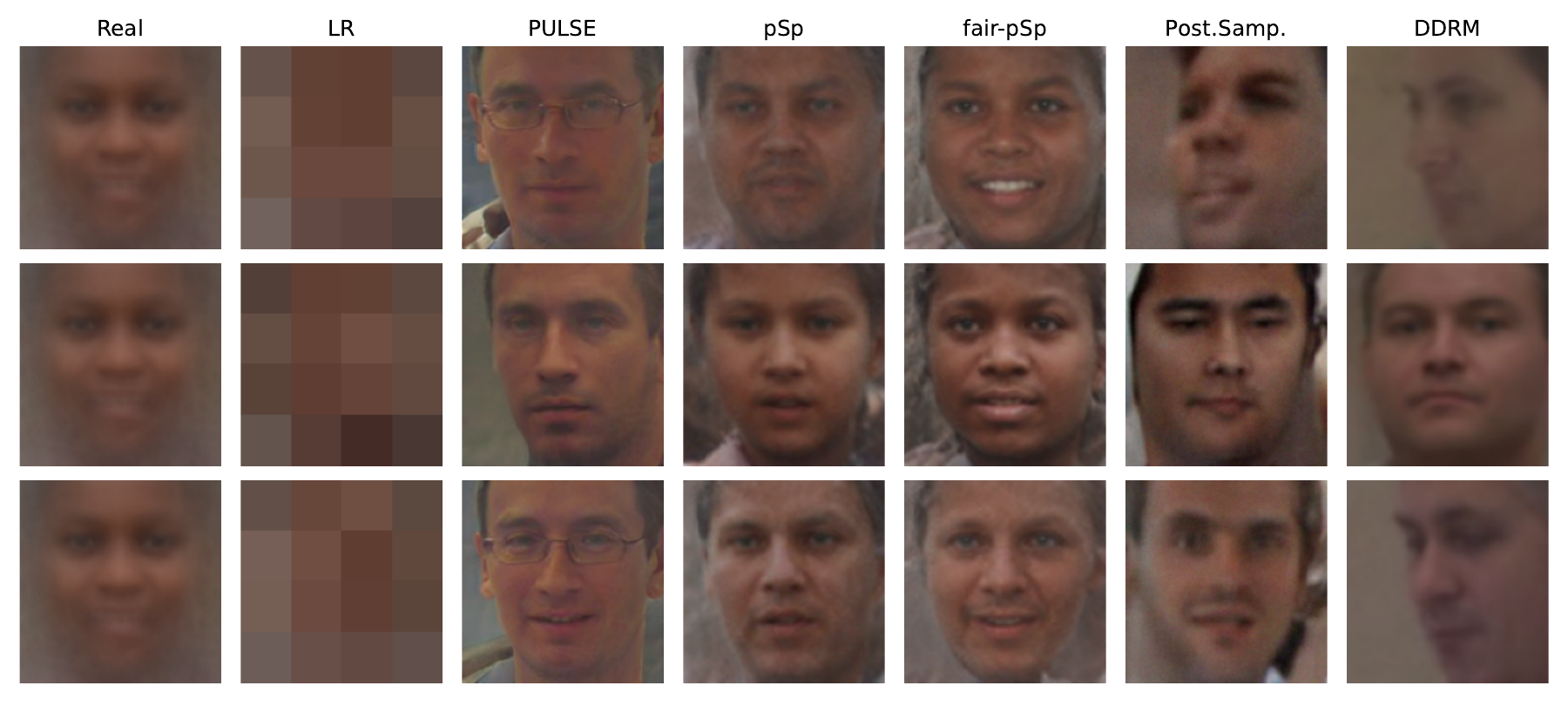}}
  \subfigure[Models trained on FairFace.]{\label{fig:fairface_noisy_avg_black}\includegraphics[width=\textwidth]{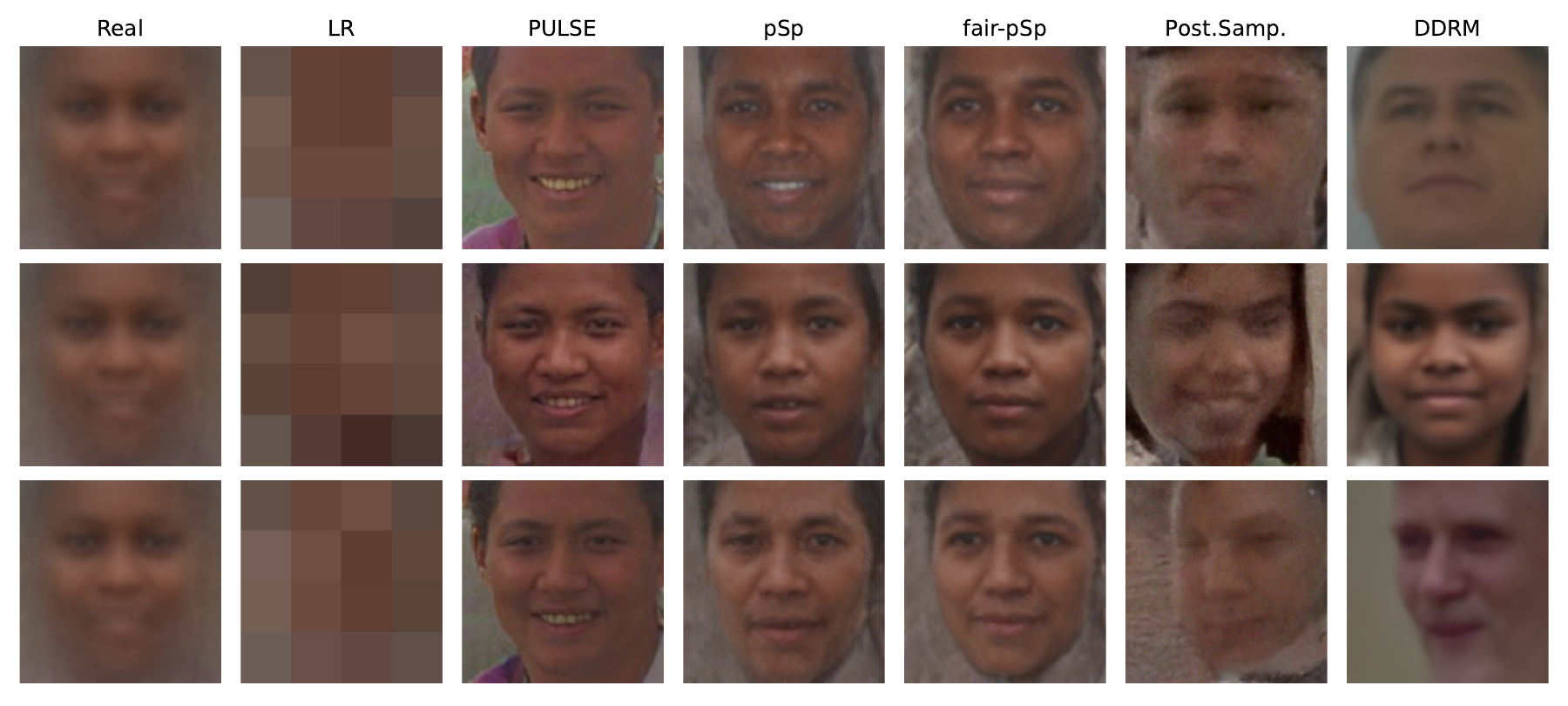}}
  \caption[]{Upsampling results for models trained on \dataset{} and FairFace using uninformative perturbed test samples. The real image is an average over images classified as ``Black''.}
  \label{fig:black_noisy_avg}
  \Description{Reconstructions based on several image upsampling algorithms if the input is an uninformative $4\times 4$ image. The inputs are perturbed averages over images classified as ``Black''.}
\end{figure*}

Lastly, we modify the experiment on diversity in a way that allows us to measure the diversity of pSp and fair-pSp. Recall that since the upsampling procedure in pSp and fair-pSp is deterministic, these algorithms cannot generate diverse reconstructions given a fixed low-resolution input. To circumvent this issue, we randomly perturbed the input by adding a Gaussian noise\footnote{In practice, the color channels of the pixels are encoded by $8$-bit integers, i.e., each color value is in $\{0, 1, \dots, 255\}$. To obtain a perturbed image, we first add the continuous noise, then clip the noisy pixel values to $[0, 255]$, and then map each color channel to an $8$-bit representation by rounding off.} $\varepsilon \sim \mathcal{N}(0, 10)$ as illustrated in Figure~\ref{fig:white_noisy_avg} and Figure~\ref{fig:black_noisy_avg}.
The resulting diversity discrepancies are presented in Table~\ref{tab:diversity_noisy}. Compared to Table~\ref{tab:diversity}, the resulting numbers tend to be slightly lower, which is not surprising given the added source of randomness. More randomness in the inputs is, intuitively, promoting more randomness and diversity in the outputs. While fair-pSp leads to the most diverse reconstructions if trained on UnfairFace, DDRM is still superior if trained on the balanced FairFace dataset. The qualitative results from Figure~\ref{fig:white_noisy_avg} and Figure~\ref{fig:black_noisy_avg} align with the superior performance of DDRM if trained on FairFace.
In both cases, fair-pSp creates more diverse reconstructions than pSp. Again, in all cases the Null hypothesis $\Pucpr=\mathcal{U}([k])$ is rejected.
Nevertheless, it must be emphasized that we cannot recommend the evaluation of diversity using noisy versions, yet. While the results of this experiment align with the original experiments in Section~\ref{sec:exp}, it is generally unclear how to pick the noise variable $\varepsilon$. Specifically, we do not conduct experiments on how the performance is influenced by the particular choice of $\varepsilon$. 
Relatedly,~\citep{richardson2021encoding} propose to replace certain components of the inferred latent code $z$ by random noise to produce multiple reconstructions given a single low-resolution input. 
But again, we think that this procedure introduces ambiguity in the specific choice of components and the noise pattern, leading to a non-trivial evaluation. Especially since a suitable $\varepsilon$ may vary from generative model to generative model.

\begin{table*}[]
    \centering
    \caption[]{Evaluating the diversity discrepancy $D(\Pucpr  \,  \Vert \, \mathcal{U}([k]) )$ for different divergences $D$
    for each algorithm trained on \dataset{} (UFF) and FairFace (FF) based on noisy inputs. Lower scores indicate more diversity. The \xmark{} illustrates that the Null hypothesis $\Pucpr=\mathcal{U}([k])$ is rejected.}
    \label{tab:diversity_noisy}
    \begin{tabular}{lrrcrrcrr}
\toprule
& \multicolumn{2}{c}{$\Delta_{\operatorname{UCPR-}\chi^2}$} && \multicolumn{2}{c}{$\Delta_{\operatorname{UCPR-Cheb}} 
$} && \multicolumn{2}{c}{$\Pucpr = \mathcal{U}([k])$} \\  
\cline{2-3} \cline{5-6} \cline{8-9} \
    & UFF & FF && UFF & FF && UFF & FF  \\ 
\midrule
PULSE & 5.11 & 1.63 &&  0.79 & 0.35&& \xmark & \xmark \\
pSp & 2.20 & 2.57 &&  0.43 & 0.55&&\xmark & \xmark \\
fair-pSp & 1.21 & 0.73 && 0.33 & 0.18&&\xmark & \xmark \\
Post.Samp. & 2.55 & 0.28 && 0.54 & 0.14&&\xmark & \xmark \\
DDRM & 4.88 & 0.13 && 0.77 & 0.08 && \xmark & \xmark\\
\bottomrule
\end{tabular}
\end{table*}

\end{document}